%% file: main.tex
\documentclass[lettersize,journal]{IEEEtran}
\usepackage{amsmath,amsfonts}
\usepackage{algorithm}
\usepackage{array}
\usepackage[caption=false,font=normalsize,labelfont=sf,textfont=sf]{subfig}
\usepackage{textcomp}
\usepackage{stfloats}
\usepackage{url}
\usepackage{verbatim}
\usepackage{graphicx}
\usepackage{cite}
\usepackage{listings}
\usepackage{algpseudocode}
\usepackage{pgf-pie}   
\usepackage{wrapfig}
\usepackage{mathrsfs}
\usepackage{multirow}
\newtheorem{myDef}{Definition}

\usepackage{pgfplots}
\usepackage{enumitem}
\usepackage{array}
\usepackage[caption=false,font=normalsize,labelfont=sf,textfont=sf]{subfig}
\usepackage{makecell}
\usepackage{tikz}
\usepgfplotslibrary{groupplots}
\pgfplotsset{compat=1.17} 
\usepackage{etoolbox}
\usepackage{algorithm}

\usepackage[utf8]{inputenc} 
\usepackage[T1]{fontenc}    
\usepackage{booktabs}       
\usepackage{nicefrac}       
\usepackage{microtype}      
\usepackage{xcolor}         
\usepackage[most]{tcolorbox}
\usepackage{cite}
\usepackage{amsmath,amssymb,amsfonts}
\usepackage{textcomp}

\lstdefinestyle{arxivcode}{
  basicstyle=\ttfamily\scriptsize,
  numbers=left,
  numberstyle=\tiny,
  stepnumber=1,
  numbersep=6pt,
  xleftmargin=10pt,
  columns=fullflexible,
  keepspaces=true,
  breaklines=true,
  breakatwhitespace=false,
  showstringspaces=false,
  upquote=true,
  literate={\"}{{\char34}}1
}

\begin{document}

\title{\textsc{Pandora:} Leveraging Code-driven Knowledge Transfer for Unified Structured Knowledge Reasoning}

\author{
Yongrui Chen\textsuperscript{\rm a,b},
Junhao He\textsuperscript{\rm a,b},
Linbo Fu\textsuperscript{\rm a,b},
Shenyu Zhang\textsuperscript{\rm a,b},
Rihui Jin\textsuperscript{\rm a,b},
Xinbang Dai\textsuperscript{\rm a,b}, \\
Jiaqi Li\textsuperscript{\rm a,b},
Dehai Min\textsuperscript{\rm a,b},
Nan Hu\textsuperscript{\rm a,b},
Yuxin Zhang\textsuperscript{\rm a,b},
Guilin Qi\textsuperscript{\rm a,b,}
Yi Huang\textsuperscript{\rm c},
Tongtong Wu\textsuperscript{\rm d}  \\

\IEEEcompsocitemizethanks{\IEEEcompsocthanksitem \textsuperscript{\rm a}
School of Computer Science and Engineering,
Southeast University, Nanjing 210096, China.
\textsuperscript{\rm b}
the Key Laboratory of New Generation Artificial Intelligence Technology and Its Interdisciplinary Applications (Southeast University), Ministry of Education, China.
\textsuperscript{\rm c} China Mobile Research.
\textsuperscript{\rm d} Department of Data Science \& AI, Monash University, Australia.
E-mail: \{yongruichen, gqi\}@seu.edu.cn}
}




\maketitle

\begin{abstract}
Unified Structured Knowledge Reasoning (USKR) aims to answer natural language questions by using structured sources such as tables, databases, and knowledge graphs in a unified way. Existing USKR methods rely on task-specific strategies or bespoke representations, which hinder their ability to dismantle barriers between different SKR tasks, thereby constraining their overall performance in cross-task scenarios.
In this paper, we introduce \textsc{Pandora}, a novel USKR framework that addresses the limitations of existing methods by leveraging two key innovations. First, we propose a code-based unified knowledge representation using \textsc{Python}'s \textsc{Pandas} API, which aligns seamlessly with the pre-training of LLMs.
This representation facilitates a cohesive approach to handling different structured knowledge sources.
Building on this foundation, we employ knowledge transfer to bolster the unified reasoning process of LLMs by automatically building cross-task memory.  By leveraging multi-stage code-driven reasoning and adaptively refining its outputs through execution feedback, \textsc{Pandora} demonstrates strong and unified reasoning capabilities across heterogeneous data sources.
Extensive experiments on seven widely used benchmarks across Text-to-SQL, KGQA, and TableQA demonstrate that \textsc{Pandora} outperforms existing unified reasoning frameworks and competes effectively with task-specific methods. Our code is available at: \url{https://github.com/Bahuia/pandora}.
\end{abstract}

\begin{IEEEkeywords}
Structured Knowledge Reasoning, Large Language Model, Unified Knowledge Representation, Database Interface, Knowledge Graph Question Answering.
\end{IEEEkeywords}

\section{Introduction}
\label{sec:motivation}
Structured knowledge reasoning (SKR)~\cite{DBLP:conf/acl/PasupatL15,DBLP:conf/emnlp/YuZYYWLMLYRZR18,DBLP:conf/acl/YihRMCS16} entails answering natural language questions over structured knowledge sources, such as tables~\cite{DBLP:journals/tkde/LiLFW24}, databases (DBs)~\cite{DBLP:journals/tkde/ZhangWSWTQCWY24}, and knowledge graphs (KGs)~\cite{DBLP:journals/tkde/PanLWCWW24}. It underpins diverse intelligent applications, including legal judgment~\cite{DBLP:journals/access/CuiSW23}, disease diagnosis~\cite{DBLP:journals/artmed/LiWYWLJSTCWL20}, and investment analysis~\cite{DBLP:conf/sigmod/ZhangMFMG0LL24}, while serving as an important way to alleviate \textit{hallucinations} of Large Language Models (LLMs). Leveraging LLM's ability to learn from massive corpus, recent work~\cite{DBLP:conf/sigir/YeHYLHL23,DBLP:journals/pacmmod/LiZLFZZWP0024,DBLP:conf/aaai/NieZW024} has achieved notable progress in reasoning over single-type structured knowledge.

However, real-world applications increasingly demand seamless integration across diverse structured knowledge representations. For instance, a medical decision support platform~\cite{antoniadi2021current} must jointly query patient DBs for clinical records and drug KGs for interaction networks, while an investment analytics tool~\cite{DBLP:conf/sigmod/ZhangMFMG0LL24} might fuse financial tables with market KGs to forecast trends. The inherent disparities in these knowledge representations, such as relational schemas in DBs versus graph-based triples in KGs, impede effective unified reasoning, stifling advancements in real-world applications~\cite{DBLP:journals/tkde/ZhangZZZWZ24}.  In particular, traditional SKR methods exacerbate this challenge through task-specific designs~\cite{DBLP:conf/emnlp/PourrezaR24,DBLP:conf/aaai/NieZW024}, \textbf{which limit generalization and fail to bridge representational gaps}, often resulting in fragmented reasoning and suboptimal performance in cross-knowledge scenarios.
\begin{figure*}[t]
\centering
	\includegraphics[width=0.92\textwidth]{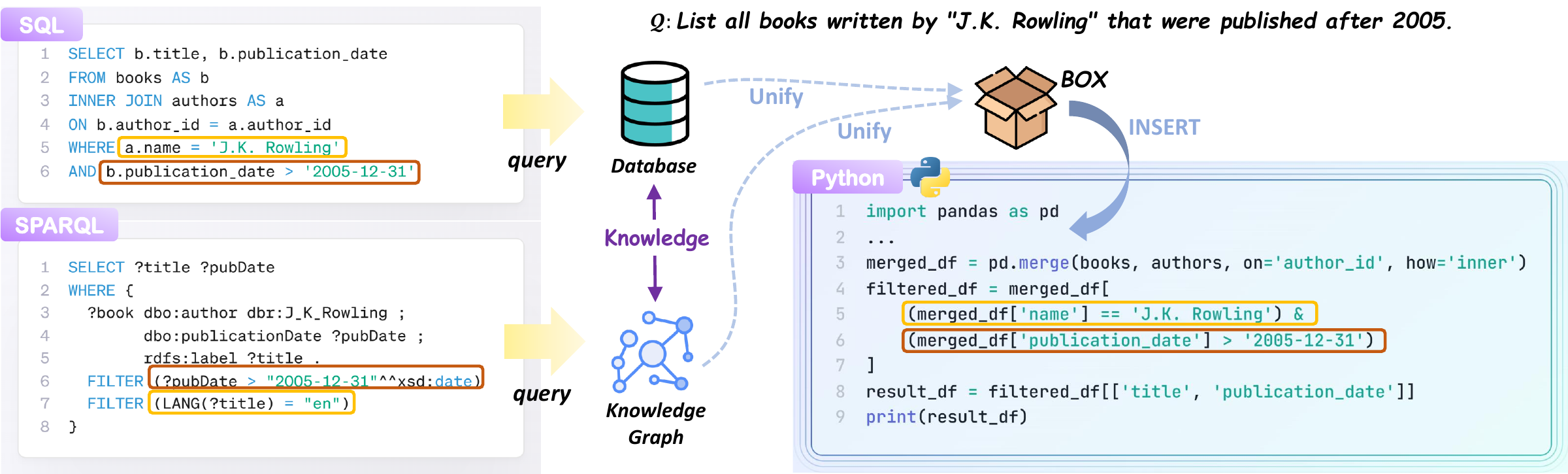}
	\caption{Comparison of SQL, SPARQL, and \textsc{Python} implementations for the same natural language question ($\mathcal{Q}$). Color-coding highlights the direct correspondence between their reasoning steps, illustrating the potential for a unified query execution framework.} \label{fig:example}
\end{figure*}
Building on well-aligned LLMs as the foundation, recent studies have proposed several unified SKR frameworks~\cite{DBLP:conf/emnlp/JiangZDYZW23,DBLP:conf/acl/ChengZXYZQHCL0R24,DBLP:journals/corr/abs-2406-18916}.
Although these methods achieved the integration of different reasoning scenarios by relying on task-specific strategies or custom-defined representations, they still \textbf{do not fundamentally break the natural barriers between different structured knowledge}, thus resulting in limited performance and coverage.
We argue that \textbf{an ideal unified SKR framework should facilitate cross-task knowledge transfer}. As illustrated in Figure~\ref{fig:example}, SQL and SPARQL queries may differ syntactically yet share equivalent semantics. Mapping such heterogeneous queries into a unified representation enables LLMs to leverage shared knowledge across SKR tasks, thereby boosting target task performance.

In this paper, we propose \textsc{Pandora}, a novel unified structured knowledge reasoning (SKR) framework that leverages Python code to uniformly represent and execute reasoning processes across diverse structured knowledge sources. Specifically, \textsc{Pandora} begins by transforming heterogeneous structured knowledge sources into a unified representation built upon \textsc{Python} \textsc{Pandas} DataFrames, referred to as BOX. This transformation is grounded in a simple yet effective strategy that proves sufficient to accommodate mainstream structured data formats, including tables, relational databases, and knowledge graphs.
For each NLQ, the \textsc{Pandora} agent leverages the LLM to perform step-by-step reasoning, iteratively generating and refining executable Python code via Pandas APIs. A key advantage of this code-driven approach is its staged reasoning pipeline: unlike SQL or SPARQL, which demand a complete, monolithic query in a single pass, Python code naturally supports incremental and fine-grained reasoning. \textsc{Pandora} decomposes each NLQ into sequential subtasks, independently generates and validates code for each, and accumulates intermediate execution states as reasoning progresses.
The final code is executed to derive the answer. Figure~\ref{fig:example} illustrates an example of Python-based reasoning for a representative query, alongside its corresponding SQL and SPARQL counterparts. To enable the LLM to learn a mapping from NLQs to Pandas APIs while maintaining robustness in low-resource scenarios, the memory module is automatically constructed from only a few training examples. Notably, our empirical validation demonstrates that the unified code-based representation facilitates cross-task knowledge sharing, enabling mutual performance gains across diverse reasoning tasks. We conduct extensive experiments across three structured knowledge reasoning tasks: Text-to-SQL, TableQA, and KGQA, on seven widely used benchmarks.
Experimental results demonstrate that \textsc{Pandora} outperforms all existing unified structured knowledge reasoning frameworks and achieves performance comparable to task-specific methods.  Remarkably, \textsc{Pandora} remains highly competitive even in zero-shot settings, highlighting its generalizability and robustness.
In summary, our contributions include:
\begin{itemize}[leftmargin=1em]
\item We introduce a code-centric representation that unifies heterogeneous structured knowledge sources, enabling seamless interoperability across diverse data formats and eliminating task-specific operational barriers.

\item We propose \textsc{Pandora}, a novel USKR framework that leverages a multi-stage, code-driven reasoning pipeline to facilitate cross-task knowledge transfer and significantly enhance the reasoning capabilities of LLMs.

\item We conduct comprehensive experiments on seven mainstream benchmarks spanning three structured knowledge reasoning tasks, demonstrating that \textsc{Pandora} achieves state-of-the-art performance among unified reasoning frameworks while remaining competitive with task-specific approaches.
\end{itemize}

\section{Preliminary}

Following ~\cite{DBLP:conf/emnlp/JiangZDYZW23}, we focus on the following three types of structured knowledge:

\noindent \textbf{Data Table}
A table is considered as $\mathcal{T}$ = ($\{c_i\}_{i=1}^C$, $\{r_j\}_{j=1}^R$, $\{v_{i,j}\}_{i=1,j=1}^{C,R}$), where $c_i$ denotes the $i$-th column name and $r_j$ denotes a data record indexed by columns. $v_{i,j}$ denotes the content of the cell located at the intersection of $c_i$ and $r_j$.

\noindent \textbf{Database}
A database $\mathcal{D}$ consists of multiple tables, represented as $\mathcal{D} = \{\mathcal{T}_1, \mathcal{T}_2, \dots, \mathcal{T}_T\}$.
The foreign keys across all tables are also available to link the data from two tables, denoted as $\{(c_i^{p}, c_j^{q})\}$, where $c_i^{p}$ and $c_j^{q}$ denote the $i$-th and $j$-th columns in the $\mathcal{T}_p$ and$\mathcal{T}_q$, respectively.

\noindent \textbf{Knowledge Graph}
A knowledge graph (KG) is typically a collection of subject-predicate-object triples, denoted by $\mathcal{K}=\{\left \langle s, p, o\right \rangle | s \in \mathcal{E}, p \in \mathcal{R}, o \in \mathcal{E} \cup \Gamma \}$, where $\mathcal{E}$, $\mathcal{R}$, and $\Gamma$ denote the entity set, relation set, and type set respectively.

\noindent \textbf{Problem Formulation}
Given a natural language question $\mathcal{Q}$ and a structured knowledge source $\mathcal{S}$ (such as a table $\mathcal{T}$, a relational database $\mathcal{D}$, or a KG $\mathcal{K}$), the objective is to accurately derive the target answer $\mathcal{A}$ by reasoning over $\mathcal{S}$.

\section{Code-based Unified Representation}
To facilitate knowledge transfer across various SKR tasks, we propose a unified knowledge representation called BOX, which is defined as follows:
\begin{myDef}[BOX]
A BOX is denoted by $\mathcal{B} = (b, \Phi, \Psi)$, where
$b$ denotes its textual name, $\Phi = \{\phi_i\}_{i=1}^N$, and $\Psi = \{[\psi_j^{\phi_i}]_{j=1}^{M}\}_{i=1}^N$. $\phi_i$ denotes a field that can be a column name or a KG relation. $\psi^{\phi_i}_j$ denotes the $j$-th value associated with the field $\phi_i$ that can be a cell content or a KG entity.
\end{myDef}
Unlike existing unified representations~\cite{DBLP:conf/emnlp/JiangZDYZW23,DBLP:conf/acl/ChengZXYZQHCL0R24,DBLP:journals/corr/abs-2406-18916}, our proposed BOX $\mathcal{B} = (b, \Phi, \Psi)$ uses Python Pandas DataFrames as a universal interface. \textbf{Code's structured, compositional nature is ideal, as LLMs excel at understanding, generating, and reasoning with it due to extensive pre-training on programming languages}~\cite{dubey2024llama}. This bridges the gap between diverse input representations and the LLM's inherent capabilities.
Here is an example of our defined BOX pattern:
\begin{tcolorbox}[colback=white, colframe=black, arc=2mm, boxrule=0.5pt]
\scriptsize
\begin{lstlisting}[style=arxivcode,language=python]
import pandas as pd
b = pd.DataFrame({phi_1: [psi_1_1, ...],
                  phi_2: [psi_2_1, ...], ...})
\end{lstlisting}
\end{tcolorbox}
\noindent where \texttt{phi\_1} and \texttt{psi\_2\_1} are code rewritings of $\phi_1$ and $\psi_1^{\phi_2}$, respectively.
The choice of \textsc{Pandas} is motivated by two key advantages:
a) \textbf{Alignment with LLM Pretraining:} LLMs are extensively trained on \textsc{Python} code, making them inherently adept at understanding and generating Pandas operations, thus narrowing the gap between input format and model capability.
b) \textbf{Flexible Manipulation:} \textsc{Pandas} provides rich APIs for data filtering, joining, aggregation, and other manipulations, enabling natural encoding of diverse reasoning patterns required for table, DBs, and KGs.
For instance, in Figure~\ref{fig:example}, \texttt{pd.merge} is first used to join the BOX \texttt{author} and the BOX \texttt{book} to form a new BOX \texttt{df}.
Then, a filtering operation is applied to \texttt{df} using a boolean index:
\begin{tcolorbox}[colback=white, colframe=black, arc=2mm, boxrule=0.5pt]
\scriptsize
\begin{lstlisting}[style=arxivcode,language=python]
result = df[(df['name'] == 'J.K.Rowling')
            & (df['publication_date'] > 2005)]
\end{lstlisting}
\end{tcolorbox}
More examples are listed in \ref{sec:pandas_example}.
\begin{figure*}
\centering
	\includegraphics[width=1\textwidth]{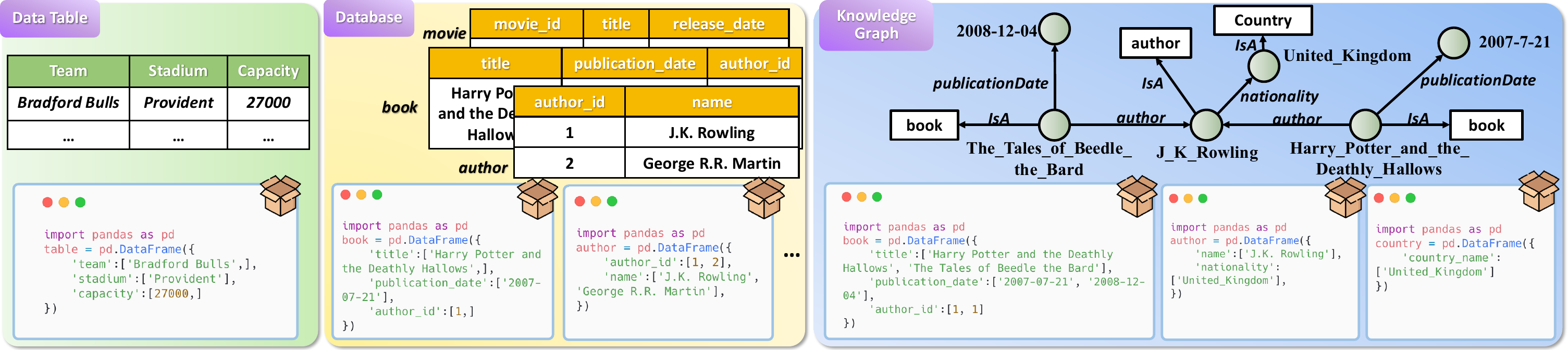}
	\caption{Examples illustrating the conversion of tables, databases, and KG subgraphs, into BOX representations. A subset of fields and values is displayed for brevity. In the KG case, each entity type is represented as a BOX that encodes both its attributes and its relationships with other entities.} \label{fig:box}
\end{figure*}
In fact, structured knowledge can be effectively represented in the form of BOX, as depicted in Figure~\ref{fig:box}. We will detail the conversion of Tables, DBs, and KGs into the BOX representation.

\subsection{Table-to-BOX}
\label{sec:table-to-box}
As shown in Figure~\ref{fig:box}(a), a data table $\mathcal{T} = (\{c_i\}_{i=1}^C, \{r_j\}_{j=1}^R, \{v_{i,j}\}_{i=1,j=1}^{C,R})$, can be seamlessly transformed into one box $\mathcal{B} = (b, \{\phi_i\}_{i=1}^C, \{[\psi_j^{\phi_i}]_{j=1}^R\}_{i=1}^C)$ by treating each column name $c_i$ as a field name $\phi_i$ and the content of each table cell $v_{i,j}$ as a field value $\psi_j^{\phi_i}$.
The detailed algorithm is provided in Algorithm~\ref{alg:table-to-box}.
\begin{algorithm}[ht]
\caption{Conversion from Table to BOX \label{alg:table-to-box}}
\begin{algorithmic}[1]
\Require A data table $\mathcal{T} = (\{c_i\}_{i=1}^C, \{r_j\}_{j=1}^R,  \{v_{i,j}\}_{i=1,j=1}^{C,R})$,  where $c_i$ denotes the $i$-th column name and each row $r_j$ denotes a data record. $v_{i,j}$ denotes the content.
\State Initialize the BOX field set as $\Phi \gets \emptyset$ and the BOX value set as $\Psi \gets \emptyset$.
\Function{TableToBOX}{$\mathcal{T}$}
\For{$i = 1$ to $C$}
\State $\Phi \gets \Phi \cup \{c_i\}$ \Comment{Treat each column as a field.}
\EndFor
\For{$i = 1$ to $C$}
\For{$j = 1$ to $R$}
\State $\Psi \gets \Psi \cup \{v_{i,j}\}$ \Comment{Treat each cell content as a field value.}
\EndFor
\EndFor
\If{$\mathcal{T}$ has a table name $t$} \Comment{Name the each BOX in \textsc{Pandas} code.}
\State $\mathcal{B} \gets (t, \Phi, \Psi)$
\Else
\State $\mathcal{B} \gets (\texttt{Table}, \Phi, \Psi)$
\EndIf
\State \textbf{return} $\mathcal{B}$
\EndFunction
\State $\mathcal{B} = \textsc{TableToBOX}(\mathcal{T})$
\State \Return $\mathcal{B}$ \Comment{A table can be converted into a single BOX.}
\end{algorithmic}
\end{algorithm}

\subsection{DB-to-BOX}
As illustrated in Figure~\ref{fig:box}(b), for a database $\mathcal{D} = \{\mathcal{T}_i\}_{i=1}^T$, each table $\mathcal{T}_i \in \mathcal{D}$ is converted to a box $\mathcal{B}_i$ following the procedure of table-to-BOX. To explicitly define the relational schema and semantic linkages among the tables, the foreign key information $\{(\phi_i^{p}, \phi_j^{q})\}$ is retained, where $\phi_i^{p}$ and $\phi_j^{q}$ represent the $i$-th field in $\mathcal{B}_p$ and the $j$-th field in $\mathcal{B}_q$, respectively.
The detailed conversion process is detailed in Algorithm~\ref{alg:db-to-box}.
\begin{algorithm}
\caption{Conversion from Database to BOX \label{alg:db-to-box}}
\begin{algorithmic}[1]
\Require A database $\mathcal{D} = \{\mathcal{T}_1, \mathcal{T}_2, \dots, \mathcal{T}_T\}$, where $\mathcal{T}_i$ denotes a table.
\State Initialize the BOX set as $\mathcal{B}^* \gets \emptyset$.
\For{$i = 1$ to $T$}
\State $\mathcal{B}_i = \textsc{TableToBOX}(\mathcal{T}_i)$ \Comment{Follow Algorithm~\ref{alg:table-to-box} to generate the BOX corresponding to each table.}
\State $\mathcal{B}^* \gets \mathcal{B}^* \cup \{\mathcal{B}_i\}$
\EndFor
\State \Return $\mathcal{B}^*$
\end{algorithmic}
\end{algorithm}

\subsection{KG-to-BOX}
\begin{algorithm}[!htbp]
\caption{Conversion from KG to BOX \label{alg:kg-to-box}}
\begin{algorithmic}[1]
\Require A knowledge graph $\mathcal{K} = \{\langle s, p, o \rangle \mid s \in \mathcal{E}, p \in \mathcal{R}, o \in \mathcal{E} \cup \Gamma\}$, where $\mathcal{E}$, $\mathcal{R}$, and $\Gamma$ denote the entity set, relation set, and type set. A topic entity set $\mathcal{E}^* \subset \mathcal{E}$. A relevant relation set $\mathcal{R}^* \subset \mathcal{R}$.
\State Initialize the BOX set as $\mathcal{B}^* \gets \emptyset$, the field record list $\Omega \gets [~]$, a visited entity set $\mathcal{V}$.
\Function{DepthFirstSearch}{$e, \omega, H$}
\If{$|\omega| = 2 \times H$}
\State Append $\omega$ to $\Omega$
\State \textbf{return}
\EndIf
\For{$r \in \mathcal{R}^*$}
\State $\mathcal{E}^+ \gets \text{GetNeighborEnt}(e, r, +)$
\State $\mathcal{E}^- \gets \text{GetNeighborEnt}(e, r, -)$
\For{$e^+ \in \mathcal{E}^+$} \Comment{Traverse the one-hop neighbor entities that start from $e$ through $r$.}
\If{$e^+ \notin \mathcal{V}$} \Comment{Prune. Prevent passing through the same entity.}
\State Append $[\Gamma(e), \Gamma(e), e]$ to $\omega$ \State Append $[\Gamma(e), r, e^+]$ to $\omega$, $\mathcal{V} \gets \mathcal{V} \cup \{e^+\}$
\State \textsc{DepthFirstSearch}($e^+, \omega, H$)
\State $\text{Pop}(\omega)$, $\text{Pop}(\omega)$, $\mathcal{V} \gets \mathcal{V} \setminus \{e^+\}$
\EndIf
\EndFor
\For{$e^- \in \mathcal{E}^-$} \Comment{Traverse the one-hop neighbor.}
\If {$e^- \notin \mathcal{V}$}
\State Append $[\Gamma(e^-), \Gamma(e^-), e^-]$ to $\omega$
\State Append $[\Gamma(e^-), r, e]$ to $\omega$, $\mathcal{V} \gets \mathcal{V} \cup \{e^-\}$
\State \textsc{DepthFirstSearch}($e^-, \omega, H$)
\State $\text{Pop}(\omega)$, $\text{Pop}(\omega)$, $\mathcal{V} \gets \mathcal{V} \setminus \{e^-\}$
\EndIf
\EndFor
\EndFor
\EndFunction
\For{$e \in \mathcal{E}^*$}
\State $\text{DepthFirstSearch}(e, [~], H)$
\EndFor
\For{$\omega \in \Omega$}  \Comment{An empty BOX with only field names.}
\For{$(b, \phi, \psi) \in \omega$}
\If{ $\not \exists~\Phi_b$ }
\State $\Phi_b \gets \emptyset$, $\Psi_b \gets \emptyset$
\EndIf
\State $\Phi_b \gets \Phi_b \cup \{\phi\}$ \Comment{Add field for each BOX.}
\EndFor
\EndFor
\For{$\omega \in \Omega$}  \Comment{Second, fill values into each field.}
\For{$(b, \phi, \psi) \in \omega$}
\If{$\not \exists~\Psi_b^{\phi}$}
\State $\Psi_b^{\phi} \gets [~]$
\EndIf
\State Append $\psi$ to $\Psi_b^{\phi}$ \Comment{Add value for each field.}
\For{$\tilde{\phi} \in \Phi \setminus \{\phi\} $}
\State Append \texttt{NA} to $\Psi_b^{\tilde{\phi}}$ \Comment{Padding with \texttt{NA}.}
\EndFor
\EndFor
\EndFor
\State \Return $\mathcal{B}^*$
\end{algorithmic}
\end{algorithm}
Figure~\ref{fig:box}(c) shows an example of KG-to-BOX.
Typically, since the full KG $\mathcal{K} = \{\langle s, p, o \rangle \mid s \in \mathcal{E}, p \in \mathcal{R}, o \in \mathcal{E} \cup \Gamma\}$ is too large for efficient querying, it is necessary to extract a topic subgraph for each NLQ $\mathcal{Q}$.
Concretely, a depth-first search is performed to extract the $H$-hop subgraph $\mathcal{K}^* \subset \mathcal{K}$ for each \textit{topic entity} mentioned in $\mathcal{Q}$.
Following a common practice~\cite{DBLP:conf/www/GuKVSLY021,DBLP:conf/acl/YihRMCS16} in KGQA, topic entities $\mathcal{E}^* \subset \mathcal{E}$ are directly obtained from the dataset annotations, in accordance with a widely adopted convention that guarantees fair and reproducible comparisons with all baseline methods. In open-domain settings, topic entities can be obtained via an external entity linking service (e.g., BLINK or TAGME).
Correspondingly, we use $\Gamma^* \subset \Gamma$ to denote the entity types appearing in $\mathcal{K}^*$.

To further narrow down the search space, we leverage the preprocessed data from~\cite{DBLP:conf/emnlp/XieW0ZSYWZYWZWL22}, retaining only the relations $\mathcal{R}^* \subseteq \mathcal{R}$ that demonstrate high embedding similarity to $\mathcal{Q}$.
Subsequently, for each entity type $\gamma \in \Gamma^*$ and its corresponding entity set $\mathcal{E}_{\gamma} = \{e \mid \exists \langle e, \text{IsA}, \gamma \rangle \in \mathcal{K}^*\}$, a BOX $\mathcal{B}_{\gamma} = (\gamma, \Phi_{\gamma}^{1:N}, \Psi_{\gamma}^{1:N})$ is constructed, where $N = |\Phi_{\gamma}|$ denotes the number of fields in the BOX for type $\gamma$; concretely, $N = 1 + |\{\phi_i \mid \phi_i \in \mathcal{R}^*, \exists \langle s, \phi_i, o \rangle \in \mathcal{K}^*, s \in \mathcal{E}_{\gamma}\}|$, where the additional field stores the entity identifier/type field.
Specifically, the field names $\Phi_{\gamma}^{1:N} = \Phi_{\gamma}^1 \cup \Phi_{\gamma}^{2:N}$, where $\Phi_{\gamma}^1 = \{\gamma\}$, and $\Phi_{\gamma}^{2:N} = \{\phi_i \mid \phi_i \in \mathcal{R}^*, \exists \langle s, \phi_i, o \rangle \in \mathcal{K}^*, s \in \mathcal{E}_{\gamma}\}_{i=2}^N$ consists of $1$-hop relations originating from the entities in $\mathcal{E}_{\gamma}$. Similarly, the field values $\Psi_{\gamma} = \Psi_{\gamma}^1 \cup \Psi_{\gamma}^{2:N}$, where $\Psi_{\gamma}^1 = \{[\psi_{1,j} \mid \psi_{1,j} \in \mathcal{E}_{\gamma}]_{j=1}^{M}\}_{i=1}^1$ corresponds to $\Phi_{\gamma}^1$ and contains the entities of type $\gamma$. $\Psi_{\gamma}^{2:N} = \{[\psi_{i,j} \mid \exists \langle s, p, \psi_{i,j} \rangle \in \mathcal{K}^*, s \in \mathcal{E}_{\gamma}, p \in \mathcal{R}^*]_{j=1}^M\}_{i=2}^N$ corresponds to $\Phi_{\gamma}^{2:N}$ and consists of the $1$-hop neighbors of the subject entities in $\mathcal{E}_{\gamma}$ through the relations in $\mathcal{R}^*$.

 To put it more intuitively, from the perspective of the KG, for a box $\mathcal{B}_{\gamma}$, $\psi_{1,j}$ serves as the subject, $\phi_i$ represents the predicate, and $\psi_{i,j}$ acts as the object. In this way, complex multi-hop reasoning is efficiently executed as a sequence of highly-optimized join operations such as \texttt{pandas.merge}, as shown in Figure~\ref{fig:example}.
After all BOXes are built, the foreign key  is defined as $\{(\phi_i^{p}, \phi_j^{q})\}$, where $\phi_i^{p}$ and $\phi_j^{q}$ share at least one common entity.
For N-ary relations encoded via intermediate nodes, such as Freebase \textit{Compound Value Type} (CVT) nodes, each CVT node is treated as an ordinary entity and assigned a dedicated BOX, with all of its incident relations represented as fields. For example, a triple pattern of the form $(x, r_1, \text{CVT}),\ (\text{CVT}, r_2, y),\ (\text{CVT}, \text{time}, 1998)$ yields a single CVT BOX that jointly captures all three relational fields. Although CVT nodes carry no intrinsic semantic meaning, they function as structural anchors that preserve a uniform BOX-based representation. During the reasoning process, the LLM reconstructs the original N-ary relational structure by invoking \texttt{pd.merge} on shared CVT entities across BOXes, thereby recovering the semantics without any loss of information.

\section{Pandora}

\begin{figure*}
\centering
	\includegraphics[width=\textwidth]{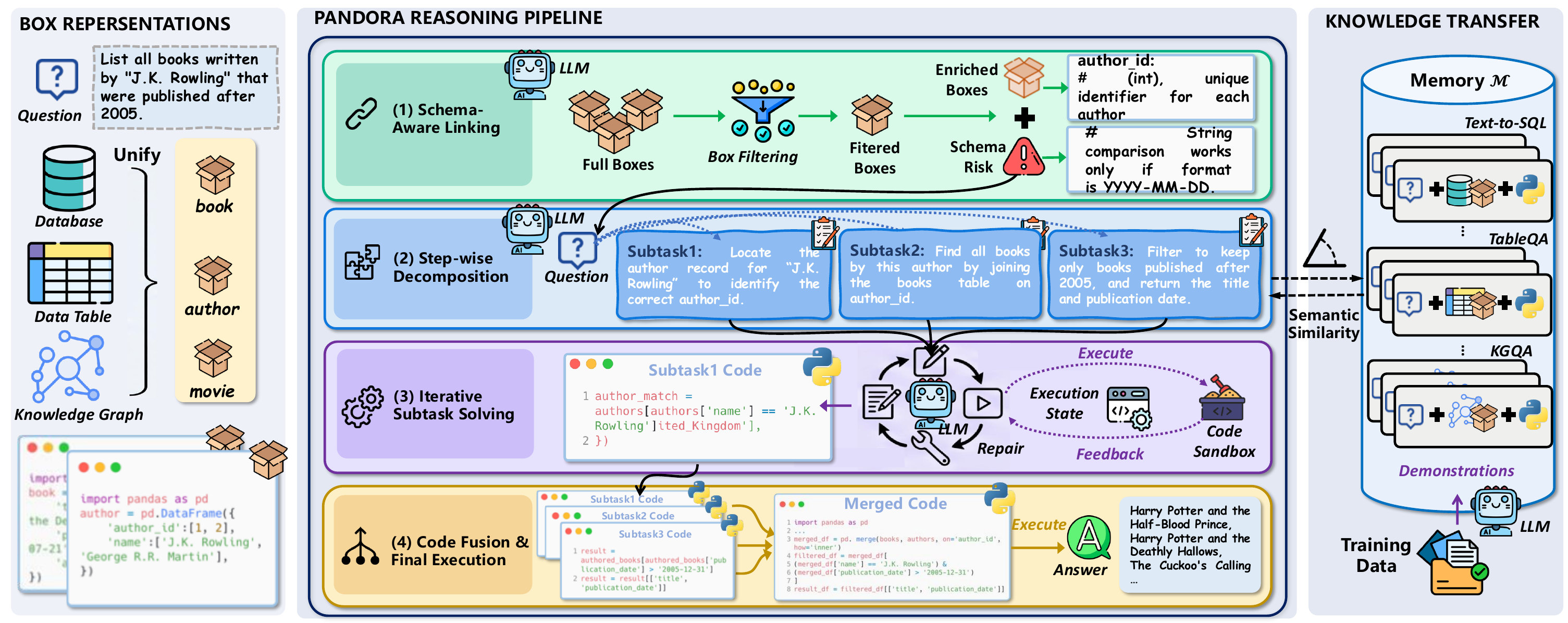}
	\caption{Architecture of our proposed \textsc{Pandora}. It takes an NLQ and converts structured data (tables, databases, KG subgraphs) into unified BOX representations. A multi-stage reasoning pipeline orchestrated by an LLM performs schema-aware linking, step-wise decomposition, iterative subtask execution with feedback loops, and code merging. A memory module enables semantic similarity-based knowledge transfer across heterogeneous reasoning tasks.} \label{fig:overview}
\end{figure*}

Figure~\ref{fig:overview} illustrates the overall architecture of \textsc{Pandora}. Structured knowledge $\mathcal{S}$ is first parsed into a collection of BOXes $\{\mathcal{B}_i\}_{i=1}^n$. \textsc{Pandora} then processes the input natural language question (NLQ) through a four-stage code-driven reasoning pipeline:
(a) \textit{Box-Aware Linking}, which identifies relevant BOXes and anticipates potential reasoning risks;
(b) \textit{Step-wise Decomposition}, which reformulates the NLQ into an ordered sequence of subtasks;
(c) \textit{Iterative Subtask Solving}, wherein code for each subtask is iteratively generated, executed, and repaired, translating semantic intent into executable \textsc{Pandas} operations (e.g., \texttt{.loc}, \texttt{.groupby}, \texttt{.merge}); and
(d) \textit{Code Merging \& Final Execution}, which consolidates all subtask code snippets and performs a full-data execution over $\{\mathcal{B}_i\}_{i=1}^n$ to produce the final answer $\mathcal{A}$.

\subsection{Box-Aware Linking}
\label{sec:schema_linking}

This stage identifies relevant fields and values across the set of BOXes for question $\mathcal{Q}$, narrowing the search space for subsequent reasoning.
Given the field set $\bigcup_i \Phi_i$ and value sets $\bigcup_i \Psi_i$ of all BOXes $\{\mathcal{B}_i\}_{i=1}^n$, the LLM $f_{\theta}$ produces:
\begin{equation}
    \mathcal{L} = f_{\theta}\!\left(\mathcal{X}_{\text{link}}
    \mid \mathcal{Q},\, \{\mathcal{B}_i\}_{i=1}^n
    \right),
\end{equation}
where $\mathcal{X}_{\text{link}}$ denotes the prompt template and $\mathcal{L}$ comprises:
\textbf{(i)}~\textit{BOX linking}, a subset of BOXes $\{\mathcal{B}_i\}_{i=1}^{n'}$ ($n' \le n$) relevant to $\mathcal{Q}$;
\textbf{(ii)}~\textit{value linking}, detect the potential values mentioned in $\mathcal{Q}$ (e.g., \texttt{2008} and \texttt{vince
mendoza});
\textbf{(iii)}~\textit{risk profiling}, schema-level ambiguities
(\emph{e.g.}, homonymous fields across BOXes, implicit join paths) propagated as
explicit risk signals to downstream stages.
In addition, to bridge the semantic gap between field names $\Phi$ and $\mathcal{Q}$, $f_{\theta}$ enriches each linked field $\phi \in \Phi$ with a
disambiguated semantic description $\mu(\phi)$, producing an optimized field map $\Phi = \{(\phi, \mu(\phi))\}_{i=1}^N$.
Finally, we denote the filtered and optimized BOXes as $\{\mathcal{B}_i^*\}_{i=1}^{n'}$.

\subsection{Step-wise Decomposition}
\label{sec:task_decomposition}

To fully exploit \textsc{Python}'s block-wise execution capabilities, \textsc{Pandora}
decomposes $\mathcal{Q}$ into a sequence of subtasks rather than
generating monolithic reasoning code in a single pass:
\begin{equation}
    T_{\text{decomp}} = f_{\theta}\!\left(\mathcal{X}_{\text{decomp}}
    \mid \mathcal{Q},\, \{\mathcal{B}_i^*\}_{i=1}^{n'},\,
    R_{\text{risk}}\right),
\end{equation}
where $\mathcal{X}_{\text{decomp}}$ denotes the prompt template and the decomposition
$T_{\text{decomp}} = [t_1, t_2, \dots, t_m]$ forms a logical chain in which each subtask $t_i$ consumes the intermediate results produced by its predecessors, enabling structured, stepwise resolution of complex queries. For example, the first subtask of the running question is that \textit{Filter the \texttt{producer} BOX to find all entries where the \texttt{producer} field matches \texttt{Vince Mendoza}.}

\subsection{Iterative Subtask Solving}
\label{sec:subtask_execution}

For each subtask $t_i \in \mathcal{T}_{\text{decomp}}$, \textsc{Pandora} follows an iterative generate–execute–validate cycle before proceeding to $t_{i+1}$, thereby preventing errors from propagating across subtasks. Concretely, given the accumulated code $\mathcal{C}_{<i}$ produced by all preceding subtasks, the model $f_{\theta}$ generates a code snippet $\mathcal{C}_i$ for the current subtask $t_i$:
\begin{equation}
    \mathcal{C}_i = f_{\theta}\!\left(\mathcal{X}_{\text{gen}} \mid t_i,\, \{\mathcal{B}_i^*\}_{i=1}^{n'},\, \mathcal{C}_{<i},\, R_{\text{risk}}\right),
\end{equation}
where $\mathcal{X}_{\text{gen}}$ denotes the prompt template. Notably, the original $\mathcal{Q}$ is intentionally omitted at this stage to avoid distracting the model with contextual information irrelevant to $t_i$.

\noindent\textbf{Execution Feedback.}
Once generated, $\mathcal{C}_i$ is executed within a sandboxed \textsc{Python} interpreter. If a runtime error occurs, the corresponding traceback is captured as feedback and supplied to the LLM, which then produces a revised code snippet:
\begin{equation}
    \mathcal{C}_i^{(r)} = f_{\theta}\!\left(\mathcal{X}_{\text{fix}} \mid t_i,\, \{\mathcal{B}_i^*\}_{i=1}^{n'},\, \mathcal{C}_i^{(r-1)},\, F_{\text{feedback}}\right),
\end{equation}
where $r \in \{1, \dots, L\}$ indexes the current repair iteration.

\noindent\textbf{Execution State Injection.}
Upon successful execution of each subtask' code snippet, all newly defined variables are inspected and their associated metadata, including shape, schema, data types, and sample rows, are captured, then serialized and injected into the context
$\mathcal{C}_{<i+1}$, allowing subsequent subtasks to condition on concrete intermediate evidence rather than relying purely on abstract symbolic representations.

\subsection{Code Merging \& Final Execution}
\label{sec:code_fusion}

After all subtasks are individually generated and validated, \textsc{Pandora} merges the accumulated code snippets into a single coherent program: $\mathcal{C}^* = \mathcal{C}_1^* \parallel \mathcal{C}_2^* \parallel \cdots \parallel \mathcal{C}_m^*,$
where $\parallel$ denotes the concatenation operation.
The merged program $\mathcal{C}^*$ is then executed over the complete BOX set. These BOXes are loaded by reading the stored \texttt{pickle} files. Should execution fail at this stage, the feedback mechanism is triggered, leveraging the error traceback to guide code repair. The final answer $\mathcal{A}$ is obtained by executing the validated program.

\subsection{Reasoning with Knowledge Transfer}
\label{sec:code_reasoning}

Across all stages of the pipeline described above, every LLM invocation follows a unified code-driven reasoning paradigm that combines \textit{Chain-of-Thought} (CoT) prompting and few-shot demonstration retrieval.
Concretely, at each stage of the pipeline, the LLM $f_{\theta}$ first generates a natural-language reasoning trace $\mathcal{Z}_{(s)}$ before producing the structured output $\mathcal{Y}_{(s)}$:
\begin{equation}
\begin{split}
    P\!\bigl(\mathcal{Z}_{(s)}, \mathcal{Y}_{(s)} \!\mid\! \mathcal{X}_{(s)}; \theta\bigr)
    =& \prod_{i=1}^{|\mathcal{Z}_{(s)}|} P\!\bigl(r_i \!\mid\! \mathcal{X}_{(s)}, r_{<i}; \theta\bigr) \\
    &\cdot \prod_{j=1}^{|\mathcal{Y}_{(s)}|} P\!\bigl(o_j \!\mid\! \mathcal{X}_{(s)}, \mathcal{Z}_{(s)}, o_{<j}; \theta\bigr),
\end{split}
\end{equation}
where $\mathcal{X}_{(s)}$ is the stage-specific prompt.

\noindent\textbf{Few-Shot Demonstration Injection.}
Every prompt $\mathcal{X}^{(s)}$ incorporates $K$ demonstrations retrieved from memory $\mathcal{M}$:
\begin{equation}
\begin{split}
    \mathcal{X}_{(s)} =\; &\texttt{Ins}_{(s)} \mid
    \bigl(\mathcal{Q}_1, \mathcal{B}_{1,1}, \dots, \mathcal{B}_{1,n_1}, \mathcal{R}_1, \mathcal{O}_1\bigr) \\
    &\mid \cdots \mid
    \bigl(\mathcal{Q}_K, \mathcal{B}_{K,1}, \dots, \mathcal{B}_{K,n_K}, \mathcal{R}_K, \mathcal{O}_K\bigr) \\
    &\mid \mathcal{Q}, \mathcal{B}_1, \dots, \mathcal{B}_n,
\end{split}
\end{equation}
where $\texttt{Ins}_{(s)}$ is the instruction for stage $s$. To control context length, only the empty BOX schemas $(b, \Phi)$ are included in demonstrations, while field values $\Psi$ are excluded.

\noindent \textbf{Knowledge Transfer.}
Unlike the conventional constraint of same-task demonstration retrieval, \textsc{Pandora} leverages the shared \textsc{Pandas} API foundation within the unified BOX representation to enable cross-task knowledge transfer.
Specifically, for each $\mathcal{Q}$, we select its $K$ demonstrations $(\mathcal{Q}_1, \dots, \mathcal{O}_k)$ from memory $\mathcal{M}$ by maximizing semantic similarity, making the retrieval process agnostic to the source task of $\mathcal{Q}_k$. The similarity is calculated as $s(\mathcal{Q}_k, \mathcal{Q}) = \cos(g_{\theta}(\mathcal{Q}_k), g_{\theta}(\mathcal{Q}))$, where $g_\theta$ is an embedding LLM.

\subsection{Progressive Memory Construction}
\label{sec:learning}
In this section, we explain how memory is constructed. Instead of exhaustively labeling samples from all tasks concurrently, we adopt a progressive annotation strategy that gradually advances from simpler to more complex instances. The entire process is structured into two stages.

\noindent \textbf{Memory Initialization}
Pandora annotates only the training samples of the DB SKR task, as these samples~\cite{DBLP:conf/emnlp/YuZYYWLMLYRZR18} typically contain high-quality, human-written SQL labels. Compared to the concise answers found in Table SKR or the SPARQL labels used in KG SKR, SQL labels provide more detailed reasoning steps and are LLM-friendly. Furthermore, the reasoning logic required by these datasets closely aligns with that of \textsc{Pandas}, thereby reducing annotation complexity when initializing the demonstration set in low-resource settings.
To enhance the quality of the generated \textsc{Python} Code, the candidate answer $\mathcal{A}^*$, obtained by executing the generated code, is compared against the ground-truth answer $\mathcal{A}$.
If $\mathcal{A}^*$ matches $\mathcal{A}$, the corresponding pair $(\mathcal{R}, \mathcal{C})$ is treated as correct and retained. Otherwise, the outcome of the comparison is further leveraged as feedback to $f_{\theta}$, facilitating iterative self-correction. Finally, samples that pass the verification check are collected to construct the initial memory buffer $\mathcal{M}_0$.

\noindent \textbf{Multi-Task Adaptation}
In the second stage, we utilize the annotated DB SKR examples from $\mathcal{M}_0$ as demonstrations, prompting $f_{\theta}$ to generate code annotations for the KG SKR and Table SKR training sets.
This approach facilitates further knowledge transfer from DB SKR to the other two tasks. Consistent with the first stage, we use the gold-standard answers to assess and refine the correctness of each generated code $\mathcal{C}$.
The successfully annotated instances from all three tasks are then aggregated to form the final memory set $\mathcal{M}$.
Notably, rather than exhaustively annotating the entire training corpus, we find that a relatively small subset of high-quality, validated examples is sufficient to obtain competitive performance.

\noindent \textbf{Memory construction details.}
For each benchmark, we first sample a small candidate subset from the training split and ask the backbone LLM to convert the available supervision into executable Pandas demonstrations. For DB SKR, the supervision is the gold SQL query; for Table SKR and KG SKR, the supervision is the gold answer set and the generated code is validated by execution against the official answer. A candidate demonstration is inserted into memory only if the produced code executes successfully and its result matches the gold answer under the corresponding task metric. The final memory item stores the natural language question, the empty BOX schema, the generated reasoning trace, the executable Pandas code, and verification metadata, while full cell values are excluded from demonstrations to control context length. As summarized in Figure~\ref{fig:task}, the resulting memory contains 600 verified examples for Spider, WTQ, WikiSQL, and GrailQA, 750 examples for BIRD, and 457 examples for WebQSP due to the smaller available training split. At inference time, retrieval is task-agnostic: \texttt{bge-large-en-v1.5} embeds the current question and all memory questions, and the top-$K$ examples are selected by cosine similarity regardless of whether they originate from DB, Table, or KG tasks.

\section{Discussion on Cross-Source Reasoning}
\label{sec:cross_source_reasoning}

A natural consequence of our proposed BOX representation is that \textsc{Pandora} can seamlessly reason over multiple heterogeneous knowledge sources within a single query.  \textsc{Pandora} then represents all retrieved sources as BOXes with a consistent field-value structure, enabling cross-source joins via standard \texttt{pd.merge} operations without any task-specific plumbing.
To bridge heterogeneous knowledge sources, \textsc{Pandora} relies on shared entity surface forms across BOXes as natural join keys. Specifically, entity alignment is performed by normalizing surface forms and linking co-referent entities across sources through exact or fuzzy string matching, thereby establishing implicit foreign-key relationships between BOXes. Rather than relying on predefined schema mappings, \textsc{Pandora} then delegates the inference of cross-source join keys entirely to $f_{\theta}$, which identifies semantically equivalent fields across the aligned BOXes. Once such cross-source links are established, \textsc{Pandora} can faithfully replicate the full reasoning process by executing $\mathcal{C}$ over the fused BOXes, completing multi-source inference in a unified and reproducible manner.

\noindent \textbf{Illustrative Example.}
Consider the query: \textit{List all movies directed by directors born in London.}, requiring information from two heterogeneous sources: a movie database and a KG. The corresponding BOX representations are:

\begin{tcolorbox}[colback=white, colframe=black, arc=2mm, boxrule=0.5pt]
\scriptsize
\begin{lstlisting}[style=arxivcode,language=python]
# BOX from DB: movies and directors
movies = pd.DataFrame({
    "title": [],     # (str), movie title
    "director_id": [] # (int), FK to directors
})
directors = pd.DataFrame({
    "director_id": [],  # (int), primary key
    "name": []           # (str), director name
})

# BOX from KG: person birthplace information
person = pd.DataFrame({
    "person_name": [],  # (str), entity name
    "birthplace": [],   # (str), city of birth
    "nationality": []   # (str), country
})
\end{lstlisting}
\end{tcolorbox}

\noindent KG and DB can be linked using names as foreign keys. The LLM generates the following cross-source reasoning code:
\begin{tcolorbox}[colback=white, colframe=black, arc=2mm, boxrule=0.5pt]
\scriptsize
\begin{lstlisting}[style=arxivcode,language=python]
# Step 1: Find directors born in London (from KG BOX)
london_directors = person[
    person['birthplace'] == 'London'
]['person_name']

# Step 2: Match KG names with DB director names (cross-source join)
matched = directors[directors['name'].isin(london_directors)]

# Step 3: Join to get movies from DB
result = pd.merge(movies, matched, on='director_id')['title']
\end{lstlisting}
\end{tcolorbox}

In summary, while our main experiments follow community protocols and evaluate on standard single-source benchmarks, the BOX representation provides a principled and extensible foundation for cross-source reasoning, naturally supporting entity alignment and heterogeneous knowledge fusion without requiring predefined schema mappings.

\section{Experiments}
%
\subsection{Datasets}
We evaluate all methods across SKR tasks on DB, Table, and KG, respectively: For \textit{DB SKR},  we use \textbf{1) Spider}~\cite{DBLP:conf/emnlp/YuZYYWLMLYRZR18}, \textbf{2) Spider-SYN} (S-SYN)~\cite{DBLP:conf/acl/GanCHPWXH20}, and \textbf{3) Bird-SQL} (Bird)~\cite{li2023bird}, which introducing challenges that demand multi-step reasoning and DB schema understanding for precise SQL synthesis;
For \textit{Table SKR}, we use \textbf{4) WTQ}~\cite{DBLP:conf/acl/PasupatL15} and \textbf{5) WikiSQL}~\cite{DBLP:journals/corr/abs-1709-00103} requiring operations such as filtering, comparison, and aggregation on real-world tables;
For \textit{KG SKR}, we use \textbf{6) GrailQA}~\cite{DBLP:conf/www/GuKVSLY021} and \textbf{7) WebQSP}~\cite{DBLP:conf/acl/YihRMCS16} focusing on multi-hop reasoning over the Freebase KG with complex S-expressions and SPARQL queries.
Following existing work~\cite{DBLP:conf/emnlp/ScholakSB21,DBLP:conf/nips/PourrezaR23,DBLP:journals/pacmmod/LiZLFZZWP0024}, we report results on the development sets for Spider and BIRD, and on the test sets for the remaining datasets.

\begin{figure}
\centering
\begin{tikzpicture}[scale=0.75]
    \begin{axis}[
        ybar,
        width=12cm, 
        height=4cm,
        bar width=8pt,
        ylabel={\# of Examples},
        symbolic x coords={Spider, Bird, WTQ, WikiSQL, GrailQA, WebQSP},
        xtick=data,
        enlarge x limits=0.2,
        ymode=log,
        log origin=infty,
        legend style={at={(0.28,1.0)}, anchor=north, legend columns=-1},
        legend entries={Memory, Train, Test}
    ]
        \addplot coordinates {(Spider, 600) (Bird, 750) (WTQ, 600) (WikiSQL, 600) (GrailQA, 600) (WebQSP, 457)};
        \addplot coordinates {(Spider, 7000) (Bird, 9428) (WTQ, 11321) (WikiSQL, 56355) (GrailQA, 44337) (WebQSP, 3044)};
        \addplot coordinates {(Spider, 1034) (Bird, 1534) (WTQ, 4344) (WikiSQL, 15878) (GrailQA, 6463) (WebQSP, 1616)};
    \end{axis}
\end{tikzpicture}
\caption{Example numbers (Logarithmic y-axis) of our memory $\mathcal{M}$.}
\label{fig:task}
\end{figure}

\subsection{Evaluation Metrics}
Consistent with previous USKR studies~\cite{DBLP:conf/emnlp/JiangZDYZW23,DBLP:journals/corr/abs-2406-18916}, we employ widely-adopted metrics for fair and direct comparison across benchmarks. Specifically, we report Execution Accuracy (EX) for the Spider and S-SYN datasets, Denotation Accuracy (DA) for WTQ and WikiSQL, and Hit@1 for GrailQA and WebQSP. To provide a more comprehensive evaluation, we further include the F1-score between predicted and ground-truth answer sets in our ablation studies.


\subsection{Implementation Details}
\label{sec:implementation}
 \textsc{Pandora} is designed to be backbone-agnostic, supporting both proprietary and open-weight LLMs.
We evaluate four distinct models across two families:
\textbf{(a) Proprietary models:} OpenAI's \texttt{gpt-3.5-turbo} and \texttt{gpt-4o-mini}, selected for their strong instruction-following and code generation capabilities.
\textbf{(b) Open-weight models:} Alibaba's \texttt{qwen3-8B} and \texttt{qwen3-coder-flash} (i.e., \texttt{qwen3-coder-30B-A3B-Instruct}).
This selection spans different model scales and specializations, enabling us to assess \textsc{Pandora}'s robustness across diverse backbones.
All models operate with temperature $0.0$ for deterministic outputs.
For the demonstration retriever $g_{\theta}$, we employ \texttt{bge-large-en-v1.5}, an encoder-only model optimized for semantic similarity. All experiments are conducted on an Intel\textregistered~Xeon\textregistered~Gold 6133 CPU @2.50GHz with 512 GB RAM and an NVIDIA A100 GPU.

The number of demonstrations ($K$) is fixed at 10. For subgraph retrieval, the maximum hops ($H$) is set to 3 for GrailQA and 2 for WebQSP, reflecting their difference in complexity. Each NLQ undergoes at most $L=3$ rounds of execution feedback, with 50 samples drawn per iteration before final execution. To simulate a data-scarce scenario, we extract only about 5\% of the training data from each dataset to build the memory. The number of examples in constructed memory are summarized in Figure~\ref{fig:task}.

 \subsection{KG Subgraph BOX Size Analysis}
\label{sec:kg_box_size}

\begin{figure}[t]
\centering
	\includegraphics[width=\columnwidth]{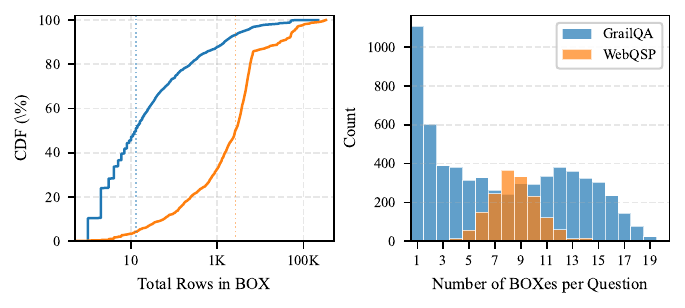}
	\caption Distribution of KG subgraph BOX sizes. \textbf{Left:} Cumulative distribution function (CDF) of total rows per BOX, where CDF($x$) denotes the fraction of questions whose BOX contains at most $x$ rows. \textbf{Right:} Histogram of the number of BOXes per question (each CSV table = one BOX). \label{fig:kg_box_size}
\end{figure}

To assess the memory footprint of the KG-to-BOX conversion (Algorithm~3), we systematically measured the BOX sizes for all questions in the GrailQA test set (6,409 questions) and the WebQSP test set (1,612 questions). As shown in Figure~\ref{fig:kg_box_size}, the vast majority of extracted subgraphs are compact. The left panel plots the \textit{cumulative distribution function} (CDF) of total rows per BOX, where CDF($x$) represents the fraction of questions whose BOX contains at most $x$ rows, revealing that GrailQA's median is only 13 rows, with 97.8\% of questions under 15K rows; WebQSP's median is 2,631 rows, with 90\% under 29K rows. The right panel shows the distribution of BOX counts per question, with GrailQA averaging 7.5 BOXes and WebQSP averaging 8.5 BOXes per question. The right-skewed row distributions are driven by a small fraction of questions connecting to large entity tables (e.g., \texttt{people.person} with up to 305K rows in WebQSP), but even the largest BOX remains well within memory capacity, the largest GrailQA BOX (217K rows, 16 BOXes) occupies approximately 45 MB in RAM, far below typical workstation limits. This empirical analysis confirms that the $H$-hop subgraph extraction strategy effectively bounds BOX sizes, and the KG-to-BOX conversion does not introduce prohibitive memory overhead in practice.

\subsection{Compared Methods}
We compared our method with:
a) \textit{Competitive Single-type SKR Methods}.
For DB SKR, we select PICARD~\cite{DBLP:conf/emnlp/ScholakSB21}, DIN-SQL~\cite{DBLP:conf/nips/PourrezaR23}, NatSQL~\cite{DBLP:conf/aaai/Li00023}, CodeS~\cite{DBLP:journals/pacmmod/LiZLFZZWP0024},
DTS-SQL~\cite{DBLP:conf/emnlp/PourrezaR24},
SQLFixAgent~\cite{DBLP:conf/aaai/CenLLW25},
DAIL~\cite{DBLP:journals/pvldb/GaoWLSQDZ24},
Mac-SQL~\cite{DBLP:conf/coling/WangR0LBCYZYSL25}, and
GenSQL~\cite{DBLP:conf/coling/ShiXLXC0WW25};
For Table SKR, we select TAPAS~\cite{DBLP:conf/acl/HerzigNMPE20}, TAPEX~\cite{DBLP:conf/iclr/LiuCGZLCL22}, Binder~\cite{DBLP:conf/iclr/ChengX0LNHXROZS23}, DATER~\cite{DBLP:conf/sigir/YeHYLHL23}, and
TIDE~\cite{DBLP:conf/iclr/YangDZDCDZ25};
For KG SKR, we select RnG-KBQA~\cite{DBLP:conf/acl/YeYHZX22}, TIARA~\cite{DBLP:conf/emnlp/ShuYLKMQL22}, DecAF~\cite{DBLP:conf/iclr/YuZNZL0HWWX23}, KB-Binder~\cite{DBLP:conf/acl/LiMZGSC23}, KB-Coder~\cite{DBLP:conf/aaai/NieZW024}, KBQA-o1~\cite{DBLP:journals/corr/abs-2501-18922},
and SG-KBQA~\cite{DBLP:conf/emnlp/GaoLQ25};
b) \textit{Unified SKR Methods}.
We select UnifiedSKG~\cite{DBLP:conf/emnlp/XieW0ZSYWZYWZWL22}, StructLM~\cite{DBLP:journals/corr/abs-2402-16671}, StructGPT~\cite{DBLP:conf/emnlp/JiangZDYZW23},  Readi~\cite{DBLP:conf/acl/ChengZXYZQHCL0R24}, TrustUQA~\cite{DBLP:journals/corr/abs-2406-18916}.
In Table~\ref{tab:baselines_db}, \ref{tab:baselines_table}, and \ref{tab:baselines_kg}, the superscript $\spadesuit$ denotes models fine-tuned on the target dataset.
See \ref{sec:source_of_results} for details on the source of the results.

\subsection{Overall result of DB SRK Tasks}

 \textsc{Pandora} demonstrates highly competitive performance on the DB SRK task (Table~\ref{tab:baselines_db}). On BIRD, the more challenging database semantic ranking benchmark that emphasizes real-world complexity and domain generalization, \textsc{Pandora} with \texttt{gpt-4o-mini} achieves 57.4\% execution accuracy, competing favorably with leading single-type methods. Notably, \textsc{Pandora} with the open-weight model \texttt{qwen3-coder-flash} (30B) and \texttt{qwen3-8B} achieves 60.6\% and 58.6\% on BIRD, respectively, substantially surpassing the proprietary \texttt{gpt-4} baseline of Mac-SQL (59.4\%) by 1.2\%. The result demonstrates that \textsc{Pandora}'s code-driven reasoning framework achieves superior performance even with smaller open-weight models, outperforming expensive proprietary approaches while maintaining computational efficiency.
\begin{table}
\centering
\caption{Results (\%) of the DB SKR task.}
\label{tab:baselines_db}
\scalebox{0.95}{
\begin{tabular}{lccc}
\toprule
\multicolumn{1}{l}{\multirow{2}[1]{*}{\textbf{Method}}}
        &\textbf{Bird} &\textbf{Spider} &\textbf{S-SYN} \\
        \cmidrule(lr){2-4}
        & EX & EX & EX  \\
        \cmidrule(lr){1-4}
\multicolumn{2}{l}{\textit{Single-type SKR Methods}} \\
\cmidrule(lr){1-4}
        PICARD$^\spadesuit$~\cite{DBLP:conf/emnlp/ScholakSB21} + \texttt{t5-3B} &- & 79.3 & 69.8  \\
        DIN-SQL~\cite{DBLP:conf/nips/PourrezaR23} + \texttt{gpt-4} &50.7 & 74.2 & -  \\
        DAIL~\cite{DBLP:journals/pvldb/GaoWLSQDZ24} + \texttt{gpt-4} &54.8 & 78.1 & - \\
        CodeS-7B$^\spadesuit$~\cite{DBLP:journals/pacmmod/LiZLFZZWP0024} &57.2 & 85.4 & 77.0 \\
        DTS-SQL$^\spadesuit$~\cite{DBLP:conf/emnlp/PourrezaR24} + \texttt{deepseek-7B}  &55.8 & 85.5 & 76.2 \\
        SQLFixAgent~\cite{DBLP:conf/aaai/CenLLW25} +  CodeS-7B$^\spadesuit$&60.2 & 86.2 & \textbf{77.7} \\
         E-SQL~\cite{DBLP:journals/corr/abs-2409-16751} +  \texttt{gpt-4o-mini}$^\spadesuit$&61.6 & - & - \\
        Mac-SQL~\cite{DBLP:conf/coling/WangR0LBCYZYSL25}  + \texttt{gpt-4} & 59.4 & \textbf{86.8} & - \\
        GenSQL~\cite{DBLP:conf/coling/ShiXLXC0WW25} + \texttt{gpt-4} & 59.8 & 85.6 & - \\
\cmidrule(lr){1-4}
\multicolumn{2}{l}{\textit{Unified SKR Methods}} \\
\cmidrule(lr){1-4}
        UnifiedSKG-3B$^\spadesuit$~\cite{DBLP:conf/emnlp/XieW0ZSYWZYWZWL22} &- & 71.8 & -  \\
        StructLM-7B$^\spadesuit$~\cite{DBLP:journals/corr/abs-2402-16671} &- & 79.6 & -  \\
        StructGPT~\cite{DBLP:conf/emnlp/JiangZDYZW23} + \texttt{gpt-3.5-turbo} &- & 77.8 & 64.0  \\
\cmidrule(lr){1-4}
    \textsc{Pandora} + \texttt{gpt-3.5-turbo} &45.6 & \textbf{81.7} & 74.7 \\
    \textsc{Pandora} + \texttt{gpt-4o-mini} &57.4 & 81.3 & \textbf{78.0}  \\
     \textsc{Pandora} + \texttt{qwen3-8B} &58.6 & 80.3 & 77.4  \\
    \textsc{Pandora} + \texttt{qwen3-coder-flash} &\textbf{60.6} & 79.2 & 78.1  \\
\bottomrule
\end{tabular}
}
\end{table}
On Spider, \textsc{Pandora} with \texttt{gpt-3.5-turbo} achieves 81.7\% execution accuracy, representing a significant advancement over the best existing unified method, StructLM-7B (79.6\%), with a gain of 2.1\%.  On S-SYN, \textsc{Pandora}'s effectiveness extends to diverse syntactic structures, where it achieves 78.0\% EX with \texttt{gpt-4o-mini}, surpassing the leading single-type method by 0.3\%.

\subsection{Overall result of Table SRK Tasks}
Table~\ref{tab:baselines_table} presents the performance of \textsc{Pandora} on the Table SRK task across WTQ and WikiSQL benchmarks. On WTQ, which contains open-domain Wikipedia tables and requires more diverse operations such as implicit row selection, comparison, superlative reasoning, and aggregation, \textsc{Pandora} achieves strong performance across all backbones. In particular, \textsc{Pandora} with \texttt{qwen3-coder-flash} obtains 70.3\% DA, outperforming the strongest unified baseline Readi (61.7\%) by 8.6 points and exceeding the fine-tuned TAPEX model (57.5\%) by 12.8 points. The remaining backbones also stay in a narrow range of 68.2\%--69.1\%, indicating that the BOX representation and code-driven reasoning pipeline are robust on complex table reasoning. At the same time, \textsc{Pandora} still trails the specialized TIDE result (75.0\%), suggesting that WTQ-style semantic ambiguity and flexible answer normalization remain challenging boundaries for a unified framework.
\begin{table}
\centering
\caption{Results (\%) of the Table SKR task.}
\label{tab:baselines_table}
\scalebox{0.95}{
\begin{tabular}{lcc}
\toprule
\multicolumn{1}{l}{\multirow{2}[1]{*}{\textbf{Method}}}
        &\textbf{WTQ} &\textbf{WikiSQL} \\
        \cmidrule(lr){2-3}
        & DA & DA  \\
        \cmidrule(lr){1-3}
\multicolumn{2}{l}{\textit{Single-type SKR Methods}} \\
\cmidrule(lr){1-3}
        TAPAS-0.34B$^\spadesuit$~\cite{DBLP:conf/acl/HerzigNMPE20} &  48.8 & 83.6  \\
        TAPEX-0.4B$^\spadesuit$~\cite{DBLP:conf/iclr/LiuCGZLCL22} &  57.5 & \textbf{89.5}  \\
        Binder~\cite{DBLP:conf/iclr/ChengX0LNHXROZS23} + \texttt{gpt-3-codex} & 64.6 & -  \\
        DATER$^\spadesuit$~\cite{DBLP:conf/sigir/YeHYLHL23} + \texttt{gpt-3-codex} & 65.9 & - \\
        TIDE~\cite{DBLP:conf/iclr/YangDZDCDZ25} + \texttt{gpt-3.5} & \textbf{75.0} & - \\
\cmidrule(lr){1-3}
\multicolumn{2}{l}{\textit{Unified SKR Methods}} \\
\cmidrule(lr){1-3}
        UnifiedSKG-3B$^\spadesuit$~\cite{DBLP:conf/emnlp/XieW0ZSYWZYWZWL22} & 49.3 & 86.0  \\
        StructLM-7B$^\spadesuit$~\cite{DBLP:journals/corr/abs-2402-16671} & 50.1 & \textbf{88.7}  \\
        StructGPT~\cite{DBLP:conf/emnlp/JiangZDYZW23} + \texttt{gpt-3.5-turbo} & 52.2 & 65.6  \\
        Readi~\cite{DBLP:conf/acl/ChengZXYZQHCL0R24} + \texttt{gpt-3.5-turbo} & \textbf{61.7} & 66.2  \\
        TrustUQA~\cite{DBLP:journals/corr/abs-2406-18916} + \texttt{gpt-3.5-turbo} & 44.2 & 85.9 \\
\cmidrule(lr){1-3}
    \textsc{Pandora} + \texttt{gpt-3.5-turbo} & 68.2 & 87.4 \\
    \textsc{Pandora} + \texttt{gpt-4o-mini} & \textbf{68.9} & 86.3 \\
    \textsc{Pandora} + \texttt{qwen3-8B} &69.1 & 84.3 \\
    \textsc{Pandora} + \texttt{qwen3-coder-flash} &\textbf{70.3} & 87.6 \\
\bottomrule
\end{tabular}
}
\end{table}
On WikiSQL, where most questions involve relatively clean single-table selection and aggregation, \textsc{Pandora} also performs competitively with task-specific and fine-tuned methods. \textsc{Pandora} with \texttt{qwen3-coder-flash} achieves 87.6\% DA, approaching TAPEX (89.5\%) and StructLM-7B (88.7\%) while requiring no task-specific fine-tuning on WikiSQL. The \texttt{gpt-3.5-turbo} variant reaches 87.4\%, and all \textsc{Pandora} variants substantially outperform prompt-based unified baselines such as StructGPT (65.6\%) and Readi (66.2\%). Compared with WTQ, the smaller gap to specialized models on WikiSQL indicates that \textsc{Pandora}'s Pandas-based execution is especially effective when the table structure is clean and the required operations can be directly mapped to filtering, grouping, and aggregation primitives.

\subsection{Overall result of KG SRK Tasks}
\begin{table}
\centering
\caption{Results (\%) of the KG SKR task.}
\label{tab:baselines_kg}
\scalebox{0.95}{
\begin{tabular}{lccc}
    \toprule
    \multirow{2}{*}{\textbf{Method}} &
    \textbf{GrailQA} &
    \multicolumn{2}{c}{\textbf{WebQSP}} \\
    \cmidrule(lr){2-4}
    & F1 & Hit@1 & F1 \\
    \cmidrule(lr){1-4}
    \multicolumn{4}{l}{\textit{Single-type SKR Methods}} \\
    \cmidrule(lr){1-4}
    RnG-KBQA$^\spadesuit$~\cite{DBLP:conf/acl/YeYHZX22} + \texttt{t5-base} & 76.9 & -      & -      \\
    DecAF$^\spadesuit$~\cite{DBLP:conf/iclr/YuZNZL0HWWX23} + \texttt{fid-3B} & 81.4 & -      & \textbf{78.7} \\
    TIARA$^\spadesuit$~\cite{DBLP:conf/emnlp/ShuYLKMQL22}  + \texttt{t5-base} & 81.9 & -      & 76.7   \\
    KB-Binder~\cite{DBLP:conf/acl/LiMZGSC23} + \texttt{code-davinci-002}              & 51.7 & -      & 68.9   \\
    KB-Coder~\cite{DBLP:conf/aaai/NieZW024} + \texttt{gpt-3.5-turbo}               & 61.3 & -      & 72.2   \\
    KBQA-o1~\cite{DBLP:journals/corr/abs-2501-18922} + \texttt{qwen2.5-72B}       & 82.1 & -      & 66.5   \\
    SG-KBQA$^\spadesuit$~\cite{DBLP:conf/emnlp/GaoLQ25} + \texttt{llama3.1-8B} &
    \textbf{84.4} & - &
    \textbf{80.3} \\
    \cmidrule(lr){1-4}
    \multicolumn{4}{l}{\textit{Unified SKR Methods}} \\
    \cmidrule(lr){1-4}
    UnifiedSKG-3B$^\spadesuit$~\cite{DBLP:conf/emnlp/XieW0ZSYWZYWZWL22} & - & - & 80.7 \\
    StructGPT~\cite{DBLP:conf/emnlp/JiangZDYZW23} + \texttt{gpt-3.5-turbo}                 & - & 69.6 & - \\
    Readi~\cite{DBLP:conf/acl/ChengZXYZQHCL0R24} + \texttt{gpt-3.5-turbo}                     & - & 74.3 & - \\
    TrustUQA~\cite{DBLP:journals/corr/abs-2406-18916} + \texttt{gpt-3.5-turbo}                & - & 83.5 & - \\
    \cmidrule(lr){1-4}
    \textsc{Pandora} + \texttt{gpt-3.5-turbo}     & 83.0 & 83.7 & 74.5 \\
    \textsc{Pandora} + \texttt{gpt-4o-mini}       & \textbf{84.6} & \textbf{84.1} & \textbf{78.9} \\
    \textsc{Pandora} + \texttt{qwen3-8B} &79.3 & 82.3 & 77.3  \\
    \textsc{Pandora} + \texttt{qwen3-coder-flash} &84.2 & 84.6 & 79.5  \\
    \bottomrule
  \end{tabular}
}
\end{table}

Table~\ref{tab:baselines_kg} reports \textsc{Pandora}'s performance on KG SRK tasks across GrailQA and WebQSP benchmarks.
On GrailQA, \textsc{Pandora} with \texttt{gpt-4o-mini} achieves an F1 score of 84.6\%, matching the leading single-type method SG-KBQA (84.4\%) and substantially outperforming all unified competitors. \texttt{qwen3-coder-flash} achieves 84.2\%, demonstrating that strong performance is attainable with open-weight models. In contrast, \texttt{qwen3-8B} achieves 79.3\% and \texttt{gpt-3.5-turbo} achieves 83.0\%, indicating consistent performance across model scales while all variants substantially exceed other unified methods.
On WebQSP, \textsc{Pandora} with \texttt{qwen3-coder-flash} achieves the highest F1 score of 79.5\%, followed by \texttt{gpt-4o-mini} (78.9\%), which approaches the best single-type method SG-KBQA (80.3\%) within 1.4\%. For Hit@1 metric, \textsc{Pandora} with \texttt{qwen3-coder-flash} achieves 84.6\%, outperforming the best unified baseline TrustUQA and competing favorably with single-type approaches. \texttt{gpt-4o-mini} achieves Hit@1 of 84.1\%, while \texttt{qwen3-8B} achieves 82.3\%, all substantially exceeding other unified methods.

 \subsection{Ablation Study}
\begin{table*}
{\caption{Ablation results (\%). Evaluation was conducted on approximately 200 randomly sampled questions from each dataset.}\label{tab:ablation_study}}
	\begin{center}
	\scalebox{0.9}{
		\begin{tabular}{lccccccccccccccc}
			\toprule
			\multicolumn{1}{l}{\multirow{3}[1]{*}{\textbf{Ablation Setting}}}
			&\multicolumn{6}{c}{\textbf{\quad\quad\quad DB SKR \quad\quad\quad }}
            &\multicolumn{4}{c}{\textbf{\quad Table SKR \quad}}
            &\multicolumn{4}{c}{\textbf{KG SKR}}
            & \multirow{3}[1]{*}{\textbf{Avg. F1}}\\
		    \cmidrule(lr){2-7} \cmidrule(lr){8-11} \cmidrule(lr){12-15}
			& \multicolumn{2}{c}{\textbf{BIRD}}
            & \multicolumn{2}{c}{\textbf{Spider}}
            & \multicolumn{2}{c}{\textbf{S-SYN}}
            & \multicolumn{2}{c}{\textbf{WTQ}}
            & \multicolumn{2}{c}{\textbf{WikiSQL}}
            & \multicolumn{2}{c}{\textbf{GrailQA}}
            & \multicolumn{2}{c}{\textbf{WebQSP}} \\
		    \cmidrule(lr){2-7} \cmidrule(lr){8-11} \cmidrule(lr){12-15} \cmidrule(lr){16-16}
            & EX & F1
            & EX & F1
            & EX & F1
            & DA & F1
            & DA & F1
            & F1 & Hit@1
            & F1 & Hit@1 \\
            \cmidrule(lr){1-1} \cmidrule(lr){2-7} \cmidrule(lr){8-11} \cmidrule(lr){12-15} \cmidrule(lr){16-16}
			\textsc{Pandora} + \texttt{gpt-4o-mini} & $\mathbf{56.8}$ & $\mathbf{59.0}$ & $\mathbf{80.5}$ & $\mathbf{83.2}$ & $\mathbf{77.2}$ & $\mathbf{79.4}$ & $\mathbf{67.8}$ & $\mathbf{69.0}$ & $\mathbf{85.5}$ & $\mathbf{86.0}$ & $\mathbf{83.8}$ & $\mathbf{84.5}$ & $\mathbf{77.8}$ & $\mathbf{83.2}$ & $\mathbf{76.9}$ \\

            \cmidrule(lr){1-1} \cmidrule(lr){2-7} \cmidrule(lr){8-11} \cmidrule(lr){12-15} \cmidrule(lr){16-16}
            \quad $-$ Execution Feedback & 50.0 & 51.8 & 73.0 & 75.2 & 70.2 & 72.0 & 61.8 & 62.8 & 79.8 & 80.3 & 76.8 & 78.8 & 71.2 & 77.5 & 70.0 \\
            \quad $-$ Code Merging & 55.2 & 57.2 & 78.8 & 81.5 & 75.5 & 77.8 & 66.0 & 67.2 & 84.0 & 84.5 & 82.2 & 83.0 & 76.2 & 81.8 & 75.2 \\
            \quad $-$ Knowledge Transfer & 54.2 & 56.0 & 77.8 & 80.5 & 74.5 & 76.5 & 65.0 & 66.0 & 82.8 & 83.5 & 81.0 & 81.8 & 75.2 & 80.5 & 74.1 \\
            \quad $-$ Step-wise Decomposition & 50.5 & 52.2 & 74.0 & 76.5 & 70.5 & 72.5 & 61.5 & 62.5 & 79.2 & 79.8 & 77.2 & 79.0 & 71.5 & 77.5 & 70.3 \\
            \cmidrule(lr){1-1} \cmidrule(lr){2-7} \cmidrule(lr){8-11} \cmidrule(lr){12-15} \cmidrule(lr){16-16}
            Vanilla Prompt & 48.5 & 50.0 & 74.0 & 76.2 & 68.5 & 70.5 & 60.2 & 61.5 & 83.0 & 83.5 & 73.5 & 75.5 & 69.8 & 75.5 & 69.3 \\
            \cmidrule(lr){1-16}
            \textsc{Pandora} + \texttt{qwen3-8B} & $\mathbf{57.8}$ & $\mathbf{59.6}$ & $\mathbf{79.5}$ & $\mathbf{82.0}$ & $\mathbf{76.5}$ & $\mathbf{78.5}$ & $\mathbf{68.5}$ & $\mathbf{69.8}$ & $\mathbf{81.5}$ & $\mathbf{82.8}$ & $\mathbf{78.5}$ & $\mathbf{81.0}$ & $\mathbf{75.8}$ & $\mathbf{80.8}$ & $\mathbf{75.3}$ \\

            \cmidrule(lr){1-1} \cmidrule(lr){2-7} \cmidrule(lr){8-11} \cmidrule(lr){12-15} \cmidrule(lr){16-16}
            \quad $-$ Execution Feedback & 51.2 & 53.0 & 72.5 & 74.8 & 69.8 & 71.5 & 62.5 & 63.5 & 76.0 & 77.2 & 72.5 & 75.5 & 69.8 & 75.5 & 68.9 \\
            \quad $-$ Code Merging & 56.0 & 58.0 & 77.8 & 80.2 & 74.8 & 76.8 & 66.8 & 68.0 & 79.8 & 81.0 & 76.8 & 79.2 & 74.2 & 79.2 & 73.6 \\
            \quad $-$ Knowledge Transfer & 55.0 & 56.8 & 77.0 & 79.5 & 73.8 & 75.5 & 65.8 & 67.0 & 78.8 & 80.0 & 76.0 & 78.5 & 73.5 & 78.2 & 72.6 \\
            \quad $-$ Step-wise Decomposition & 50.8 & 52.5 & 73.0 & 75.5 & 69.5 & 71.5 & 62.0 & 63.0 & 75.5 & 76.5 & 72.0 & 74.8 & 69.2 & 75.0 & 68.6 \\
            \cmidrule(lr){1-1} \cmidrule(lr){2-7} \cmidrule(lr){8-11} \cmidrule(lr){12-15} \cmidrule(lr){16-16}
            Vanilla Prompt & 45.5 & 47.0 & 70.5 & 72.5 & 65.0 & 66.8 & 56.5 & 57.8 & 80.0 & 80.5 & 70.5 & 72.8 & 66.5 & 72.0 & 65.9 \\

			\bottomrule
		\end{tabular}
		}
	\end{center}
\end{table*}

To systematically evaluate each component's contribution, we conduct ablation studies by disabling individual mechanisms while keeping the rest intact.

\begin{itemize}
    \item[a)] \textbf{Vanilla Prompt:} $f_{\theta}$ directly generates final outputs via few-shot ICL, without \textsc{Python} code, multi-stage decomposition, or execution feedback, producing SQL for DB SKR tasks, direct answers for Table SKR tasks, and SPARQL for KG SKR tasks.

    \item[b)] \textbf{w/o Execution Feedback:} Code is generated per subtask but executed only after all subtasks complete, eliminating incremental self-correction.

    \item[c)] \textbf{w/o Code Merging:} Validated subtask codes are naively concatenated rather than merged.

    \item[d)] \textbf{w/o Knowledge Transfer:} Demonstrations are retrieved solely from the same dataset as the test query, isolating the benefit of task-agnostic retrieval afforded by the unified BOX representation.

    \item[e)] \textbf{w/o Step-wise Decomposition:} A single-stage code generation pass is used, measuring the overall impact of staged fine-grained reasoning.
\end{itemize}

To manage computational costs, we randomly sampled approximately 200 from each benchmark for ablation studies across 5 settings, 2 models, and 7 datasets, which yields reliable performance estimates consistent with full-dataset evaluation.
Table~\ref{tab:ablation_study} reports the results.
The full \textsc{Pandora} system achieves the best performance across all settings, with average F1 scores of \textbf{78.5\%} for \texttt{gpt-4o-mini} and \textbf{75.5\%} for \texttt{qwen3-8B}, validating the synergy among all proposed components.
Removing all components and falling back to few-shot ICL with direct output generation results in average F1 scores of 68.8\% and 65.5\% for the two backbones, corresponding to decreases of 9.7\% and 10.0\%, respectively.
Disabling per-subtask execution feedback leads to the largest single-component performance drops, reducing average F1 by 8.4\% and 6.1\%, respectively.
This underscores its critical role in incremental error detection and repair, a capability that is inherently unavailable in conventional SQL/SPARQL generation.
Replacing staged decomposition with single-pass code generation causes comparable degradations of 8.5\% and 8.7\%, respectively, confirming that fine-grained step-wise reasoning substantially outperforms monolithic generation.
Substituting LLM-based code merging with naive concatenation reduces average F1 by 2.4\% and 2.3\%, respectively.
Restricting demonstration retrieval to same-dataset examples decreases performance by 3.4\% and 4.0\%, respectively, with the most pronounced impacts on WTQ and GrailQA.
This highlights the benefit of cross-task demonstration sharing enabled by the unified \textsc{Box} representation.

\subsection{Zero-shot Evaluation}
\begin{figure}
    \centering
    \includegraphics[width=\columnwidth]{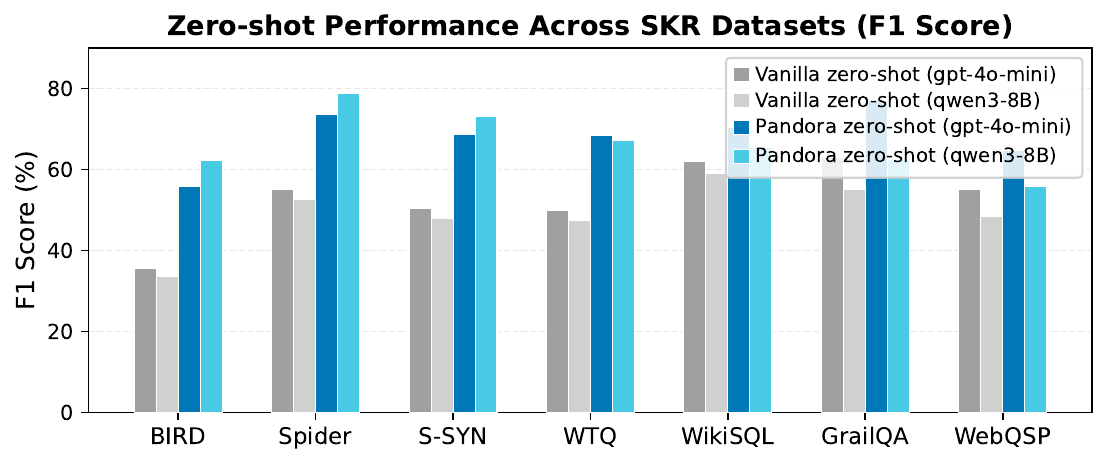}
    \caption{Zero-shot performance ($K{=}0$) across seven SKR datasets. We compare vanilla zero-shot (direct LLM generation without demonstrations) against \textsc{Pandora} zero-shot (full code-driven reasoning pipeline without any demonstrations). F1 score is reported for all datasets.}\label{fig:zero_shot}
\end{figure}

We evaluated zero-shot capability by setting $K{=}0$ (no demonstrations) and comparing vanilla direct generation against \textsc{Pandora}'s code-driven reasoning pipeline.
As shown in Figure~\ref{fig:zero_shot}, \textsc{Pandora} zero-shot with \texttt{gpt-4o-mini} achieves an average F1 of 68.4\%, substantially outperforming vanilla zero-shot by 15.4\%. Similarly, \textsc{Pandora} zero-shot with \texttt{qwen3-8B} averages 66.5\%, exceeding vanilla zero-shot by 17.4\%. These results confirm that the structured code reasoning pipeline, encompassing schema analysis, step-wise decomposition, execution feedback, and Code Merging, yields substantial gains even without any demonstrations, validating the effectiveness of \textsc{Pandora}'s execution-guided reasoning paradigm.
Beyond the overall comparison, a finer-grained analysis reveals complementary strengths between the two models. \texttt{qwen3-8B} outperforms \texttt{gpt-4o-mini} on DB tasks, reflecting its stronger code generation capability. Conversely, \texttt{gpt-4o-mini} maintains a clear advantage on KG tasks, suggesting superior multi-hop reasoning over knowledge graphs.

\subsection{LLM-as-judge Evaluation}
\label{sec:llm_judge}

\begin{figure}[t]
    \centering
    \includegraphics[width=\columnwidth]{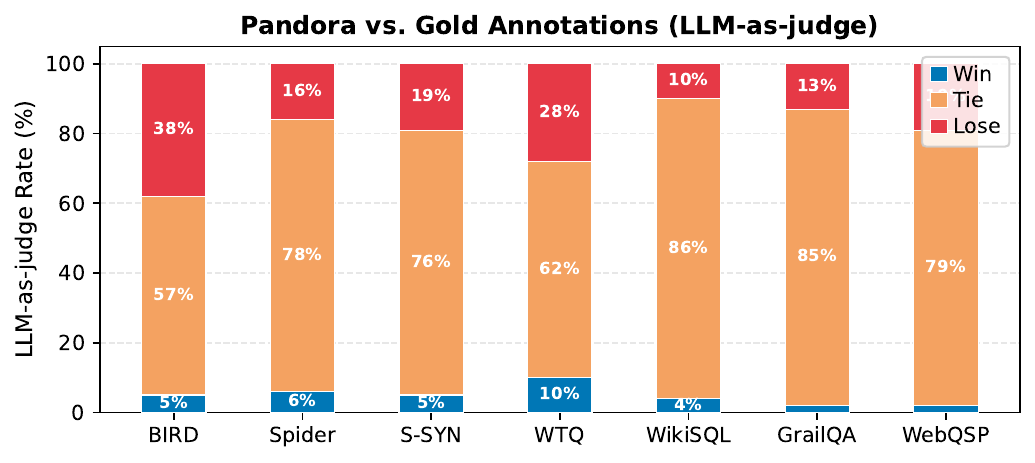}
    \caption{ LLM-as-judge comparison between \textsc{Pandora} + \texttt{gpt-4o-mini} predictions and gold annotations across seven datasets. Each stacked bar shows the Win/Tie/Lose rate, judged by GPT-4o using a rubric-based pairwise comparison that evaluates semantic equivalence beyond exact-match.}\label{fig:llm_judge}
\end{figure}

 To complement the exact-match evaluation, we conduct an LLM-as-judge assessment using GPT-4o with a rubric-based pairwise comparison protocol, which better captures semantic correctness in cases where valid answers may differ in format or phrasing from gold annotations. We classify each prediction into three categories: \textbf{Win} (Pandora's answer is correct but was penalized by the strict evaluation metric), \textbf{Tie} (Pandora and gold agree, confirming the original metric was correct), and \textbf{Lose} (Pandora's answer is genuinely incorrect).

As shown in Figure~\ref{fig:llm_judge}, the Win+Tie rate exceeds the main experimental results across all datasets, with improvements ranging from +2.1\% (WebQSP) to +4.6\% (BIRD), confirming that a small but consistent fraction of Pandora's answers were incorrectly penalized by strict EX/DA/F1 scoring. \textbf{Tie dominates} across all datasets (57\%--86\%), indicating that the majority of gold annotations are accurate and the original metrics are reliable for most cases. \textbf{Win accounts for a small portion} (2\%--10\%), revealing dataset-specific annotation quality differences: WTQ exhibits the highest Win rate (10\%), suggesting more frequent annotation ambiguities in table-based QA where equivalent answers may be expressed in diverse formats; in contrast, GrailQA and WebQSP show only 2\% Win rates, reflecting the more standardized and reliable gold labels in KGQA benchmarks.

Overall, the LLM-as-judge results demonstrate that \textsc{Pandora}'s code-driven reasoning produces outputs that are semantically well-aligned with expected answers, and that strict evaluation metrics, while generally reliable, may slightly underestimate performance on datasets with more flexible answer formats.

\subsection{Exploration of Code-driven Knowledge Transfer}
\begin{figure}
    \centering
	\includegraphics[width=0.45\textwidth]{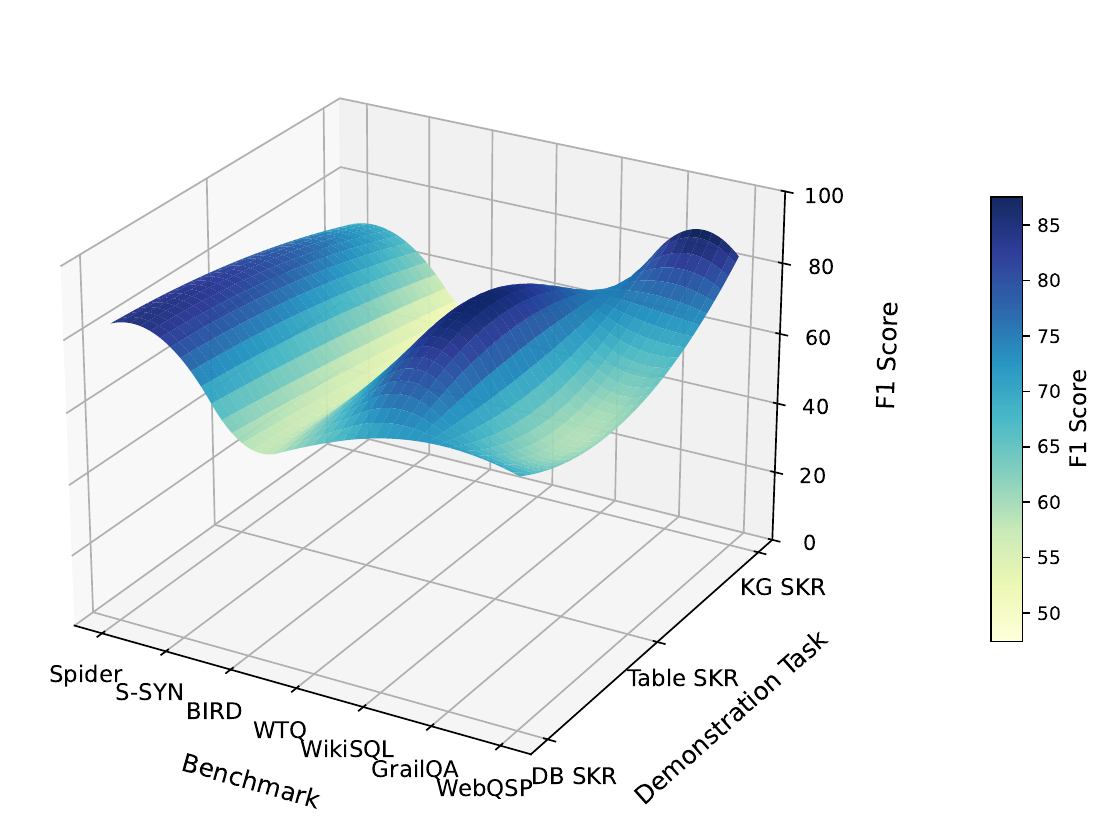}
	\caption{Cross-task memory transfer results (\%) of \textsc{Pandora} with \texttt{gpt-4o-mini}. Each entry reports the performance on a target task (column) using demonstrations from a source task (row); e.g., (\texttt{Table}, \texttt{GrailQA}) denotes the F1 score on GrailQA with table-task demonstrations. Approximately 200 questions are sampled per dataset to ensure effiency and cost.} \label{fig:heat_map}
\end{figure}
 To delve deeper into the potential of shared demonstrations for facilitating knowledge transfer, we evaluated \textsc{Pandora} on each benchmark using demonstrations from only one SKR task type (DB, Table, or KG).
The results are shown in Figure~\ref{fig:heat_map}, where the diagonal (e.g., DB-demos $\rightarrow$ Spider) represents the optimal same-type configuration, while off-diagonal entries reveal the cross-task transfer capability enabled by the unified BOX representation.
Several key observations emerge.
\textbf{First, cross-task knowledge transfer between DB and Table tasks is remarkably effective}: using Table demonstrations on DB benchmarks (Spider, S-SYN, BIRD) incurs only a modest 4--6\% drop compared to same-type DB demonstrations, and using DB demonstrations on Table benchmarks (WTQ, WikiSQL) incurs a similar 5--16\% drop. This relatively small gap indicates that the \textsc{Pandas}-based BOX representation successfully abstracts the shared structured reasoning patterns between databases and tables.
\textbf{Second, KG demonstrations still provide meaningful performance on DB/Table tasks:} on Spider, KG-demos achieve 68.4\% F1 (vs.\ 83.7\% for DB-demos), and on WikiSQL, KG-demos reach 68.6\% (vs.\ 86.1\% for Table-demos). While these drops are larger than the DB$\leftrightarrow$Table gap, they are far from catastrophic --- the model retains substantial reasoning capability even when demonstrations come from a completely different knowledge modality.
\textbf{Third, and most notably, DB and Table demonstrations each contribute strong signals to KG tasks: }DB-demos yield 72.6\% on GrailQA and 69.5\% on WebQSP, while Table-demos achieve 65.4\% and 62.7\%, respectively. The fact that DB-demos reach 72.6\% on GrailQA (only 11.6 points below the 84.2\% achieved with KG-demos) is particularly striking --- it demonstrates that structured data reasoning skills learned from DB tasks, such as multi-table joins, filtering, and aggregation, generalize meaningfully to knowledge graph query construction.

\noindent \textbf{Comparison with zero-shot baselines.}
To verify that heterogeneous demonstrations do not introduce harmful context noise, we further compare the cross-task transfer results with the vanilla zero-shot baselines in Figure~\ref{fig:zero_shot}. Vanilla zero-shot prompting achieves an average F1 of 53.0\% with \texttt{gpt-4o-mini}, whereas \textsc{Pandora} zero-shot already reaches 68.4\%. More importantly, the off-diagonal transfer settings in Figure~\ref{fig:heat_map} remain consistently above the corresponding vanilla zero-shot results. For example, using only DB demonstrations for KG reasoning reaches 72.6\% on GrailQA and 69.5\% on WebQSP, compared with vanilla zero-shot scores of 63.0\% and 55.0\%, respectively. Similarly, using KG demonstrations for DB/Table reasoning still improves over vanilla zero-shot on Spider, S-SYN, BIRD, WTQ, and WikiSQL. These results indicate that cross-task retrieval contributes useful executable reasoning patterns rather than merely injecting distracting heterogeneous context.

\subsection{Impact of Demonstration Number}
\begin{figure*}
 \centering
 \includegraphics[width=\textwidth]{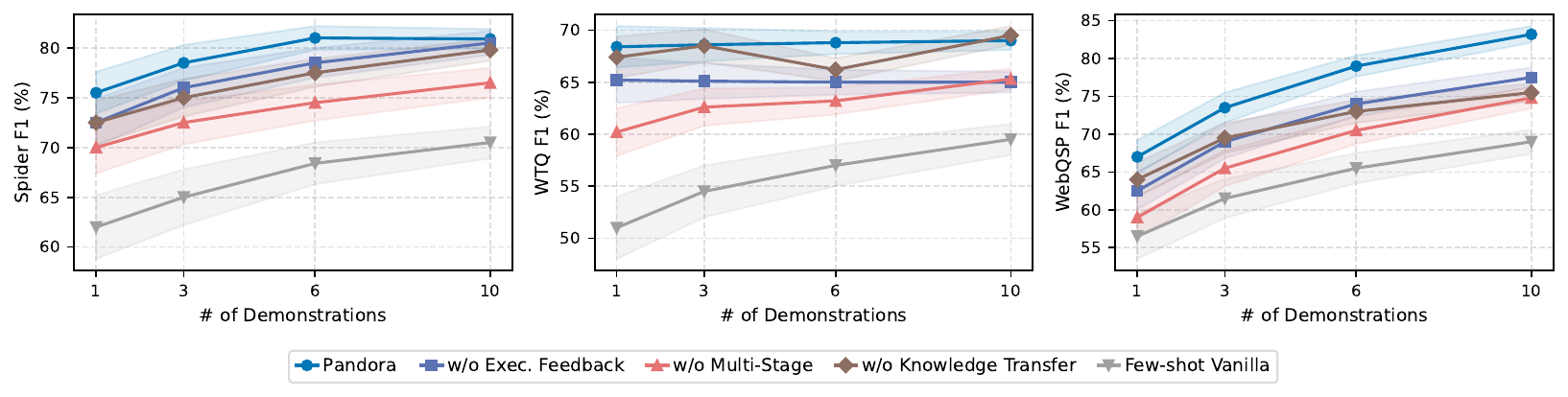}
 \caption{Effect of demonstration count (K) on \textsc{Pandora} + \texttt{qwen3-8B} across Spider, WTQ and WebQSP. Each data point reports the mean over three independent runs of approximately 200 randomly sampled examples, with error bars indicating standard deviation.}
 \label{fig:demo_number}
\end{figure*}
We vary the number of demonstrations ($K$) from $\{1, 3, 6, 10\}$, independently sampling 200 examples from the dev set and running the full pipeline three times per setting, reporting mean and standard deviation across runs. Results are illustrated in Figure~\ref{fig:demo_number}.
Overall, performance consistently improves with more demonstrations for both the full pipeline and vanilla baselines, confirming that in-context learning benefits from larger demonstration pools. The gain is especially pronounced on KG tasks (WebQSP), where performance increases by +16.2\% from $K{=}1$ to $K{=}10$.
Across all values of $K$, removing execution feedback causes a consistent 3--6\% drop, confirming that per-subtask error detection and repair provides orthogonal benefits independent of demonstration count.
Similarly, removing multi-stage decomposition yields a stable 8\% gap across all $K$ settings, indicating that step-wise reasoning is a structural advantage rather than a retrieval-dependent one.
The knowledge transfer memory exhibits notable $K$-sensitivity: at $K{=}1$, removing it causes minimal degradation, but the gap widens substantially as $K$ increases, reaching 3.4\% on Spider and 7.7\% on WebQSP at $K{=}10$. This demonstrates that cross-task demonstration retrieval becomes increasingly valuable as the pool of available examples grows.

\subsection{Impact of Memory Sampling Ratio}
\label{sec:memory_ratio}

\begin{figure}[t]
 \centering
 \includegraphics[width=0.9\columnwidth]{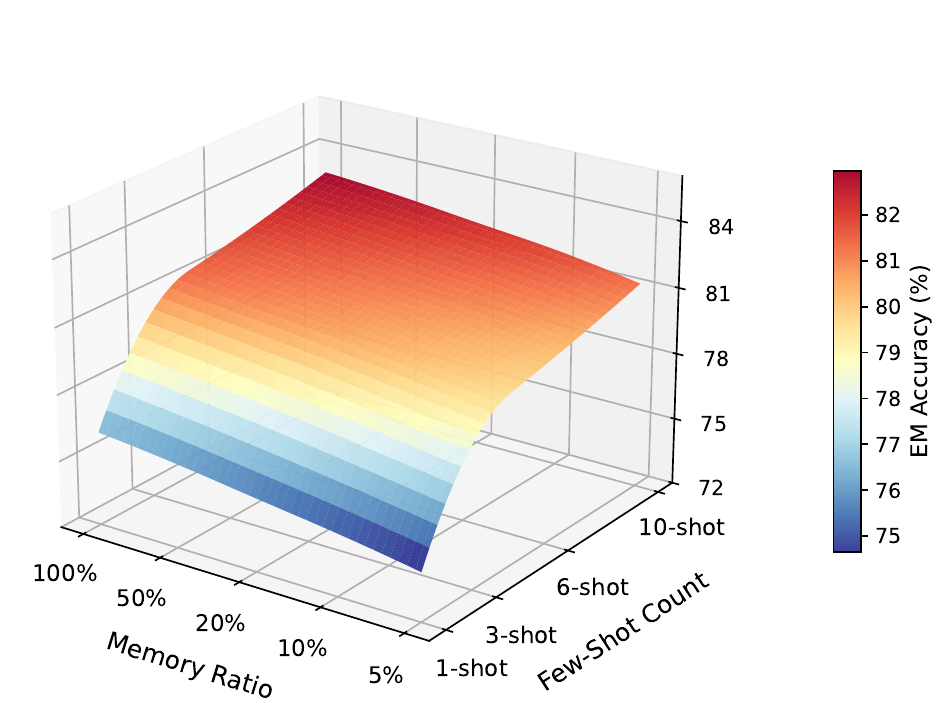}
 \caption{Execution accuracy (\%) on Spider as a function of memory construction ratio and few-shot count K, evaluated on 200 randomly sampled questions using \textsc{Pandora} + \texttt{gpt-4o-mini}.}\label{fig:memory_ratio}
\end{figure}

\begin{figure}[t]
 \centering
 \includegraphics[width=\columnwidth]{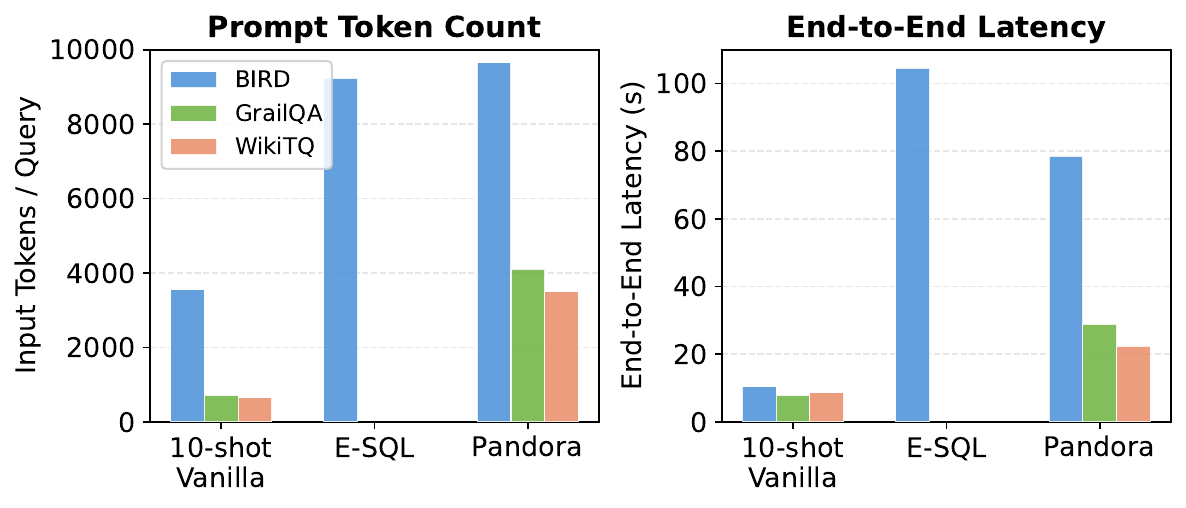}
 \caption{Prompt token count and end-to-end latency comparison across BIRD, GrailQA, and WikiTQ. Zero-shot and 10-shot Vanilla prompt baselines.}\label{fig:efficiency}
\end{figure}

 We conduct a systematic test on Spider to study the joint effect of memory construction ratio (percentage of training data sampled to build the demonstration memory) and demonstration count $K$, evaluated across all $5 \times 4$ configurations (ratios: 5\%, 10\%, 20\%, 50\%, 100\%; $K \in \{1, 3, 6, 10\}$). All evaluations are performed on about 200 randomly sampled test examples using \texttt{gpt-4o-mini} with the full \textsc{Pandora} pipeline. Results are shown in Figure~\ref{fig:memory_ratio}.
Two clear trends emerge from the 3D surface visualization. First, increasing the memory construction ratio yields strongly diminishing returns: expanding memory from 5\% to 100\% improves accuracy by only 1.7\% at $K{=}10$, confirming that a compact memory pool already captures a sufficiently diverse set of reasoning patterns. Second, the few-shot count $K$ has a far more pronounced effect, with gains of 6.8--7.1\% observed when increasing from $K{=}1$ to $K{=}10$ across all memory ratios. The steeper gradient along the $K$-axis indicates that the model benefits more directly from richer in-context demonstrations than from a larger retrieval pool, suggesting that demonstration quality and quantity matter more than memory coverage in practice.

\subsection{End-to-End Latency and Token Cost.}
\label{sec:end_to_end_efficiency}

 Figure~\ref{fig:efficiency} reports the prompt token overhead and end-to-end latency of 10-shot Vanilla, E-SQL, and \textsc{Pandora} across three representative datasets spanning DB, KG, and Table SKR tasks. E-SQL is evaluated exclusively on BIRD, as it is designed specifically for SQL generation and cannot be directly applied to GrailQA or WikiTQ. We include E-SQL as a reference point because it offers an open-source implementation, achieves competitive performance on the challenging BIRD benchmark, and thus represents a realistic baseline for token consumption in large-scale structured database processing.
\textsc{Pandora} inevitably incurs higher per-call token consumption than the 10-shot Vanilla baseline, primarily due to the rich schema context embedded within each BOX unit — including table schemas, sampled rows, value-linking hints, and retrieval-augmented in-context demonstrations. Aggregated across its full multi-stage pipeline, \textsc{Pandora} processes approximately 9{,}600 tokens per query on average, resulting in end-to-end latencies of 22.4s (WikiTQ), 28.8s (GrailQA), and 78.5s (BIRD) when powered by \texttt{gpt-4o-mini}, compared to 8.6--10.4s for the single-step 10-shot baseline.
This additional overhead is an inherent consequence of the multi-stage, context-enriched pipeline design; however, it is precisely this design that underlies the substantial accuracy improvements documented in Tables\ref{tab:baselines_db} and\ref{tab:baselines_kg}. Importantly, this latency is not a fundamental bottleneck: independent pipeline stages can be parallelized, API calls can be batched, and schema contexts can be cached across queries, all of which would substantially reduce wall-clock latency. We therefore consider the current overhead to be an acceptable and addressable cost for accuracy-critical deployment scenarios.

\subsection{Error Analysis.}

\begin{figure}[t]
 \centering
 \includegraphics[width=\columnwidth]{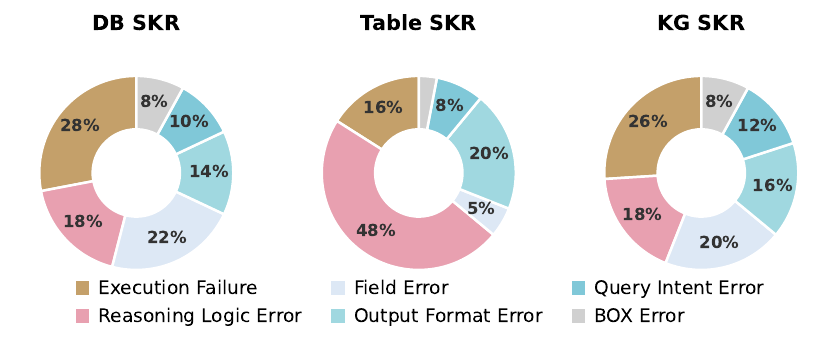}
 \caption{Breakdown of error categories across DB, Table, and KG reasoning tasks produced by \textsc{Pandora} with \texttt{gpt-4o-mini}, highlighting the dominant failure modes for each task type.}\label{fig:error_cause}
\end{figure}

We analyze failure cases of \textsc{Pandora} with \texttt{gpt-4o-mini} by manually inspecting 50 random error instances per task, classifying them into six categories: \textbf{Execution Failure} and five types of \textbf{Silent Logical Failures} (i.e., errors that produce no runtime exception but yield incorrect answers), namely \textbf{BOX Error}, \textbf{Field Error}, \textbf{Reasoning Logic Error}, \textbf{Query Intent Error}, and \textbf{Output Format Error}. The analysis results are shown in Figure~\ref{fig:error_cause}.

Several patterns emerge. \textbf{Execution failures} are most prevalent in DB (28\%) and KG (26\%) tasks, reflecting the higher schema complexity of multi-table joins and KG entity linking, whereas Table SKR exhibits the lowest rate (16\%) owing to its simpler single-table representation. \textbf{Silent logical failures} dominate across all tasks, with \textbf{reasoning logic errors} being the leading cause in Table SKR (48\%), as multi-step aggregations and compositional operations on flat tables pose significant orchestration challenges for the LLM. \textbf{Field errors} are substantial in DB (22\%) and KG (20\%) tasks, suggesting that schema linking hints alone are insufficient to resolve ambiguities among semantically similar columns. \textbf{Output format errors} (14\%--20\%) persist across all tasks, pointing to the need for stricter answer postprocessing. By contrast, \textbf{BOX errors} and \textbf{query intent errors} remain minor (3\%--12\%), confirming that the schema risk analyzer and task decomposition components are generally effective at grounding queries to correct schema elements.

\subsection{Cross-Source Reasoning Evaluation}
\label{sec:cross_source_exp}

\begin{figure}[t]
 \centering
 \includegraphics[width=\columnwidth]{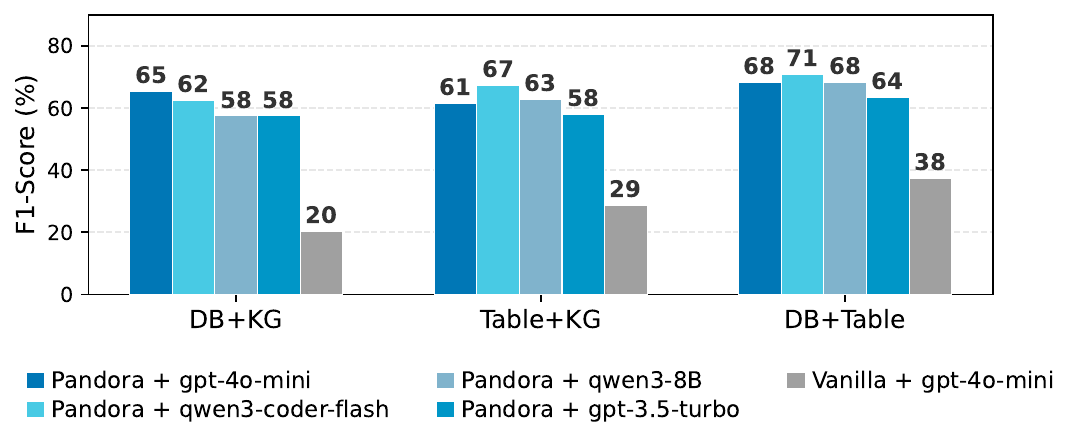}
 \caption{Cross-source reasoning F1-Score on 23 manually constructed queries spanning three heterogeneous source combinations.}\label{fig:cross_source}
\end{figure}

To evaluate \textsc{Pandora}'s cross-source reasoning capability, we manually constructed a benchmark of 23 queries requiring joint reasoning over two heterogeneous knowledge sources, covering three combinations: DB+KG (8 queries), Table+KG (7 queries), and DB+Table (8 queries). For DB+KG, we pair Spider databases with Freebase subgraphs (e.g., querying a movie DB while referencing director birthplaces from a KG); for Table+KG, WikiTQ tables are paired with Freebase subgraphs (e.g., filtering universities by alumni information); for DB+Table, Spider databases are paired with manually constructed CSV tables (e.g., combining employee records with financial metrics). Each query is manually checked to ensure that the answer cannot be obtained from either source alone. Gold answers are derived by executing the corresponding source-specific operations and then verifying the final joined answer set.

All sources are converted into BOX representations and processed through \textsc{Pandora}'s standard multi-stage pipeline without adding any cross-source examples to the memory. Thus, the experiment tests whether the memory constructed from ordinary single-source DB, Table, and KG tasks can generalize to multi-source joins. The Vanilla Prompt baseline receives the same source descriptions and BOX schemas, but directly predicts the final answer without box-aware linking, step-wise decomposition, execution feedback, or retrieved code demonstrations. We report answer-set F1 after normalizing string values and entity names.

Results are shown in Figure~\ref{fig:cross_source}. \textsc{Pandora} with \texttt{gpt-4o-mini} achieves an average F1 score of 65.0\% across all three source combinations, outperforming the Vanilla Prompt baseline by more than 35 points on average. This demonstrates that the BOX representation and code-driven reasoning pipeline provide a principled and generalizable foundation for cross-source joint reasoning. Several noteworthy patterns emerge from the results.
\textbf{Source structure alignment drives peak performance.} DB+Table yields the highest F1 score (68.2\%), as both sources share inherently tabular structures that align naturally under the unified BOX representation, minimizing the structural gap that the reasoning pipeline must bridge.
\textbf{Code-specialized models exhibit a surprising advantage.} \texttt{qwen3-coder-flash} not only performs competitively but even surpasses \texttt{gpt-4o-mini} on Table+KG (67.3\% vs.\ 61.4\%) and DB+Table (70.9\% vs.\ 68.2\%), suggesting that models pre-trained on large volumes of code may possess an inherent advantage in the structured join and traversal operations required for cross-source reasoning.
\textbf{Table+KG poses a moderate but consistent challenge} across all evaluated models, primarily attributed to entity resolution mismatches between raw table cell values and canonical KG entity names --- a known difficulty in heterogeneous data integration that warrants dedicated treatment in future work.
\textbf{The necessity of the full pipeline is further confirmed} by the stark underperformance of the Vanilla baseline across all combinations (20.3\%--37.5\%), indicating that schema linking, query decomposition, and execution feedback are collectively essential for effective cross-source reasoning, rather than any single component alone.
While this benchmark remains preliminary in scale, the results consistently confirm that \textsc{Pandora} enables non-trivial cross-source reasoning out of the box without task-specific adaptation. We leave large-scale benchmark evaluation to future work.

\subsection{Security Considerations}
\label{sec:security}
Executing LLM-generated Python code introduces inherent security risks, as the model could in principle produce unintended or malicious instructions, such as file system operations, network requests, or process spawning. To mitigate this, \textsc{Pandora} adopts a practical multi-layered defense strategy. All generated code is executed within an isolated subprocess whose access to the host environment is strictly limited: only a small whitelist of data-oriented libraries (\texttt{pandas}, \texttt{numpy}, \texttt{math}, \texttt{re}, etc.) is permitted, and each execution is subject to configurable CPU time and memory quotas to prevent runaway processes. Furthermore, the instruction templates explicitly guide the LLM to focus on data manipulation operations, and actively discourage any system-level calls.
We acknowledge that no sandboxing mechanism offers absolute guarantees. The measures described above are intended to provide a reasonable security baseline suitable for research and development settings. For production deployments where stricter isolation is required, additional hardening strategies, such as containerization via Docker, would be advisable.

\subsection{Applicability and Limitations}
\label{sec:applicability_limitations}
\textsc{Pandora} is most applicable to structured reasoning scenarios where schemas are available, source data can be exposed as tables or query-local KG subgraphs, and the application can tolerate multi-stage LLM latency in exchange for higher reasoning accuracy and executable intermediate states. The current implementation assumes that topic entities for KGQA can be obtained from benchmark annotations or an external entity linker, and that cross-source alignment can be approximated through normalized surface forms and LLM-inferred join keys. Therefore, entity-linking errors, noisy table cell values, and ambiguous cross-source aliases may directly affect the quality of the generated BOXes and downstream reasoning.

The framework is less suitable for scenarios that require loading an entire enterprise database or billion-scale KG into in-memory Pandas DataFrames. In such deployments, BOX should be viewed as a logical reasoning interface containing schemas, sampled rows, and query-local evidence, while the physical execution layer should push computation down to database engines, SPARQL endpoints, or scalable dataframe systems such as Dask or Polars. \textsc{Pandora} is also not optimized for ultra-low-latency applications, because its decomposition, execution feedback, and code-merging stages introduce additional model calls. Finally, although the sandboxing strategy in Section~\ref{sec:security} mitigates common risks, production use requires stronger isolation and auditing of generated code. These boundaries clarify that \textsc{Pandora}'s main contribution is a unified and executable reasoning abstraction, rather than a replacement for large-scale storage engines, entity linking systems, or production-grade secure execution infrastructure.

\section{Related Work}
\subsection{DB Structured Knowledge Reasoning}
The task of translating natural language questions into executable SQL queries, commonly known as text-to-SQL, has a rich history. Traditional methods predominantly focused on advancing model architectures, often employing specialized encoder-decoder frameworks~\cite{DBLP:conf/iclr/0009WLWTYRSX21}, and designing sophisticated intermediate representations (IRs)~\cite{DBLP:conf/acl/GuoZGXLLZ19}. These IRs, such as domain-specific languages or abstract syntax trees, aimed to bridge the semantic gap between natural language and the formal structure of SQL. More recently, the advent of Large Language Models (LLMs) has catalyzed a paradigm shift. Advanced techniques such as decomposing complex questions into simpler, manageable sub-problems, leveraging chain-of-thought prompting to generate explicit reasoning steps for query construction~\cite{DBLP:conf/nips/Wei0SBIXCLZ22}, and employing self-consistency to sample multiple reasoning paths and select the most robust query~\cite{DBLP:conf/iclr/0002WSLCNCZ23} have markedly improved performance. These approaches have led to state-of-the-art results on challenging benchmarks~\cite{DBLP:conf/nips/PourrezaR23,DBLP:conf/aaai/CenLLW25,DBLP:journals/pvldb/GaoWLSQDZ24,DBLP:journals/corr/abs-2405-16755,DBLP:journals/corr/abs-2410-01943}. Concurrently, a significant line of work has focused on fine-tuning open-source LLMs, demonstrating that with specialized training, these models can achieve performance competitive with proprietary, closed-box models, offering a more accessible and adaptable alternative~\cite{DBLP:journals/pacmmod/LiZLFZZWP0024,DBLP:conf/emnlp/PourrezaR24}.

\subsection{Table Structured Knowledge Reasoning}
Reasoning over semi-structured tabular data~\cite{DBLP:journals/tkde/ZhangWSWTQCWY24} to answer NLQs has long been a core challenge in NLP. Traditional approaches typically relied on two main strategies: semantic parsing, which aimed to translate an NLQ into a logical form executable over the table schema~\cite{DBLP:conf/acl/PasupatL15}, and embedding-based methods. The latter focused on learning dense vector representations for both the question and the tabular data, enabling query-table matching and cell selection through similarity metrics~\cite{DBLP:conf/acl/YinNYR20, DBLP:conf/aaai/DengXLYDFLS19}. The current landscape is dominated by LLMs, which have introduced more powerful and flexible reasoning capabilities. Modern methods leverage the remarkable ability of LLMs to understand and manipulate data contextually. Models like TAPEX~\cite{DBLP:conf/iclr/LiuCGZLCL22} pioneered pre-training objectives that mimic table-based question answering, while Binder~\cite{DBLP:conf/iclr/ChengX0LNHXROZS23} explicitly models the connections between question mentions and table cells. More recent work, such as DATER~\cite{DBLP:conf/sigir/YeHYLHL23}, focuses on decomposing the reasoning process itself, allowing the model to tackle complex multi-step operations over tables and achieve superior performance on diverse benchmarks.

\subsection{KG Structured Knowledge Reasoning}
Knowledge Graph Question Answering (KGQA)~\cite{DBLP:journals/tkde/ChenLQWW23} seeks to answer natural language questions by reasoning over large-scale knowledge graphs (KGs). Historically, this field was characterized by methods based on semantic parsing, which converted questions into executable logical forms like SPARQL queries~\cite{DBLP:conf/emnlp/BerantCFL13, DBLP:conf/acl/YihCHG15}. An alternative and parallel direction involved embedding-based approaches, where questions and KG entities/relations were projected into a shared low-dimensional space to find answers via query matching and neighborhood exploration~\cite{DBLP:conf/iclr/DasDZVDKSM18}. Recent innovations have effectively harnessed the power of LLMs to overcome the brittleness of traditional parsers. For example, DecAF~\cite{DBLP:conf/iclr/YuZNZL0HWWX23} integrates the generation of logical forms with direct answer generation in a decompose-and-fuse framework. Other methods focus on enhancing the LLM's access to relevant knowledge; KB-BINDER~\cite{DBLP:conf/acl/LiMZGSC23}, for instance, uses the classic BM25 algorithm to retrieve a relevant subgraph from the KG before prompting the LLM. Furthermore, KB-Coder~\cite{DBLP:conf/aaai/NieZW024} reframes the KGQA task within a code-style paradigm, leveraging in-context learning to guide the LLM in generating code-like reasoning paths, which has shown to produce highly effective and interpretable results.

\subsection{Unified Structured Knowledge Reasoning}
The pursuit of a single model capable of reasoning over heterogeneous structured knowledge is a key frontier. Early unified frameworks, such as UnifiedSKG~\cite{DBLP:conf/emnlp/XieW0ZSYWZYWZWL22} and StructLM~\cite{DBLP:journals/corr/abs-2402-16671}, made significant strides by integrating multiple structured knowledge datasets. These models typically involved fine-tuning pre-trained language models like T5~\cite{DBLP:journals/jmlr/RaffelSRLNMZLL20} or code-centric models like CodeLlama~\cite{DBLP:journals/corr/abs-2308-12950} on a unified input-output format, thereby enhancing the model's general understanding of structured knowledge. More recent unified QA frameworks have been developed to handle diverse structured data types within a single inference process. For instance, StructGPT~\cite{DBLP:conf/emnlp/JiangZDYZW23} employs an iterative reading-then-reasoning strategy, using the LLM to call external tools to retrieve evidence and progressively generate an answer. Similarly, Readi~\cite{DBLP:conf/acl/ChengZXYZQHCL0R24} iteratively refines reasoning paths to extract evidence from various sources and synthesize a final response.

 \subsection{Graph-based and Agentic RAG Frameworks}
GraphRAG~\cite{edge2024graphrag} constructs entity knowledge graphs from source documents and synthesizes community summaries for global sensemaking, with subsequent work extending it via dual-level retrieval~\cite{guo2024lightrag}, cost-efficient RL-based routing~\cite{fan2026graphragrouter}, and path-aware subgraph reasoning~\cite{yuan2026pathgraphrag}. Agentic RAG frameworks further delegate multi-step retrieval and reasoning to autonomous LLM agents, with applications spanning clinical decision support~\cite{chen2026medcopilot} and plugin-style episodic memory~\cite{yang2026plugmem}. However, all these approaches are fundamentally oriented toward \textit{summarization over unstructured text corpora} — relying on pre-computed community summaries for approximate answering — whereas \textsc{Pandora} targets \textit{logical reasoning over structured, heterogeneous knowledge sources} (tables, databases, and knowledge graphs) through executable code generation on a unified BOX representation, yielding deterministic and verifiable results with step-wise execution feedback and post-execution reflection.

\section{Conclusion}
\label{sec:conclusion}
In this paper, we introduced \textsc{Pandora}, a novel Unified Structured Knowledge Reasoning framework that addresses the limitations of existing task-specific approaches through two key innovations. By adopting \textsc{Pandas}-based Python code as a unified representation for heterogeneous structured knowledge sources, including tables, databases, and knowledge graphs, \textsc{Pandora} establishes a cohesive and LLM-aligned foundation for cross-task reasoning. Building upon this representation, \textsc{Pandora} further leverages automated cross-task memory construction and multi-stage code-driven reasoning with execution feedback, enabling adaptive refinement and effective knowledge transfer across diverse SKR tasks. Extensive experiments on seven widely used benchmarks spanning three SKR tasks demonstrate that \textsc{Pandora} consistently outperforms existing unified reasoning frameworks while remaining competitive with task-specific methods. We also present a preliminary exploration of cross-source reasoning, confirming that \textsc{Pandora}'s design generalizes naturally to joint reasoning over multiple heterogeneous sources without task-specific adaptation. Due to cost constraints, evaluation on the most advanced reasoning models such as \texttt{gpt-5.5}, \texttt{qwen3.6-plus} and \texttt{Claude 4.6} remains outstanding. Future work will focus on scaling the framework to broader knowledge types, constructing large-scale cross-source benchmarks, and exploring tighter integration with advanced LLMs for real-world deployment scenarios.

\section{Acknowledgements}
This work is supported by National Nature Science Foundation of China under No.62506071, No.6250072425, and No.62476058. We thank the Big Data Computing Center of Southeast University for providing the facility support on the numerical calculations in this paper.

\input{main.bbl}

\clearpage

\appendices
\setcounter{secnumdepth}{1}

\section{Table of Notations}
\label{sec:notations}
To aid readability, we summarize the most frequently used mathematical notations in Table~\ref{tab:notations}.

\begin{table*}[h]
\centering
\caption{Summary of key notations.}
\label{tab:notations}
\scalebox{1.0}{
\begin{tabular}{lp{11cm}}
\toprule
\textbf{Symbol} & \textbf{Description} \\
\midrule
$\mathcal{Q}$ & Natural language question \\
$\mathcal{S}$ & Structured knowledge source (table $\mathcal{T}$, database $\mathcal{D}$, or KG $\mathcal{K}$) \\
$\mathcal{A}$ & Target answer \\
$\mathcal{B}_i = (b_i, \Phi_i, \Psi_i)$ & The $i$-th BOX, with name $b_i$, field set $\Phi_i$, and value sets $\Psi_i$ \\
$\{\mathcal{B}_1, \dots, \mathcal{B}_n\}$ & Set of all BOXes derived from $\mathcal{S}$ \\
$\{\mathcal{B}'_1, \dots, \mathcal{B}'_k\}$ & Subset of linked BOXes identified during schema linking \\
$\phi$ & A field name (column name or KG relation) \\
$\psi$ & A field value (cell content or KG entity) \\
$\hat{\Phi}$ & LLM-enriched field semantics: $\{(\phi, \mu(\phi))\}$ \\
$\mathcal{L}$ & Schema/value linking result \\
$\mathcal{R}_{\text{risk}}$ & Unified risk hints from schema analysis and linking \\
$\mathcal{T}_{\text{decomp}} = [t_1, \dots, t_m]$ & Decomposed subtask sequence \\
$\mathcal{C}_i$ & Generated code for subtask $t_i$ \\
$\mathcal{C}_{\text{a}}$ & Accumulated code after all subtasks \\
$\mathcal{C}_{\text{merge}}$ & LLM-fused code \\
$\mathcal{M}$ & Memory storing demonstration examples \\
$K$ & Number of few-shot demonstrations \\
$L$ & Maximum repair iterations \\
$f_{\theta}$ & Backbone LLM for code generation \\
$g_{\theta}$ & Encoder LLM for demonstration retrieval \\
$\mathcal{K}^*$ & Extracted $H$-hop subgraph from KG \\
$\mathcal{E}^*, \mathcal{R}^*$ & Entity set and relation set of the extracted subgraph \\
$\Phi_{\gamma}^{1:N}$ & Field names for the BOX of entity type $\gamma$, where $N = |\Phi_{\gamma}|$ is the number of fields in that BOX. \\
\bottomrule
\end{tabular}
}
\end{table*}

\section{More Examples of Pandas Code}
\label{sec:pandas_example}

\subsection{Examples of Text-to-SQL}
Below are examples of converting SQL queries into equivalent Pandas code, covering cases like filtering, ordering, grouping, and nested queries.

\noindent \texttt{\footnotesize NLQ 1: Show the name and number of employees for the departments managed \\by heads whose temporary acting value is 'Yes'?}

\begin{tcolorbox}[colback=white, colframe=black, arc=2mm, boxrule=0.5pt]
\scriptsize
\begin{lstlisting}[style=arxivcode,language=sql]
SELECT T1.name , T1.num_employees
FROM department AS T1
JOIN management AS T2
ON T1.department_id = T2.department_id
WHERE T2.temporary_acting = 'Yes'
\end{lstlisting}
\end{tcolorbox}

\begin{tcolorbox}[colback=white, colframe=black, arc=2mm, boxrule=0.5pt]
\scriptsize
\begin{lstlisting}[style=arxivcode,language=python]
merged_df = pd.merge(department, management,                                on='department_id')
result = merged_df[merged_df['temporary_acting']
            == 'Yes'][['name', 'num_employees']]
\end{lstlisting}
\end{tcolorbox}


\noindent \texttt{\footnotesize NLQ 2: What are the ids of the students who registered for course 301 most recently?}

\begin{tcolorbox}[colback=white, colframe=black, arc=2mm, boxrule=0.5pt]
\scriptsize
\begin{lstlisting}[style=arxivcode,language=sql]
SELECT student_id
FROM student_course_attendance
WHERE course_id = 301
ORDER BY date_of_attendance DESC LIMIT 1
\end{lstlisting}
\end{tcolorbox}

\begin{tcolorbox}[colback=white, colframe=black, arc=2mm, boxrule=0.5pt]
\scriptsize
\begin{lstlisting}[style=arxivcode,language=python]
filtered_df = student_course_attendance[
    student_course_attendance['course_id'] == 301
]
sorted_df = filtered_df.sort_values(
    by='date_of_attendance',
    ascending=False
)
result = sorted_df.iloc[0]['student_id']
\end{lstlisting}
\end{tcolorbox}


\noindent \texttt{\footnotesize NLQ 3: For each zip code, what is the average mean temperature for all \\ dates that start with '8'?}

\begin{tcolorbox}[colback=white, colframe=black, arc=2mm, boxrule=0.5pt]
\scriptsize
\begin{lstlisting}[style=arxivcode,language=sql]
SELECT zip_code, avg(mean_temperature_f)
FROM weather
WHERE date LIKE "8/%"
GROUP BY zip_code
\end{lstlisting}
\end{tcolorbox}

\begin{tcolorbox}[colback=white, colframe=black, arc=2mm, boxrule=0.5pt]
\scriptsize
\begin{lstlisting}[style=arxivcode,language=python]
filtered_df = weather[
    weather['date'].str.startswith('8/')
]
result = filtered_df.groupby('zip_code')
         ['mean_temperature_f'].mean().reset_index()
result.columns = [
'zip_code',
'avg_mean_temperature_f'
]
\end{lstlisting}
\end{tcolorbox}


\noindent \texttt{\footnotesize NLQ 4: Which department has more than 1 head at a time? List the id, name and the number of heads.}

\begin{tcolorbox}[colback=white, colframe=black, arc=2mm, boxrule=0.5pt]
\scriptsize
\begin{lstlisting}[style=arxivcode,language=sql]
SELECT T1.department_id , T1.name , count(*)
FROM management AS T2 JOIN department AS T1
ON T1.department_id = T2.department_id
GROUP BY T1.department_id
HAVING count(*) > 1
\end{lstlisting}
\end{tcolorbox}

\begin{tcolorbox}[colback=white, colframe=black, arc=2mm, boxrule=0.5pt]
\scriptsize
\begin{lstlisting}[style=arxivcode,language=python]
merged_df = pd.merge(department, management,
                    on='department_id', how='inner')
grouped = merged_df.groupby(
    ['department_id', 'name']
).size().reset_index(name='count')
result = grouped[grouped['count'] > 1]
\end{lstlisting}
\end{tcolorbox}


\noindent \texttt{\footnotesize NLQ 5: What is the average bike availability in stations that are not \\located in Palo Alto?}

\begin{tcolorbox}[colback=white, colframe=black, arc=2mm, boxrule=0.5pt]
\scriptsize
\begin{lstlisting}[style=arxivcode,language=sql]
SELECT avg(bikes_available)
FROM status
WHERE station_id NOT IN (
    SELECT id
    FROM station
    WHERE city = "Palo Alto"
)
\end{lstlisting}
\end{tcolorbox}

\begin{tcolorbox}[colback=white, colframe=black, arc=2mm, boxrule=0.5pt]
\scriptsize
\begin{lstlisting}[style=arxivcode,language=python]
palo_alto_stations = station[
    station['city'] == "Palo Alto"
]['id']
filtered_status = status[
    ~status['station_id'].isin(palo_alto_stations)
]
result = filtered_status['bikes_available'].mean()
\end{lstlisting}
\end{tcolorbox}


\subsection{Examples of KGQA}
Here are examples of converting SPARQL queries into their equivalent Pandas code. These examples cover cases such as multi-hop queries, counting, filtering by type, and finding argmax/argmin.

\noindent \texttt{\footnotesize NLQ 1: which hotel grading authority awards servigroup papa luna hotel?}

\begin{tcolorbox}[colback=white, colframe=black, arc=2mm, boxrule=0.5pt]
\scriptsize
\begin{lstlisting}[style=arxivcode,language={}]
PREFIX : <http://rdf.freebase.com/ns/>
SELECT (?x0 AS ?value) WHERE {
    SELECT DISTINCT ?x0  WHERE {
        ?x0 :type.object.type
            :travel.hotel_grading_authority .
        ?x1 :type.object.type :travel.hotel_grade .
        VALUES ?x2 { :m.011nyts3 }
        ?x1 :travel.hotel_grade.awarded_by ?x0 .
        ?x2 :travel.hotel.grade ?x1 .
        FILTER ( ?x0 != ?x1
                && ?x0 != ?x2
                && ?x1 != ?x2 )
    }
}
\end{lstlisting}
\end{tcolorbox}

\begin{tcolorbox}[colback=white, colframe=black, arc=2mm, boxrule=0.5pt]
\scriptsize
\begin{lstlisting}[style=arxivcode,language=python]
result = hotel_grade[
    hotel_grade['hotel'] == 'Servigroup Papa
    Luna Hotel']['awarded_by'].tolist()
\end{lstlisting}
\end{tcolorbox}



\noindent \texttt{\footnotesize NLQ 3: which countries does russia border?}

\begin{tcolorbox}[colback=white, colframe=black, arc=2mm, boxrule=0.5pt]
\scriptsize
\begin{lstlisting}[style=arxivcode,language={}]
PREFIX ns: <http://rdf.freebase.com/ns/>
SELECT DISTINCT ?x
WHERE {
    ns:m.06bnz ns:location.location.adjoin_s ?y .
    ?y ns:location.adjoining_relationship.adjoins ?x .
    ?x ns:common.topic.notable_types ns:m.01mp .
}
\end{lstlisting}
\end{tcolorbox}

\begin{tcolorbox}[colback=white, colframe=black, arc=2mm, boxrule=0.5pt]
\scriptsize
\begin{lstlisting}[style=arxivcode,language=python]
merged_data = location.merge(
    adjoining_relationship, left_on='adjoin_s',
    right_on='adjoining_relationship'
)
result = merged_data.loc[
    (merged_data['location'] == 'Russia'),
    'adjoins'
].unique().tolist()
result = [x for x in result if x != 'Russia']
\end{lstlisting}
\end{tcolorbox}

\noindent \texttt{\footnotesize NLQ 4: how many religious leaders are thai chinese?}

\begin{tcolorbox}[colback=white, colframe=black, arc=2mm, boxrule=0.5pt]
\scriptsize
\begin{lstlisting}[style=arxivcode,language={}]
PREFIX : <http://rdf.freebase.com/ns/>
SELECT (COUNT(?x0) AS ?value) WHERE {
    SELECT DISTINCT ?x0  WHERE {
        ?x0 :type.object.type
            :religion.religious_leader .
        VALUES ?x1 { :m.04l_pt }
        ?x0 :people.person.ethnicity ?x1 .
        FILTER ( ?x0 != ?x1  )
    }
}
\end{lstlisting}
\end{tcolorbox}

\begin{tcolorbox}[colback=white, colframe=black, arc=2mm, boxrule=0.5pt]
\scriptsize
\begin{lstlisting}[style=arxivcode,language=python]
result = [[
    len(religious_leader[religious_leader[
    'religious_leader'].isin(
        person['person'][person['ethnicity'] == 'jew']
    )])
]]
\end{lstlisting}
\end{tcolorbox}


\subsection{Examples of TableQA}

\noindent \texttt{\footnotesize NLQ 1: in which county did brown receive the most votes?}

\begin{tcolorbox}[colback=white, colframe=black, arc=2mm, boxrule=0.5pt]
\scriptsize
\begin{lstlisting}[style=arxivcode,language=python]
brown_votes = df['Votes'].str.replace(',',
                '').astype(int)
max_votes_index = brown_votes.idxmax()
result = [[df.loc[max_votes_index, 'County']]]
\end{lstlisting}
\end{tcolorbox}

\noindent \texttt{\footnotesize NLQ 2: who was the top position american driver?}

\begin{tcolorbox}[colback=white, colframe=black, arc=2mm, boxrule=0.5pt]
\scriptsize
\begin{lstlisting}[style=arxivcode,language=python]
df['Pos'] = pd.to_numeric(df['Pos'], errors='coerce')
result = df.nsmallest(1, 'Pos')['Driver'].tolist()
\end{lstlisting}
\end{tcolorbox}

\noindent \texttt{\footnotesize NLQ 3: what is the total number of points india has scored throughout the \\ rivalry?}

\begin{tcolorbox}[colback=white, colframe=black, arc=2mm, boxrule=0.5pt]
\scriptsize
\begin{lstlisting}[style=arxivcode,language=python]
valid_results = df['Results'].dropna().str.replace(
                    'Draw', '0 - 0')
scores = valid_results.str.extract(
                    r'(\d+)\s*-\s*(\d+)')
scores = scores.fillna(0).astype(int)
india_scores = scores[1]
total_points_india = india_scores.sum()
result = [[total_points_india]]
\end{lstlisting}
\end{tcolorbox}

\noindent \texttt{\footnotesize NLQ 4: how long did they play before they won a game?}

\begin{tcolorbox}[colback=white, colframe=black, arc=2mm, boxrule=0.5pt]
\scriptsize
\begin{lstlisting}[style=arxivcode,language=python]
first_game_date = pd.to_datetime(df['Date'].iloc[0])
wins = df[df['Result'].str.startswith('W')]
if not wins.empty:
    first_win_date = pd.to_datetime(
                    wins['Date'].iloc[0])
    days_before_first_win = (first_win_date
                    - first_game_date).days
else:
    days_before_first_win = 0
result = [[days_before_first_win]]
\end{lstlisting}
\end{tcolorbox}

\noindent \texttt{\footnotesize NLQ 5: how far removed were the times from the fifth place and the tenth\\ place contestant?}

\begin{tcolorbox}[colback=white, colframe=black, arc=2mm, boxrule=0.5pt]
\scriptsize
\begin{lstlisting}[style=arxivcode,language=python]
print(df['Rank'].unique())

# Creating a boolean series for rank 10
rank_5 = df['Rank'].astype(int).eq(5)

fifth_place_time = df.loc[rank_5, 'Time'].values[0]
      if df.loc[rank_5, 'Time'].size > 0 else None

tenth_place_time = df.loc[rank_10, 'Time'].values[0]
      if df.loc[rank_10, 'Time'].size > 0 else None

if fifth_place_time is None
    or tenth_place_time is None:
    result = [['Times not found for
                specified ranks.']]
else:
    def convert_to_seconds(time_str):
        minutes, seconds = time_str.split(':')
        return int(minutes) * 60 + float(seconds)

    fifth_place_seconds = convert_to_seconds(
                            fifth_place_time)
    tenth_place_seconds = convert_to_seconds(
                            tenth_place_time)

    difference = abs(fifth_place_seconds
                        - tenth_place_seconds)
    result = [[difference]]
\end{lstlisting}
\end{tcolorbox}

\noindent \texttt{\footnotesize NLQ 6: how many times did the cowboys win the nfc championship in the 1970s?}

\begin{tcolorbox}[colback=white, colframe=black, arc=2mm, boxrule=0.5pt]
\scriptsize
\begin{lstlisting}[style=arxivcode,language=python]
nfc_cowboys_70s = df[
    (df['Team'] == 'Dallas Cowboys')
    & (df['Season'] >= 1970) & (df['Season'] < 1980)
]
result = [[nfc_cowboys_70s.shape[0]]]
\end{lstlisting}
\end{tcolorbox}

\noindent \texttt{\footnotesize NLQ 7: who was the only administrator to have just a b.s.?}

\begin{tcolorbox}[colback=white, colframe=black, arc=2mm, boxrule=0.5pt]
\scriptsize
\begin{lstlisting}[style=arxivcode,language=python]
filtered_df = df[
    df['Educational Background'].str.contains(
    'B.S.') &
    ~df['Educational Background'].str.contains(
    'Ph.D|M.S|M.A|Master')
]
result = [[filtered_df['Name'].iloc[0]]]
         if not filtered_df.empty else [[]]
\end{lstlisting}
\end{tcolorbox}


\section{More Examples of Knowledge Transfer}
\label{sec:transfer_examples}

Section \ref{sec:motivation} highlights that different query languages originating from diverse structured knowledge sources can formulate semantically equivalent questions. Additional examples below further strengthen this point.

\noindent \texttt{\footnotesize NLQ 1: find all movies directed by Christopher Nolan or starring Leonardo \\DiCaprio.}

\begin{tcolorbox}[colback=white, colframe=black, arc=2mm, boxrule=0.5pt]
\scriptsize
\begin{lstlisting}[style=arxivcode,language=sql]
SELECT DISTINCT m.title
FROM Movies AS m
LEFT JOIN Directors AS d
ON m.director_id = d.director_id
LEFT JOIN Movie_Actors AS ma
ON m.movie_id = ma.movie_id
LEFT JOIN Actors AS a ON ma.actor_id = a.actor_id
WHERE d.name = 'Christopher Nolan'
OR a.name = 'Leonardo DiCaprio';
\end{lstlisting}
\end{tcolorbox}
\begin{tcolorbox}[colback=white, colframe=black, arc=2mm, boxrule=0.5pt]
\scriptsize
\begin{lstlisting}[style=arxivcode,language={}]
PREFIX dbo: <http://dbpedia.org/ontology/>
PREFIX dbr: <http://dbpedia.org/resource/>
PREFIX rdfs: <http://www.w3.org/2000/01/rdf-schema#>
SELECT DISTINCT ?title
WHERE {
  {
    ?movie dbo:director dbr:Christopher_Nolan ;
           rdfs:label ?title .
    FILTER (LANG(?title) = "en")
  }
  UNION
  {
    ?movie dbo:starring dbr:Leonardo_DiCaprio ;
           rdfs:label ?title .
    FILTER (LANG(?title) = "en")
  }
}
\end{lstlisting}
\end{tcolorbox}
This example demonstrates handling multiple relationships (director, actor) and logical disjunction (\texttt{OR}), which often translates to \texttt{UNION} in SPARQL or \texttt{OR} in SQL's \texttt{WHERE} clause.

\noindent \texttt{\footnotesize NLQ 2: list all research papers and, if available, their corresponding PDF URLs.}

\begin{tcolorbox}[colback=white, colframe=black, arc=2mm, boxrule=0.5pt]
\scriptsize
\begin{lstlisting}[style=arxivcode,language=sql]
SELECT p.title, pdf.url
FROM Papers AS p
LEFT JOIN Paper_PDFs AS pdf
ON p.paper_id = pdf.paper_id;
\end{lstlisting}
\end{tcolorbox}
\begin{tcolorbox}[colback=white, colframe=black, arc=2mm, boxrule=0.5pt]
\scriptsize
\begin{lstlisting}[style=arxivcode,language={}]
PREFIX dbo: <http://dbpedia.org/ontology/>
PREFIX dbr: <http://dbpedia.org/resource/>
PREFIX rdfs: <http://www.w3.org/2000/01/rdf-schema#>
SELECT DISTINCT ?title
WHERE {
  {
    ?movie dbo:director dbr:Christopher_Nolan ;
           rdfs:label ?title .
    FILTER (LANG(?title) = "en")
  }
  UNION
  {
    ?movie dbo:starring dbr:Leonardo_DiCaprio ;
           rdfs:label ?title .
    FILTER (LANG(?title) = "en")
  }
}
\end{lstlisting}
\end{tcolorbox}
This example demonstrates querying for entities and their optional attributes with optional information (\texttt{LEFT JOIN} in SQL, \texttt{OPTIONAL} in SPARQL), which is crucial when dealing with incomplete knowledge graphs or databases.

\noindent \texttt{\footnotesize NLQ 3: find the grandparents of Charlie Brown.}

\begin{tcolorbox}[colback=white, colframe=black, arc=2mm, boxrule=0.5pt]
\scriptsize
\begin{lstlisting}[style=arxivcode,language=sql]
SELECT gp.person_name AS grandparent_name
FROM People AS c
JOIN Parent_Child_Relationship AS pc1
ON c.person_id = pc1.child_id
JOIN People AS p ON pc1.parent_id = p.person_id
JOIN Parent_Child_Relationship AS pc2
ON p.person_id = pc2.child_id
JOIN People AS gp ON pc2.parent_id = gp.person_id
WHERE c.person_name = 'Charlie Brown';
\end{lstlisting}
\end{tcolorbox}
\begin{tcolorbox}[colback=white, colframe=black, arc=2mm, boxrule=0.5pt]
\scriptsize
\begin{lstlisting}[style=arxivcode,language={}]
PREFIX ex: <http://example.org/ontology/>
PREFIX foaf: <http://xmlns.com/foaf/0.1/>
SELECT ?grandparentName
WHERE {
  dbr:Charlie_Brown ex:hasParent/ex:hasParent
                    ?grandparent .
  ?grandparent foaf:name ?grandparentName .
}
\end{lstlisting}
\end{tcolorbox}
This example illustrates how multi-hop relationships (finding a parent of a parent) are expressed. SPARQL's property paths offer a concise way to represent this, while SQL requires multiple joins. This highlights a common pattern in graph data.

\noindent \texttt{\footnotesize NLQ 4: what is the highest salary for an Engineer in New York?}

\begin{tcolorbox}[colback=white, colframe=black, arc=2mm, boxrule=0.5pt]
\scriptsize
\begin{lstlisting}[style=arxivcode,language={}]
# Table: Job_Postings
| JobTitle | Company     | Location | Salary |
|----------|-------------|----------|--------|
| Engineer | TechCorp    | New York | 120000 |
| Manager  | BizInc      | San Fran | 130000 |
| Engineer | InnovateLLC | New York | 135000 |
| Analyst  | AlphaCo     | Chicago  | 90000  |
| Engineer | TechCorp    | Boston   | 115000 |
# TableQA Operation:
    1. Identify the target table: 'Job_Postings'.
    2. Filter rows where 'JobTitle' is 'Engineer'
                AND 'Location' is 'New York'.
    3. From the filtered rows, find the maximum
                value in the 'Salary' column.
\end{lstlisting}
\end{tcolorbox}
\begin{tcolorbox}[colback=white, colframe=black, arc=2mm, boxrule=0.5pt]
\scriptsize
\begin{lstlisting}[style=arxivcode,language=sql]
SELECT MAX(Salary)
FROM Job_Postings
WHERE JobTitle = 'Engineer'
  AND Location = 'New York';
\end{lstlisting}
\end{tcolorbox}
\begin{tcolorbox}[colback=white, colframe=black, arc=2mm, boxrule=0.5pt]
\scriptsize
\begin{lstlisting}[style=arxivcode,language={}]
PREFIX ex: <http://example.org/ontology/>
PREFIX dbo: <http://dbpedia.org/ontology/>
PREFIX xsd: <http://www.w3.org/2001/XMLSchema#>
SELECT (MAX(?salaryValue) AS ?maxSalary)
WHERE {
  ?jobPosting ex:jobTitle "Engineer" ;
              dbo:location dbr:New_York ;
              ex:salary ?salary .
  BIND(REPLACE(STR(?salary), "[^0-9.]", "")
                AS ?salaryCleaned)
  BIND(xsd:decimal(?salaryCleaned) AS ?salaryValue)
}
\end{lstlisting}
\end{tcolorbox}
This example illustrates how an aggregation function (MAX) is applied to a subset of data defined by multiple conditions. TableQA first filters the table rows based on targeted values, and then identifying the maximum value from filtered rows. SQL achieves this by using the \texttt{MAX()} aggregate function directly in the \texttt{SELECT} statement, with conditions specified in the \texttt{WHERE} clause. SPARQL performs a similar operation by defining a graph pattern to bind the relevant salaries, filtering them with \texttt{FILTER} conditions, and then applying the \texttt{MAX()} aggregate in the \texttt{SELECT} expression. This highlights a fundamental data analysis pattern: isolating specific data points based on criteria and then computing a summary statistic over them.

\noindent \texttt{\footnotesize NLQ 5: what are the songs in the alubum Night at Opera by Quean?}

\begin{tcolorbox}[colback=white, colframe=black, arc=2mm, boxrule=0.5pt]
\scriptsize
\begin{lstlisting}[style=arxivcode,language={}]
# Table 1: Artists
    | ArtistID | ArtistName   |
    | :------- | :----------- |
    | 1        | Quean        |
    | 2        | Lead Zipelin |
# Table 2: Albums
    | AlbumID | AlbumTitle      | Ref | Release |
    | :------ | :-------------- | :-- | :------ |
    | 101     | Night at Opera  | 1   | 1975    |
    | 102     | Heart Attack    | 1   | 1974    |
    | 201     | Led Zeppelin IV | 2   | 1971    |
# Table 3: Songs
    | ID   | SongTitle         | Ref | Duration |
    | :--- | :---------------- | :-- | :------- |
    | 1001 | Pohemian Rhapsody | 101 | 05:50    |
    | 1002 | My Best Friend    | 101 | 02:50    |
    | 1003 | Killa Queen       | 102 | 03:00    |

# TableQA Operation:
    1. In 'Artists' table, find 'ArtistID' where
        'ArtistName' is 'Quean'. (Result: ArtistID=1)
    2. In 'Albums' table, filter rows where
        'Ref' is 1 AND 'AlbumTitle' is 'Night
       at Opera'. (Result: AlbumID=101)
    3. In 'Songs' table, filter rows where
            'Ref' is 101.
    4. Select 'SongTitle' from the resulting rows.
\end{lstlisting}
\end{tcolorbox}
\begin{tcolorbox}[colback=white, colframe=black, arc=2mm, boxrule=0.5pt]
\scriptsize
\begin{lstlisting}[style=arxivcode,language=sql]
SELECT s.SongTitle
FROM Songs AS s
INNER JOIN Albums AS al ON s.Ref = al.AlbumID
INNER JOIN Artists AS ar ON al.Ref = ar.ArtistID
WHERE ar.ArtistName = 'Quean'
  AND al.AlbumTitle = 'Night at Opera';
\end{lstlisting}
\end{tcolorbox}
\begin{tcolorbox}[colback=white, colframe=black, arc=2mm, boxrule=0.5pt]
\scriptsize
\begin{lstlisting}[style=arxivcode,language={}]
PREFIX dbo: <http://dbpedia.org/ontology/>
PREFIX mo: <http://purl.org/ontology/mo/>
PREFIX rdfs: <http://www.w3.org/2000/01/rdf-schema#>
PREFIX dbr: <http://dbpedia.org/resource/>
SELECT ?songTitle
WHERE {
  ?artist rdfs:label "Quean"@en .
  ?album dbo:artist ?artist ;
         rdfs:label "Night at Opera"@en ;
         mo:record ?record .
  ?record rdfs:label ?songTitle .
}
\end{lstlisting}
\end{tcolorbox}
This example demonstrates how information is retrieved by navigating through multiple related entities (or tables) while applying specific filtering conditions at different stages of this navigation. SQL explicitly uses \texttt{JOIN} clauses to connect related tables, combining these with \texttt{WHERE} clauses to filter based on artist name and album title. SPARQL achieves this by linking triple patterns: variables like ?album and ?artist bridge different statements, effectively traversing relationships, while \texttt{FILTERs} or direct matching in patterns apply the conditions. TableQA simulates this by requiring sequential lookups or joins. This highlights the core concept of relational joins or graph traversal, essential for combining and constraining information that is distributed across.

\noindent \texttt{\footnotesize NLQ 6: list all European countries by population in ascending order,\\ excluding Germany.}

\begin{tcolorbox}[colback=white, colframe=black, arc=2mm, boxrule=0.5pt]
\scriptsize
\begin{lstlisting}[style=arxivcode,language={}]
# Table: Country_Info
| CountryName | Continent  | Population | GDP  |
| :---------- | :--------- | :--------- | :--- |
| France      | Europe     | 65273511   | 2716 |
| Germany     | Europe     | 83783942   | 3846 |
| UK          | Europe     | 67886011   | 2827 |
| Italy       | Europe     | 59554023   | 1886 |
| Canada      | N. America | 37742154   | 1736 |
# TableQA Operation:
    1. Identify the target table: 'Country_Info'.
    2. Filter rows where 'Continent' is 'Europe'.
    3. Further filter out rows where 'CountryName'
            is 'Germany'.
    4. Select 'CountryName' and 'Population'.
    5. Order the results by 'Population'
            in ascending order.
\end{lstlisting}
\end{tcolorbox}

\begin{tcolorbox}[colback=white, colframe=black, arc=2mm, boxrule=0.5pt]
\scriptsize
\begin{lstlisting}[style=arxivcode,language=sql]
SELECT CountryName, Population
FROM Country_Info
WHERE Continent = 'Europe'
  AND CountryName != 'Germany'
ORDER BY Population ASC;
\end{lstlisting}
\end{tcolorbox}

\begin{tcolorbox}[colback=white, colframe=black, arc=2mm, boxrule=0.5pt]
\scriptsize
\begin{lstlisting}[style=arxivcode,language={}]
PREFIX dbo: <http://dbpedia.org/ontology/>
PREFIX dbr: <http://dbpedia.org/resource/>
PREFIX rdfs: <http://www.w3.org/2000/01/rdf-schema#>
SELECT ?countryName ?population
WHERE {
  ?country dbo:continent dbr:Europe ;
           dbo:populationTotal ?population ;
           rdfs:label ?countryName .
  FILTER (LANG(?countryName) = "en")
  FILTER (?country != dbr:Germany)
}
ORDER BY ASC(?population)
\end{lstlisting}
\end{tcolorbox}
This example showcases how datasets are refined by first applying inclusionary and exclusionary criteria (negation) and then structuring the presentation of the results through ordering. In a TableQA scenario, one would first filter the table for Europe in the continent column, then remove any rows where the country is Germany, and finally sort the remaining rows in ascending order. SQL handles this using a \texttt{WHERE} clause with multiple conditions, followed by an \texttt{ORDER BY} and \texttt{ASC} clause. SPARQL employs \texttt{FILTER} conditions within its graph pattern to achieve both the inclusion and exclusion. This illustrates the common data processing workflow of filtering (both to include and exclude data based on specific attributes) and then sorting the qualified data to facilitate analysis or presentation.

By showcasing these examples, it becomes clearer that despite the syntactic and structural differences between SQL databases, knowledge graphs, and tables, the underlying logical operations required to answer user questions can be very similar. A unified SKR framework would aim to abstract these common operations, allowing LLMs to learn them once and apply them across these diverse structured knowledge sources, thus effectively transferring knowledge and improving performance on SKR tasks such as text-to-SQL, KBQA, and TableQA tasks.

\section{Detailed Introduction to Compared Methods}
\label{sec:source_of_results}

The main experiment involved a comparative analysis of \textsc{Pandora} with 25 baselines categorized as:  \textbf{DB SKR}, \textbf{Table SKR}, \textbf{KG SKR} and \textbf{Unified SKR Methods}. Table \ref{tab:baseline_details} outlines the sources of the reported results for these baselines from their respective papers, and their principal ideas are summarized below.

\subsection{DB SKR (Text-to-SQL) Methods}
\begin{itemize}[leftmargin=1em]
\item PICARD~\cite{DBLP:conf/emnlp/ScholakSB21}: A constrained decoding approach for text-to-SQL that incrementally parses partial outputs to ensure syntactic SQL validity.

\item DIN-SQL~\cite{DBLP:conf/nips/PourrezaR23}: A decomposed in-context learning method for text-to-SQL that splits the generation into smaller sub-tasks and uses chain-of-thought prompting with self-correction.

\item NatSQL~\cite{DBLP:conf/aaai/Li00023}: An intermediate representation for text-to-SQL that simplifies SQL parsing by removing or rewriting hard clauses while preserving query semantics.

\item CodeS~\cite{DBLP:journals/pacmmod/LiZLFZZWP0024}: An open-source family of code-centric language models, specialized for text-to-SQL tasks.

\item DTS-SQL~\cite{DBLP:conf/emnlp/PourrezaR24}: A two-stage fine-tuning pipeline tailored for smaller language models, separating schema linking and SQL synthesis.

\item SQLFixAgent~\cite{DBLP:conf/aaai/CenLLW25}: A multi-agent framework to identify and rectify errors in initially generated SQL queries.

\item DAIL~\cite{DBLP:journals/pvldb/GaoWLSQDZ24}: A demonstration-based in-context learning approach that selects and formats examples for text-to-SQL, leveraging LLMs with carefully designed demonstration strategies.

\item Mac-SQL~\cite{DBLP:conf/coling/WangR0LBCYZYSL25}: A multi-agent collaborative framework for text-to-SQL that decomposes the task into specialized agent roles for schema linking, SQL generation, and self-correction.

\item  GenSQL~\cite{DBLP:conf/coling/ShiXLXC0WW25}: A generative approach for text-to-SQL that combines LLM-based SQL generation with execution feedback and self-refinement mechanisms.
\end{itemize}

\subsection{Table SKR (TableQA) Methods}
\begin{itemize}[leftmargin=1em]

\item TAPAS~\cite{DBLP:conf/acl/HerzigNMPE20}: A weakly supervised table question answering model that predicts answers by directly selecting a subset of table cells and, optionally, an aggregation operator (e.g., SUM, COUNT, AVERAGE).

\item TAPEX~\cite{DBLP:conf/iclr/LiuCGZLCL22}: A seq2seq table reasoning model pre-trained as a neural SQL execution engine.

\item Binder~\cite{DBLP:conf/iclr/ChengX0LNHXROZS23}: A training-free neural-symbolic framework designed for table question answering, which translates natural language questions into an executable program (e.g., SQL or Python) that is augmented with specialized Language Model (LM) API calls.

\item DATER~\cite{DBLP:conf/sigir/YeHYLHL23}: A few-shot prompting strategy for table QA that decomposes tables by extracting relevant sub-tables and breaks complex questions into simpler, step-by-step sub-questions answered on those sub-tables.

\item TIDE~\cite{DBLP:conf/iclr/YangDZDCDZ25}: A triple-guided method for LLM-based table QA, which employs a triple-based structure (head entity, relation, tail entity) to guide both question decomposition and answer verification.

\end{itemize}

\subsection{KG SKR (KGQA) Methods}
\begin{itemize}[leftmargin=1em]

\item RnG-KBQA~\cite{DBLP:conf/acl/YeYHZX22}: A hybrid Rank-and-Generate approach for knowledge-base QA that first retrieves a set of candidate logical forms and then refines them.

\item DecAF~\cite{DBLP:conf/iclr/YuZNZL0HWWX23}: A joint decoding framework for knowledge-base QA that concurrently generates both a direct textual answer and a logical form, and subsequently validate or correct the direct answer using the execution result of the logical form.

\item TIARA~\cite{DBLP:conf/emnlp/ShuYLKMQL22}: A knowledge-base QA model that enhances pre-trained language mode performance, by employing a multi-grained retrieval strategy to supply relevant KB context.

\item KB-Binder~\cite{DBLP:conf/acl/LiMZGSC23}: A training-free, few-shot framework for knowledge-base QA that first generates preliminary logical forms, subsequently grounded onto the knowledge base by a Knowledge Base Binder module.

\item KB-Coder~\cite{DBLP:conf/aaai/NieZW024}: A training-free KBQA framework using a code-style in-context learning method to minimize formatting errors in logical forms.

\item KBQA-o1~\cite{DBLP:journals/corr/abs-2501-18922}: An agentic KBQA method that employs Monte Carlo Tree Search (MCTS) to facilitate stepwise logical form generation through exploration of the knowledge base environment.

\item SG-KBQA~\cite{DBLP:conf/emnlp/GaoLQ25}: A subgraph-guided KBQA approach that leverages retrieved subgraph structures to constrain and guide logical form generation, achieving strong performance on complex multi-hop questions.

\end{itemize}

\subsection{Unified SKR Methods:}
\begin{itemize}[leftmargin=1em]
\item StructLM-7B~\cite{DBLP:journals/corr/abs-2402-16671}: A generalist language model instruction-tuned on a broad mixture of structured data tasks (text-to-SQL, table QA, KG QA, etc.).

\item UnifiedSKG~\cite{DBLP:conf/emnlp/XieW0ZSYWZYWZWL22}: A unified text-to-text framework that recasts 21 structured knowledge tasks (spanning SQL, tables, and graphs) into a standardized sequence format for seq2seq models.

\item StructGPT~\cite{DBLP:conf/emnlp/JiangZDYZW23}: An LLM-augmented QA framework that iteratively reads and reasons over structured data by invoking external tools.

\item Readi~\cite{DBLP:conf/acl/ChengZXYZQHCL0R24}: A reasoning-path editing framework that guides an LLM to construct and iteratively refine a step-by-step reasoning plan.

\item TrustUQA~\cite{DBLP:journals/corr/abs-2406-18916}: A unified question-answering system that ensures trustworthy answers by translating structured data into a general graph-based format, over which an LLM generates pseudo-queries that is automatically converted into executable queries to retrieve answers.

\end{itemize}

\begin{table*}
{\caption{The specific table number from which the reported results are selected within the original baseline papers. Here - denotes either that the dataset was not included in the cited work or that the reported metric differs from the metric employed in this paper. Specifically, \cite{DBLP:journals/pacmmod/LiZLFZZWP0024}-7 and \cite{DBLP:conf/iclr/YuZNZL0HWWX23}-5 indicate results taken from Table 7 of \cite{DBLP:journals/pacmmod/LiZLFZZWP0024} and Table 5 of \cite{DBLP:conf/iclr/YuZNZL0HWWX23}, respectively.}\label{tab:baseline_details}}
	\begin{center}
	\scalebox{1.0}{
		\begin{tabular}{lcccccc}
\toprule
\multicolumn{1}{l}{\multirow{2}[1]{*}{\textbf{Method}}}
        &\multicolumn{2}{c}{\textbf{DB SKR}} &\multicolumn{2}{c}{\textbf{Table SKR}} &\multicolumn{2}{c}{\textbf{KG SKR}} \\
        \cmidrule(lr){2-3} \cmidrule(lr){4-5} \cmidrule(lr){6-7}
        & \textbf{Spider}
            & \textbf{S-SYN}
            & \textbf{WTQ}
            & \textbf{WikiSQL}
            & \textbf{GrailQA}
            & \textbf{WebQSP} \\
        \cmidrule(lr){1-7}
\multicolumn{7}{c}{\textit{Single-type SKR Methods}} \\
\cmidrule(lr){1-7}
        PICARD~\cite{DBLP:conf/emnlp/ScholakSB21} & 1 & \cite{DBLP:journals/pacmmod/LiZLFZZWP0024}-7 & - & - & - & -  \\
        DIN-SQL~\cite{DBLP:conf/nips/PourrezaR23} & 4 & - & - & - & - & -  \\
        NatSQL~\cite{DBLP:conf/aaai/Li00023} & 2 & 3 & - & - & - & -  \\
        CodeS~\cite{DBLP:journals/pacmmod/LiZLFZZWP0024} & 5 & 7 & - & - & - & - \\
        DTS-SQL~\cite{DBLP:conf/emnlp/PourrezaR24} & 3 & 7 & - & - & - & - \\
        SQLFixAgent~\cite{DBLP:conf/aaai/CenLLW25} & 2 & 3 & - & - & - & - \\
        DAIL~\cite{DBLP:journals/pvldb/GaoWLSQDZ24} & 3 & - & - & - & - & - \\
        Mac-SQL~\cite{DBLP:conf/coling/WangR0LBCYZYSL25} & 2 & - & - & - & - & - \\
        GenSQL~\cite{DBLP:conf/coling/ShiXLXC0WW25} & 3 & - & - & - & - & - \\
        TAPAS~\cite{DBLP:conf/acl/HerzigNMPE20} & - & -  &  4 & 3 & - & - \\
        TAPEX~\cite{DBLP:conf/iclr/LiuCGZLCL22} & - & - &  2 & 1 & - & -  \\
        Binder~\cite{DBLP:conf/iclr/ChengX0LNHXROZS23} & - & - & 1 & - & - & -  \\
        DATER~\cite{DBLP:conf/sigir/YeHYLHL23} & - & - & 3 & - & - & - \\
        TIDE~\cite{DBLP:conf/iclr/YangDZDCDZ25} & - & - & 1 & - & - & - \\
        RnG-KBQA~\cite{DBLP:conf/acl/YeYHZX22} & - & - & - & - &\cite{DBLP:conf/iclr/YuZNZL0HWWX23}-5 &- \\
        DecAF~\cite{DBLP:conf/iclr/YuZNZL0HWWX23} & - & - & - & - &2 &5 \\
        TIARA~\cite{DBLP:conf/emnlp/ShuYLKMQL22} & - & - & - & - &3 &2 \\
        KB-Binder~\cite{DBLP:conf/acl/LiMZGSC23} & - & - & - & - &2  &3 \\
        KB-Coder~\cite{DBLP:conf/aaai/NieZW024} & - & - & - & - &2  &3 \\
        KBQA-o1~\cite{DBLP:journals/corr/abs-2501-18922} & - & - & - & - &2  &3 \\
        SG-KBQA~\cite{DBLP:conf/emnlp/GaoLQ25} & - & - & - & - & 3  & 3 \\
\cmidrule(lr){1-7}
\multicolumn{7}{c}{\textit{Unified SKR Methods}} \\
\cmidrule(lr){1-7}
        StructLM-7B~\cite{DBLP:journals/corr/abs-2402-16671} & 2 & - & 2 & 2 & - & -  \\
        UnifiedSKG~\cite{DBLP:conf/emnlp/XieW0ZSYWZYWZWL22} & 11 & - & 2 & 2 & - & 2 \\
        StructGPT~\cite{DBLP:conf/emnlp/JiangZDYZW23} & 3 & 3  & 2 & 2 & - & 1  \\
        Readi~\cite{DBLP:conf/acl/ChengZXYZQHCL0R24} & - & - & 2 & 2 & - & 1 \\
        TrustUQA~\cite{DBLP:journals/corr/abs-2406-18916} & - & - & 2 & 2 & - & 3 \\
\bottomrule
\end{tabular}
		}
	\end{center}
\end{table*}

\section{Case Studies}
\label{sec:bad_cases}
\subsection{Execution Failure}

\noindent \textbf{NLQ:}
\noindent \texttt{\footnotesize What is the xbox 360 kinect game version that supports the north american \\continent region?}

\noindent \textbf{Gold Query (S-expression):}
\begin{tcolorbox}[colback=white, colframe=black, arc=2mm, boxrule=0.5pt]
\scriptsize
\begin{lstlisting}[style=arxivcode,language=sql]
(AND game_version (AND (
    JOIN game_version.regions north \
         american continent)
    (JOIN (R computer_peripheral.supporting_game\
             _versions)
    xbox 360 kinect)))
\end{lstlisting}
\end{tcolorbox}

\noindent \textbf{Incorrect Reasoning Steps:}
\begin{tcolorbox}[colback=white, colframe=black, arc=2mm, boxrule=0.5pt]
\scriptsize
\begin{lstlisting}[style=arxivcode,language={}]
To answer the question, we first need to filter the
'computer_peripheral' DataFrame to find the Xbox 360
Kinect. Then, we cross-reference its supported game
versions with the 'game_version' DataFrame to find
which of these game versions support the 'north
american continent' region. We can achieve this using
a merge operation to join these two DataFrames based
on the game versions and filter for the region
specified. Finally, we will store the resulting game
versions in the desired list format.
\end{lstlisting}
\end{tcolorbox}

\noindent \textbf{Incorrect Code:}
\begin{tcolorbox}[colback=white, colframe=black, arc=2mm, boxrule=0.5pt]
\scriptsize
\begin{lstlisting}[style=arxivcode,language=python]
kinect_games = computer_peripheral[
    computer_peripheral['computer_peripheral']
                        == 'xbox 360 kinect'
                ]['supporting_game_versions'].values
if len(kinect_games) > 0:
    kinect_games = kinect_games[0]
    supported_versions = game_version[
    game_version['game_version'].isin(kinect_games)
    & (game_version['regions'] == 'north american
        continent')]
    result = supported_versions[[
            'game_version']].values.tolist()
else:
    result = []
\end{lstlisting}
\end{tcolorbox}

\noindent \textbf{Runtime Error:}
\begin{tcolorbox}[colback=white, colframe=black, arc=2mm, boxrule=0.5pt]
\scriptsize
\begin{lstlisting}[style=arxivcode,language={}]
Execution failed on python:
TypeError: only list-like objects are allowed to be
passed to isin(), you passed a `str`",
"Traceback (most recent call last):
File \"/tmp/tmpwoi1e79t.py\", line 19, in <module>
supported_versions = game_version[game_version['
game_version'].isin(kinect_games) &
(game_version['regions'] == 'north
american continent')]
\end{lstlisting}
\end{tcolorbox}

The incorrect code fails due to a TypeError in pandas' isin() method, as it passes a single string (from extracting kinect\_games[0]) instead of a list-like object, assuming only one entry exists for Xbox 360 Kinect.
This stems from flawed reasoning that oversimplifies data handling without properly checking if \texttt{supporting\_game\_versions} is a list, mismatching the gold S-expression's structured JOIN and AND logic for filtering regions and game versions.

\subsection{BOX Error}

\noindent \texttt{\footnotesize How many battles did not lose any ship with tonnage '225'?}

\noindent \textbf{Gold Query (SQL):}
\begin{tcolorbox}[colback=white, colframe=black, arc=2mm, boxrule=0.5pt]
\scriptsize
\begin{lstlisting}[style=arxivcode,language=sql]
SELECT count(*) FROM battle WHERE id NOT IN (
    SELECT lost_in_battle FROM ship
    WHERE tonnage  =  '225'
);
\end{lstlisting}
\end{tcolorbox}

\noindent \textbf{Incorrect Reasoning Steps:}
\begin{tcolorbox}[colback=white, colframe=black, arc=2mm, boxrule=0.5pt]
\scriptsize
\begin{lstlisting}[style=arxivcode,language={}]
To determine how many battles did not lose any ship
with tonnage '225', we need to follow these steps: 1.
Filter the ships that have a tonnage of '225' and
check if any of them lost the battle by joining with
the battle table using the foreign key. 2. Identify
battles where there are no ships with tonnage '225'
that were lost. 3. Count the number of unique battles
that fit this criterion. This requires using a
combination of filtering and deduplication techniques
provided by pandas.
\end{lstlisting}
\end{tcolorbox}

\noindent \textbf{Incorrect Code:}
\begin{tcolorbox}[colback=white, colframe=black, arc=2mm, boxrule=0.5pt]
\scriptsize
\begin{lstlisting}[style=arxivcode,language=python]
ships_with_lost_tonnage_225 = ship[ship['tonnage']
                                == 225]['id']
battles_with_losses = battle[battle['id'].isin(
        death[death['caused_by_ship_id'].isin(
        ships_with_lost_tonnage_225)]['id'])]
battles_without_losses = battle[~battle['id'].isin(
            battles_with_losses['id'])]
result = [(battles_without_losses['id'].nunique(),)]
\end{lstlisting}
\end{tcolorbox}

\noindent \textbf{Gold Code:}
\begin{tcolorbox}[colback=white, colframe=black, arc=2mm, boxrule=0.5pt]
\scriptsize
\begin{lstlisting}[style=arxivcode,language=python]
stadium['capacity'] = pd.to_numeric(stadium[
                    'capacity'], errors='coerce')
max_capacity = stadium['capacity'].max()
result = [(max_capacity, stadium['average'])]
\end{lstlisting}
\end{tcolorbox}
The incorrect code exhibits multiple compounding errors that collectively produce a semantically and structurally invalid query. Most critically, it fabricates a non-existent \texttt{death} table alongside a spurious \texttt{caused\_by\_ship\_id} column, introducing fictitious schema elements that yield flawed join operations entirely misaligned with the gold SQL's straightforward subquery logic --- namely, filtering \texttt{ship.lost\_in\_battle} where \texttt{tonnage = '225'} and directly counting the resulting battle IDs.
Beyond this hallucinated schema, the code further commits a type inconsistency by treating \texttt{tonnage} as an integer value \texttt{225} rather than the string literal \texttt{'225'}, thereby risking silent type mismatch errors that may produce empty or incorrect results depending on the underlying engine's coercion behavior. Compounding these issues, the final output is incorrectly formulated as a tuple of unique battle IDs rather than a scalar count, deviating from both the structural and semantic expectations of the NLQ and yielding a result that is neither directly interpretable nor consistent with the intended answer.

\subsection{Field Error}
\noindent \textbf{NLQ:}

\noindent \texttt{\footnotesize What is the maximum capacity and the average of all stadiums ?}

\noindent \textbf{Gold Query (SQL):}
\begin{tcolorbox}[colback=white, colframe=black, arc=2mm, boxrule=0.5pt]
\scriptsize
\begin{lstlisting}[style=arxivcode,language=sql]
select max(capacity), average from stadium
\end{lstlisting}
\end{tcolorbox}

\noindent \textbf{Incorrect Reasoning Steps:}
\begin{tcolorbox}[colback=white, colframe=black, arc=2mm, boxrule=0.5pt]
\scriptsize
\begin{lstlisting}[style=arxivcode,language={}]
To determine the maximum capacity and the average
capacity of all stadiums, I will first convert the
'capacity' column of the stadium DataFrame to numeric
format in order to perform mathematical calculations.
After that, I will calculate both the maximum and
average capacities. Finally, I will store the results
in a list of tuples format for the final output.
\end{lstlisting}
\end{tcolorbox}

\noindent \textbf{Incorrect Code:}
\begin{tcolorbox}[colback=white, colframe=black, arc=2mm, boxrule=0.5pt]
\scriptsize
\begin{lstlisting}[style=arxivcode,language=python]
stadium['capacity'] = pd.to_numeric(
        stadium['capacity'], errors='coerce')
max_capacity = stadium['capacity'].max()
avg_capacity = stadium['capacity'].mean()
result = [(max_capacity, avg_capacity)]
\end{lstlisting}
\end{tcolorbox}

\noindent \textbf{Gold Code:}
\begin{tcolorbox}[colback=white, colframe=black, arc=2mm, boxrule=0.5pt]
\scriptsize
\begin{lstlisting}[style=arxivcode,language=python]
stadium['capacity'] = pd.to_numeric(
            stadium['capacity'], errors='coerce')
max_capacity = stadium['capacity'].max()
result = [(max_capacity, stadium['average'])]
\end{lstlisting}
\end{tcolorbox}
The incorrect code introduces a critical field error by computing the mean of the \texttt{capacity} column via \texttt{stadium['capacity'].mean()}, fundamentally misinterpreting the term \textit{``average''} in the NLQ as an aggregation operation rather than recognizing it as a pre-existing column in the stadium DataFrame, a distinction made explicit in both the gold SQL and the reference code.
This error originates from flawed schema grounding, wherein the model presupposes that \texttt{average} must be derived through explicit calculation, entirely overlooking the fact that \texttt{average} is a native field in the schema. Such an assumption is particularly hazardous when the \texttt{average} column encodes domain-specific precomputed statistics, for instance, historical attendance averages, that are semantically distinct from a raw \texttt{capacity} mean, thereby rendering the incorrect output not merely redundant but potentially misleading.

\subsection{Reasoning Logic Error}

\noindent \textbf{NLQ:}

\noindent \texttt{\footnotesize The successor agency of national aeronautic and space administration was \\which government agency?}

\noindent \textbf{Gold Query (S-expression):}
\begin{tcolorbox}[colback=white, colframe=black, arc=2mm, boxrule=0.5pt]
\scriptsize
\begin{lstlisting}[style=arxivcode,language=sql]
(AND government_agency (JOIN
government_agency.successor_agency
national aeronautic and space administration))
\end{lstlisting}
\end{tcolorbox}

\noindent \textbf{Incorrect Reasoning Steps:}
\begin{tcolorbox}[colback=white, colframe=black, arc=2mm, boxrule=0.5pt]
\scriptsize
\begin{lstlisting}[style=arxivcode,language={}]
First, I need to locate the 'national aeronautic and
space administration' in the 'government_agency'
DataFrame to find its associated 'successor_agency'. I
will filter the DataFrame to get the row that matches
this government agency, and then access the
'successor_agency' column to find the successor.
\end{lstlisting}
\end{tcolorbox}

\noindent \textbf{Incorrect Code:}
\begin{tcolorbox}[colback=white, colframe=black, arc=2mm, boxrule=0.5pt]
\scriptsize
\begin{lstlisting}[style=arxivcode,language=python]
successor_agency = government_agency[
    government_agency['government_agency'] ==
    'national aeronautic and space administration'
    ]['successor_agency'].iloc[0]
result = [[successor_agency]]
\end{lstlisting}
\end{tcolorbox}

\noindent \textbf{Gold Code:}
\begin{tcolorbox}[colback=white, colframe=black, arc=2mm, boxrule=0.5pt]
\scriptsize
\begin{lstlisting}[style=arxivcode,language=python]
successor_agency = government_agency[
    government_agency['successor_agency'] ==
    'national aeronautic and space administration'
    ]['government_agency'].iloc[0]
result = [[successor_agency]]
\end{lstlisting}
\end{tcolorbox}

The incorrect reasoning inverts the successor-predecessor relationship by filtering on \texttt{government\_agency = `national aeronautic and space administration'} to retrieve its successor, whereas the NLQ explicitly seeks the \textit{predecessor} agency that was subsequently succeeded by NASA. The correct approach requires filtering on \texttt{successor\_agency = `nasa'} and retrieving the corresponding \texttt{government\_agency} value.
This logic error stems from a fundamental misinterpretation of the schema's directional semantics: the model erroneously assumes that \texttt{successor\_agency} emanates \textit{from} the queried entity, when in fact it points \textit{toward} it --- thereby reversing the intended traversal direction and yielding a result that is semantically opposite to what the NLQ demands.

\subsection{Query Intent Error}

\noindent \textbf{NLQ:}

\noindent \texttt{\footnotesize Which airlines have less than 200 flights?}

\noindent \textbf{Gold Query (SQL):}
\begin{tcolorbox}[colback=white, colframe=black, arc=2mm, boxrule=0.5pt]
\scriptsize
\begin{lstlisting}[style=arxivcode,language=sql]
SELECT T1.Airline
FROM AIRLINES AS T1
JOIN FLIGHTS AS T2
ON T1.uid  =  T2.Airline
GROUP BY T1.Airline
HAVING count(*)  <  200
\end{lstlisting}
\end{tcolorbox}

\noindent \textbf{Incorrect Reasoning Steps:}
\begin{tcolorbox}[colback=white, colframe=black, arc=2mm, boxrule=0.5pt]
\scriptsize
\begin{lstlisting}[style=arxivcode,language={}]
In light of the previous responses, the goal is to
see the complete list of airline counts rather than
filtering out those with flight counts under 200
initially. This enables us to understand what data we
have when we check the counts. Once confirmed, I can
collect how many airlines have at least 200 flights
or display all available airlines with their counts,
allowing better analysis of their distribution.
\end{lstlisting}
\end{tcolorbox}

\noindent \textbf{Incorrect Code:}
\begin{tcolorbox}[colback=white, colframe=black, arc=2mm, boxrule=0.5pt]
\scriptsize
\begin{lstlisting}[style=arxivcode,language=python]
# Step 1: Group the flights by airline and count the
# number of flights.
flight_counts = flights.groupby('airline').size(
                 ).reset_index(name='flight_count')
# Step 2: Prepare the final result as a list of tuples
# containing airline names and their flight counts.
result = list(flight_counts.itertuples(
                index=False, name=None))
\end{lstlisting}
\end{tcolorbox}
The incorrect code misinterprets the query intent by computing and outputting flight counts for \textbf{all} airlines, rather than isolating only those with fewer than 200 flights. Since the NLQ explicitly asks \textit{which airlines have less than 200 flights}, this overbroad result introduces unnecessary data that obscures the intended answer.
This error originates from flawed reasoning that prioritizes displaying a complete frequency distribution for exploratory purposes, thereby overlooking the gold SQL's focused logic --- namely, grouped aggregation, a \texttt{HAVING COUNT(*) < 200} filter, and the selection of airline names exclusively.
A straightforward fix would involve applying a threshold filter via \texttt{flight\_counts[flight\_counts['flight\_count'] < 200]} and extracting only the airline name column as the final output. If the aggregated DataFrame lacks readable airline names, an additional merge with a reference airlines DataFrame may be required to align with the intended result.

\section{Complexity Analysis of KG-to-BOX}

\subsection{Notation}
Let $|\mathcal{E}^*|$ denote the number of topic entities, $|\mathcal{R}^*|$ the number of relevant relations after filtering, $H$ the maximum hop depth, $d$ the average number of neighbors per entity per relation, $N$ the maximum number of fields per BOX, and $M$ the maximum number of rows per BOX.

\subsection{Phase 1: Depth-First Search}

For each topic entity $e \in \mathcal{E}^*$, \textsc{DepthFirstSearch} is invoked with hop limit $H$.
At each recursive call, the algorithm iterates over all $|\mathcal{R}^*|$ relations and, for each relation, fetches both forward neighbors $\mathcal{E}^+$ and backward neighbors $\mathcal{E}^-$, yielding an average branching factor of $B = 2|\mathcal{R}^*|d$.
The visited-entity pruning ($e^\pm \notin \mathcal{V}$) prevents revisiting entities, so the recursion tree has depth $H$ and at most $B^H$ leaf nodes.
The total number of recursive calls is therefore $\sum_{h=0}^{H} B^h = O(B^H)$, giving:
\begin{equation}
    T_{\text{Phase 1}} = O\!\left(|\mathcal{E}^*| \cdot (2|\mathcal{R}^*|d)^H\right).
\end{equation}
For space, the recursion stack holds at most $H$ frames, each carrying a path $\omega$ of length $O(H)$, contributing $O(H^2)$ stack space.
The path list $\Omega$ accumulates all leaf-level paths, each of length $2H$, so:
\begin{equation}
    S_{\text{Phase 1}} = O\!\left(|\mathcal{E}^*| \cdot (2|\mathcal{R}^*|d)^H \cdot H\right).
\end{equation}

\subsection{Phase 2: BOX Field Registration}

The algorithm iterates over every recorded path $\omega \in \Omega$ and every triple $(b, \phi, \psi)$ within $\omega$ (of length $2H$) to register field names $\Phi_b$.
Since $|\Omega| = O(|\mathcal{E}^*| \cdot (2|\mathcal{R}^*|d)^H)$, the time complexity of this phase is:
\begin{equation}
    T_{\text{Phase 2}} = O\!\left(|\Omega| \cdot H\right) = O\!\left(|\mathcal{E}^*| \cdot (2|\mathcal{R}^*|d)^H \cdot H\right).
\end{equation}
The space required to store all field name sets $\{\Phi_b\}$ is bounded by $O(|\Omega| \cdot H)$, which is the same order as above.

\subsection{Phase 3: BOX Value Filling}

For each path $\omega \in \Omega$ and each triple $(b, \phi, \psi)$, the algorithm appends the value $\psi$ to $\Psi_b^\phi$ and pads all remaining $N - 1$ fields with \texttt{NA}.
The cost per triple is thus $O(N)$, yielding:
\begin{equation}
    T_{\text{Phase 3}} = O\!\left(|\Omega| \cdot H \cdot N\right) = O\!\left(|\mathcal{E}^*| \cdot (2|\mathcal{R}^*|d)^H \cdot H \cdot N\right).
\end{equation}
The space required to store all BOX value arrays is $O(|\mathcal{B}^*| \cdot N \cdot M)$, where $|\mathcal{B}^*|$ is the total number of constructed BOXes.

\subsection{Overall Complexity}

The overall time and space complexities are dominated by Phase 3:
\begin{equation}
    T_{\text{total}} = O\!\left(|\mathcal{E}^*| \cdot (2|\mathcal{R}^*|d)^H \cdot H \cdot N\right),
\end{equation}
\begin{equation}
    S_{\text{total}} = O\!\left(|\mathcal{E}^*| \cdot (2|\mathcal{R}^*|d)^H \cdot H \cdot N\right).
\end{equation}

\subsection{Practical Tractability}

Although the worst-case complexity is exponential in $H$, several design choices render the algorithm tractable in practice.
\textbf{(i) Relation pruning}: retaining only $\mathcal{R}^* \subseteq \mathcal{R}$ via embedding similarity significantly reduces the branching factor $B$.
\textbf{(ii) Visited-entity pruning}: the constraint $e^\pm \notin \mathcal{V}$ eliminates cycles and redundant paths, further contracting the effective search tree.
\textbf{(iii) Small hop depth}: KGQA benchmarks such as WebQSP and ComplexWebQuestions require at most $H = 2$ hops, keeping $(2|\mathcal{R}^*|d)^H$ manageable.
\textbf{(iv) Few topic entities}: each natural language question typically mentions only one or two topic entities, so $|\mathcal{E}^*|$ is effectively a small constant.


\end{document}

%% file: main.bbl

%% file: main.bbl
\begin{thebibliography}{10}
\providecommand{\url}[1]{#1}
\csname url@samestyle\endcsname
\providecommand{\newblock}{\relax}
\providecommand{\bibinfo}[2]{#2}
\providecommand{\BIBentrySTDinterwordspacing}{\spaceskip=0pt\relax}
\providecommand{\BIBentryALTinterwordstretchfactor}{4}
\providecommand{\BIBentryALTinterwordspacing}{\spaceskip=\fontdimen2\font plus
\BIBentryALTinterwordstretchfactor\fontdimen3\font minus
  \fontdimen4\font\relax}
\providecommand{\BIBforeignlanguage}[2]{{%
\expandafter\ifx\csname l@#1\endcsname\relax
\typeout{** WARNING: IEEEtran.bst: No hyphenation pattern has been}%
\typeout{** loaded for the language `#1'. Using the pattern for}%
\typeout{** the default language instead.}%
\else
\language=\csname l@#1\endcsname
\fi
#2}}
\providecommand{\BIBdecl}{\relax}
\BIBdecl

\bibitem{DBLP:conf/acl/PasupatL15}
\BIBentryALTinterwordspacing
P.~Pasupat and P.~Liang, ``Compositional semantic parsing on semi-structured
  tables,'' in \emph{Proceedings of the 53rd Annual Meeting of the Association
  for Computational Linguistics and the 7th International Joint Conference on
  Natural Language Processing of the Asian Federation of Natural Language
  Processing, {ACL} 2015, July 26-31, 2015, Beijing, China, Volume 1: Long
  Papers}.\hskip 1em plus 0.5em minus 0.4em\relax The Association for Computer
  Linguistics, 2015, pp. 1470--1480. [Online]. Available:
  \url{https://doi.org/10.3115/v1/p15-1142}
\BIBentrySTDinterwordspacing

\bibitem{DBLP:conf/emnlp/YuZYYWLMLYRZR18}
\BIBentryALTinterwordspacing
T.~Yu, R.~Zhang, K.~Yang, M.~Yasunaga, D.~Wang, Z.~Li, J.~Ma, I.~Li, Q.~Yao,
  S.~Roman, Z.~Zhang, and D.~R. Radev, ``Spider: {A} large-scale human-labeled
  dataset for complex and cross-domain semantic parsing and text-to-sql task,''
  in \emph{Proceedings of the 2018 Conference on Empirical Methods in Natural
  Language Processing, Brussels, Belgium, October 31 - November 4, 2018},
  E.~Riloff, D.~Chiang, J.~Hockenmaier, and J.~Tsujii, Eds.\hskip 1em plus
  0.5em minus 0.4em\relax Association for Computational Linguistics, 2018, pp.
  3911--3921. [Online]. Available: \url{https://doi.org/10.18653/v1/d18-1425}
\BIBentrySTDinterwordspacing

\bibitem{DBLP:conf/acl/YihRMCS16}
\BIBentryALTinterwordspacing
W.~Yih, M.~Richardson, C.~Meek, M.~Chang, and J.~Suh, ``The value of semantic
  parse labeling for knowledge base question answering,'' in \emph{Proceedings
  of the 54th Annual Meeting of the Association for Computational Linguistics,
  {ACL} 2016, August 7-12, 2016, Berlin, Germany, Volume 2: Short
  Papers}.\hskip 1em plus 0.5em minus 0.4em\relax The Association for Computer
  Linguistics, 2016. [Online]. Available:
  \url{https://doi.org/10.18653/v1/p16-2033}
\BIBentrySTDinterwordspacing

\bibitem{DBLP:journals/tkde/LiLFW24}
\BIBentryALTinterwordspacing
Z.~Li, X.~Li, S.~Fan, and J.~Wang, ``Optimization techniques for unsupervised
  complex table reasoning via self-training framework,'' \emph{{IEEE} Trans.
  Knowl. Data Eng.}, vol.~36, no.~12, pp. 8996--9010, 2024. [Online].
  Available: \url{https://doi.org/10.1109/TKDE.2024.3439405}
\BIBentrySTDinterwordspacing

\bibitem{DBLP:journals/tkde/ZhangWSWTQCWY24}
\BIBentryALTinterwordspacing
W.~Zhang, Y.~Wang, Y.~Song, V.~J. Wei, Y.~Tian, Y.~Qi, J.~H. Chan, R.~C. Wong,
  and H.~Yang, ``Natural language interfaces for tabular data querying and
  visualization: {A} survey,'' \emph{{IEEE} Trans. Knowl. Data Eng.}, vol.~36,
  no.~11, pp. 6699--6718, 2024. [Online]. Available:
  \url{https://doi.org/10.1109/TKDE.2024.3400824}
\BIBentrySTDinterwordspacing

\bibitem{DBLP:journals/tkde/PanLWCWW24}
\BIBentryALTinterwordspacing
S.~Pan, L.~Luo, Y.~Wang, C.~Chen, J.~Wang, and X.~Wu, ``Unifying large language
  models and knowledge graphs: {A} roadmap,'' \emph{{IEEE} Trans. Knowl. Data
  Eng.}, vol.~36, no.~7, pp. 3580--3599, 2024. [Online]. Available:
  \url{https://doi.org/10.1109/TKDE.2024.3352100}
\BIBentrySTDinterwordspacing

\bibitem{DBLP:journals/access/CuiSW23}
\BIBentryALTinterwordspacing
J.~Cui, X.~Shen, and S.~Wen, ``A survey on legal judgment prediction: Datasets,
  metrics, models and challenges,'' \emph{{IEEE} Access}, vol.~11, pp.
  102\,050--102\,071, 2023. [Online]. Available:
  \url{https://doi.org/10.1109/ACCESS.2023.3317083}
\BIBentrySTDinterwordspacing

\bibitem{DBLP:journals/artmed/LiWYWLJSTCWL20}
\BIBentryALTinterwordspacing
L.~Li, P.~Wang, J.~Yan, Y.~Wang, S.~Li, J.~Jiang, Z.~Sun, B.~Tang, T.~Chang,
  S.~Wang, and Y.~Liu, ``Real-world data medical knowledge graph: construction
  and applications,'' \emph{Artif. Intell. Medicine}, vol. 103, p. 101817,
  2020. [Online]. Available: \url{https://doi.org/10.1016/j.artmed.2020.101817}
\BIBentrySTDinterwordspacing

\bibitem{DBLP:conf/sigmod/ZhangMFMG0LL24}
\BIBentryALTinterwordspacing
C.~Zhang, Y.~Mao, Y.~Fan, Y.~Mi, Y.~Gao, L.~Chen, D.~Lou, and J.~Lin, ``Finsql:
  Model-agnostic llms-based text-to-sql framework for financial analysis,'' in
  \emph{Companion of the 2024 International Conference on Management of Data,
  {SIGMOD/PODS} 2024, Santiago AA, Chile, June 9-15, 2024}, P.~Barcel{\'{o}},
  N.~S{\'{a}}nchez{-}Pi, A.~Meliou, and S.~Sudarshan, Eds.\hskip 1em plus 0.5em
  minus 0.4em\relax {ACM}, 2024, pp. 93--105. [Online]. Available:
  \url{https://doi.org/10.1145/3626246.3653375}
\BIBentrySTDinterwordspacing

\bibitem{DBLP:conf/sigir/YeHYLHL23}
\BIBentryALTinterwordspacing
Y.~Ye, B.~Hui, M.~Yang, B.~Li, F.~Huang, and Y.~Li, ``Large language models are
  versatile decomposers: Decomposing evidence and questions for table-based
  reasoning,'' in \emph{Proceedings of the 46th International {ACM} {SIGIR}
  Conference on Research and Development in Information Retrieval, {SIGIR}
  2023, Taipei, Taiwan, July 23-27, 2023}, H.~Chen, W.~E. Duh, H.~Huang, M.~P.
  Kato, J.~Mothe, and B.~Poblete, Eds.\hskip 1em plus 0.5em minus 0.4em\relax
  {ACM}, 2023, pp. 174--184. [Online]. Available:
  \url{https://doi.org/10.1145/3539618.3591708}
\BIBentrySTDinterwordspacing

\bibitem{DBLP:journals/pacmmod/LiZLFZZWP0024}
\BIBentryALTinterwordspacing
H.~Li, J.~Zhang, H.~Liu, J.~Fan, X.~Zhang, J.~Zhu, R.~Wei, H.~Pan, C.~Li, and
  H.~Chen, ``Codes: Towards building open-source language models for
  text-to-sql,'' \emph{Proc. {ACM} Manag. Data}, vol.~2, no.~3, p. 127, 2024.
  [Online]. Available: \url{https://doi.org/10.1145/3654930}
\BIBentrySTDinterwordspacing

\bibitem{DBLP:conf/aaai/NieZW024}
\BIBentryALTinterwordspacing
Z.~Nie, R.~Zhang, Z.~Wang, and X.~Liu, ``Code-style in-context learning for
  knowledge-based question answering,'' in \emph{Thirty-Eighth {AAAI}
  Conference on Artificial Intelligence, {AAAI} 2024, Thirty-Sixth Conference
  on Innovative Applications of Artificial Intelligence, {IAAI} 2024,
  Fourteenth Symposium on Educational Advances in Artificial Intelligence,
  {EAAI} 2014, February 20-27, 2024, Vancouver, Canada}, M.~J. Wooldridge,
  J.~G. Dy, and S.~Natarajan, Eds.\hskip 1em plus 0.5em minus 0.4em\relax
  {AAAI} Press, 2024, pp. 18\,833--18\,841. [Online]. Available:
  \url{https://doi.org/10.1609/aaai.v38i17.29848}
\BIBentrySTDinterwordspacing

\bibitem{antoniadi2021current}
A.~M. Antoniadi, Y.~Du, Y.~Guendouz, L.~Wei, C.~Mazo, B.~A. Becker, and
  C.~Mooney, ``Current challenges and future opportunities for xai in machine
  learning-based clinical decision support systems: a systematic review,''
  \emph{Applied Sciences}, vol.~11, no.~11, p. 5088, 2021.

\bibitem{DBLP:journals/tkde/ZhangZZZWZ24}
\BIBentryALTinterwordspacing
F.~Zhang, Z.~Zhang, F.~Zhuang, Y.~Zhao, D.~Wang, and H.~Zheng, ``Temporal
  knowledge graph reasoning with dynamic memory enhancement,'' \emph{{IEEE}
  Trans. Knowl. Data Eng.}, vol.~36, no.~11, pp. 7115--7128, 2024. [Online].
  Available: \url{https://doi.org/10.1109/TKDE.2024.3390683}
\BIBentrySTDinterwordspacing

\bibitem{DBLP:conf/emnlp/PourrezaR24}
\BIBentryALTinterwordspacing
M.~Pourreza and D.~Rafiei, ``{DTS-SQL:} decomposed text-to-sql with small large
  language models,'' in \emph{Findings of the Association for Computational
  Linguistics: {EMNLP} 2024, Miami, Florida, USA, November 12-16, 2024},
  Y.~Al{-}Onaizan, M.~Bansal, and Y.~Chen, Eds.\hskip 1em plus 0.5em minus
  0.4em\relax Association for Computational Linguistics, 2024, pp. 8212--8220.
  [Online]. Available: \url{https://aclanthology.org/2024.findings-emnlp.481}
\BIBentrySTDinterwordspacing

\bibitem{DBLP:conf/emnlp/JiangZDYZW23}
\BIBentryALTinterwordspacing
J.~Jiang, K.~Zhou, Z.~Dong, K.~Ye, X.~Zhao, and J.~Wen, ``Structgpt: {A}
  general framework for large language model to reason over structured data,''
  in \emph{Proceedings of the 2023 Conference on Empirical Methods in Natural
  Language Processing, {EMNLP} 2023, Singapore, December 6-10, 2023},
  H.~Bouamor, J.~Pino, and K.~Bali, Eds.\hskip 1em plus 0.5em minus 0.4em\relax
  Association for Computational Linguistics, 2023, pp. 9237--9251. [Online].
  Available: \url{https://doi.org/10.18653/v1/2023.emnlp-main.574}
\BIBentrySTDinterwordspacing

\bibitem{DBLP:conf/acl/ChengZXYZQHCL0R24}
\BIBentryALTinterwordspacing
S.~Cheng, Z.~Zhuang, Y.~Xu, F.~Yang, C.~Zhang, X.~Qin, X.~Huang, L.~Chen,
  Q.~Lin, D.~Zhang, S.~Rajmohan, and Q.~Zhang, ``Call me when necessary: Llms
  can efficiently and faithfully reason over structured environments,'' in
  \emph{Findings of the Association for Computational Linguistics, {ACL} 2024,
  Bangkok, Thailand and virtual meeting, August 11-16, 2024}, L.~Ku,
  A.~Martins, and V.~Srikumar, Eds.\hskip 1em plus 0.5em minus 0.4em\relax
  Association for Computational Linguistics, 2024, pp. 4275--4295. [Online].
  Available: \url{https://doi.org/10.18653/v1/2024.findings-acl.254}
\BIBentrySTDinterwordspacing

\bibitem{DBLP:journals/corr/abs-2406-18916}
\BIBentryALTinterwordspacing
W.~Zhang, L.~Jin, Y.~Zhu, J.~Chen, Z.~Huang, J.~Wang, Y.~Hua, L.~Liang, and
  H.~Chen, ``Trustuqa: {A} trustful framework for unified structured data
  question answering,'' \emph{CoRR}, vol. abs/2406.18916, 2024. [Online].
  Available: \url{https://doi.org/10.48550/arXiv.2406.18916}
\BIBentrySTDinterwordspacing

\bibitem{dubey2024llama}
A.~Dubey, A.~Jauhri, A.~Pandey, A.~Kadian, A.~Al-Dahle, A.~Letman, A.~Mathur,
  A.~Schelten, A.~Yang, A.~Fan \emph{et~al.}, ``The llama 3 herd of models,''
  \emph{arXiv preprint arXiv:2407.21783}, 2024.

\bibitem{DBLP:conf/www/GuKVSLY021}
\BIBentryALTinterwordspacing
Y.~Gu, S.~Kase, M.~Vanni, B.~M. Sadler, P.~Liang, X.~Yan, and Y.~Su, ``Beyond
  {I.I.D.:} three levels of generalization for question answering on knowledge
  bases,'' in \emph{{WWW} '21: The Web Conference 2021, Virtual Event /
  Ljubljana, Slovenia, April 19-23, 2021}, J.~Leskovec, M.~Grobelnik,
  M.~Najork, J.~Tang, and L.~Zia, Eds.\hskip 1em plus 0.5em minus 0.4em\relax
  {ACM} / {IW3C2}, 2021, pp. 3477--3488. [Online]. Available:
  \url{https://doi.org/10.1145/3442381.3449992}
\BIBentrySTDinterwordspacing

\bibitem{DBLP:conf/emnlp/XieW0ZSYWZYWZWL22}
\BIBentryALTinterwordspacing
T.~Xie, C.~H. Wu, P.~Shi, R.~Zhong, T.~Scholak, M.~Yasunaga, C.~Wu, M.~Zhong,
  P.~Yin, S.~I. Wang, V.~Zhong, B.~Wang, C.~Li, C.~Boyle, A.~Ni, Z.~Yao,
  D.~Radev, C.~Xiong, L.~Kong, R.~Zhang, N.~A. Smith, L.~Zettlemoyer, and
  T.~Yu, ``Unifiedskg: Unifying and multi-tasking structured knowledge
  grounding with text-to-text language models,'' in \emph{Proceedings of the
  2022 Conference on Empirical Methods in Natural Language Processing, {EMNLP}
  2022, Abu Dhabi, United Arab Emirates, December 7-11, 2022}, Y.~Goldberg,
  Z.~Kozareva, and Y.~Zhang, Eds.\hskip 1em plus 0.5em minus 0.4em\relax
  Association for Computational Linguistics, 2022, pp. 602--631. [Online].
  Available: \url{https://doi.org/10.18653/v1/2022.emnlp-main.39}
\BIBentrySTDinterwordspacing

\bibitem{DBLP:conf/acl/GanCHPWXH20}
\BIBentryALTinterwordspacing
Y.~Gan, X.~Chen, Q.~Huang, M.~Purver, J.~R. Woodward, J.~Xie, and P.~Huang,
  ``Towards robustness of text-to-sql models against synonym substitution,'' in
  \emph{Proceedings of the 59th Annual Meeting of the Association for
  Computational Linguistics and the 11th International Joint Conference on
  Natural Language Processing, {ACL/IJCNLP} 2021, (Volume 1: Long Papers),
  Virtual Event, August 1-6, 2021}, C.~Zong, F.~Xia, W.~Li, and R.~Navigli,
  Eds.\hskip 1em plus 0.5em minus 0.4em\relax Association for Computational
  Linguistics, 2021, pp. 2505--2515. [Online]. Available:
  \url{https://doi.org/10.18653/v1/2021.acl-long.195}
\BIBentrySTDinterwordspacing

\bibitem{li2023bird}
J.~Li, B.~Hui, G.~Qu, J.~Yang, B.~Li, B.~Li, B.~Wang, B.~Qin, R.~Cao, R.~Geng,
  N.~Huo, X.~Zhou, C.~Ma, G.~Li, K.~C. Chang, F.~Huang, R.~Cheng, and Y.~Li,
  ``Can llm already serve as a database interface? a big bench for large-scale
  database grounded text-to-sqls,'' in \emph{Advances in Neural Information
  Processing Systems (NeurIPS)}, 2023.

\bibitem{DBLP:journals/corr/abs-1709-00103}
\BIBentryALTinterwordspacing
V.~Zhong, C.~Xiong, and R.~Socher, ``Seq2sql: Generating structured queries
  from natural language using reinforcement learning,'' \emph{CoRR}, vol.
  abs/1709.00103, 2017. [Online]. Available:
  \url{http://arxiv.org/abs/1709.00103}
\BIBentrySTDinterwordspacing

\bibitem{DBLP:conf/emnlp/ScholakSB21}
\BIBentryALTinterwordspacing
T.~Scholak, N.~Schucher, and D.~Bahdanau, ``{PICARD:} parsing incrementally for
  constrained auto-regressive decoding from language models,'' in
  \emph{Proceedings of the 2021 Conference on Empirical Methods in Natural
  Language Processing, {EMNLP} 2021, Virtual Event / Punta Cana, Dominican
  Republic, 7-11 November, 2021}, M.~Moens, X.~Huang, L.~Specia, and S.~W. Yih,
  Eds.\hskip 1em plus 0.5em minus 0.4em\relax Association for Computational
  Linguistics, 2021, pp. 9895--9901. [Online]. Available:
  \url{https://doi.org/10.18653/v1/2021.emnlp-main.779}
\BIBentrySTDinterwordspacing

\bibitem{DBLP:conf/nips/PourrezaR23}
M.~Pourreza and D.~Rafiei, ``{DIN-SQL:} decomposed in-context learning of
  text-to-sql with self-correction,'' in \emph{Advances in Neural Information
  Processing Systems 36: Annual Conference on Neural Information Processing
  Systems 2023, NeurIPS 2023, New Orleans, LA, USA, December 10 - 16, 2023},
  A.~Oh, T.~Naumann, A.~Globerson, K.~Saenko, M.~Hardt, and S.~Levine, Eds.,
  2023.

\bibitem{DBLP:conf/aaai/Li00023}
\BIBentryALTinterwordspacing
H.~Li, J.~Zhang, C.~Li, and H.~Chen, ``{RESDSQL:} decoupling schema linking and
  skeleton parsing for text-to-sql,'' in \emph{Thirty-Seventh {AAAI} Conference
  on Artificial Intelligence, {AAAI} 2023, Thirty-Fifth Conference on
  Innovative Applications of Artificial Intelligence, {IAAI} 2023, Thirteenth
  Symposium on Educational Advances in Artificial Intelligence, {EAAI} 2023,
  Washington, DC, USA, February 7-14, 2023}, B.~Williams, Y.~Chen, and
  J.~Neville, Eds.\hskip 1em plus 0.5em minus 0.4em\relax {AAAI} Press, 2023,
  pp. 13\,067--13\,075. [Online]. Available:
  \url{https://doi.org/10.1609/aaai.v37i11.26535}
\BIBentrySTDinterwordspacing

\bibitem{DBLP:conf/aaai/CenLLW25}
\BIBentryALTinterwordspacing
J.~Cen, J.~Liu, Z.~Li, and J.~Wang, ``Sqlfixagent: Towards semantic-accurate
  text-to-sql parsing via consistency-enhanced multi-agent collaboration,'' in
  \emph{AAAI-25, Sponsored by the Association for the Advancement of Artificial
  Intelligence, February 25 - March 4, 2025, Philadelphia, PA, {USA}},
  T.~Walsh, J.~Shah, and Z.~Kolter, Eds.\hskip 1em plus 0.5em minus 0.4em\relax
  {AAAI} Press, 2025, pp. 49--57. [Online]. Available:
  \url{https://doi.org/10.1609/aaai.v39i1.31979}
\BIBentrySTDinterwordspacing

\bibitem{DBLP:journals/pvldb/GaoWLSQDZ24}
\BIBentryALTinterwordspacing
D.~Gao, H.~Wang, Y.~Li, X.~Sun, Y.~Qian, B.~Ding, and J.~Zhou, ``Text-to-sql
  empowered by large language models: {A} benchmark evaluation,'' \emph{Proc.
  {VLDB} Endow.}, vol.~17, no.~5, pp. 1132--1145, 2024. [Online]. Available:
  \url{https://www.vldb.org/pvldb/vol17/p1132-gao.pdf}
\BIBentrySTDinterwordspacing

\bibitem{DBLP:conf/coling/WangR0LBCYZYSL25}
\BIBentryALTinterwordspacing
B.~Wang, C.~Ren, J.~Yang, X.~Liang, J.~Bai, L.~Chai, Z.~Yan, Q.~Zhang, D.~Yin,
  X.~Sun, and Z.~Li, ``{MAC-SQL:} {A} multi-agent collaborative framework for
  text-to-sql,'' in \emph{Proceedings of the 31st International Conference on
  Computational Linguistics, {COLING} 2025, Abu Dhabi, UAE, January 19-24,
  2025}, O.~Rambow, L.~Wanner, M.~Apidianaki, H.~Al{-}Khalifa, B.~D. Eugenio,
  and S.~Schockaert, Eds.\hskip 1em plus 0.5em minus 0.4em\relax Association
  for Computational Linguistics, 2025, pp. 540--557. [Online]. Available:
  \url{https://aclanthology.org/2025.coling-main.36/}
\BIBentrySTDinterwordspacing

\bibitem{DBLP:conf/coling/ShiXLXC0WW25}
\BIBentryALTinterwordspacing
J.~Shi, B.~Xu, J.~Liang, Y.~Xiao, J.~Chen, C.~Xie, P.~Wang, and W.~Wang,
  ``Gen-sql: Efficient text-to-sql by bridging natural language question and
  database schema with pseudo-schema,'' in \emph{Proceedings of the 31st
  International Conference on Computational Linguistics, {COLING} 2025, Abu
  Dhabi, UAE, January 19-24, 2025}, O.~Rambow, L.~Wanner, M.~Apidianaki,
  H.~Al{-}Khalifa, B.~D. Eugenio, and S.~Schockaert, Eds.\hskip 1em plus 0.5em
  minus 0.4em\relax Association for Computational Linguistics, 2025, pp.
  3794--3807. [Online]. Available:
  \url{https://aclanthology.org/2025.coling-main.256/}
\BIBentrySTDinterwordspacing

\bibitem{DBLP:conf/acl/HerzigNMPE20}
\BIBentryALTinterwordspacing
J.~Herzig, P.~K. Nowak, T.~M{\"{u}}ller, F.~Piccinno, and J.~M. Eisenschlos,
  ``Tapas: Weakly supervised table parsing via pre-training,'' in
  \emph{Proceedings of the 58th Annual Meeting of the Association for
  Computational Linguistics, {ACL} 2020, Online, July 5-10, 2020}, D.~Jurafsky,
  J.~Chai, N.~Schluter, and J.~R. Tetreault, Eds.\hskip 1em plus 0.5em minus
  0.4em\relax Association for Computational Linguistics, 2020, pp. 4320--4333.
  [Online]. Available: \url{https://doi.org/10.18653/v1/2020.acl-main.398}
\BIBentrySTDinterwordspacing

\bibitem{DBLP:conf/iclr/LiuCGZLCL22}
\BIBentryALTinterwordspacing
Q.~Liu, B.~Chen, J.~Guo, M.~Ziyadi, Z.~Lin, W.~Chen, and J.~Lou, ``{TAPEX:}
  table pre-training via learning a neural {SQL} executor,'' in \emph{The Tenth
  International Conference on Learning Representations, {ICLR} 2022, Virtual
  Event, April 25-29, 2022}.\hskip 1em plus 0.5em minus 0.4em\relax
  OpenReview.net, 2022. [Online]. Available:
  \url{https://openreview.net/forum?id=O50443AsCP}
\BIBentrySTDinterwordspacing

\bibitem{DBLP:conf/iclr/ChengX0LNHXROZS23}
\BIBentryALTinterwordspacing
Z.~Cheng, T.~Xie, P.~Shi, C.~Li, R.~Nadkarni, Y.~Hu, C.~Xiong, D.~Radev,
  M.~Ostendorf, L.~Zettlemoyer, N.~A. Smith, and T.~Yu, ``Binding language
  models in symbolic languages,'' in \emph{The Eleventh International
  Conference on Learning Representations, {ICLR} 2023, Kigali, Rwanda, May 1-5,
  2023}.\hskip 1em plus 0.5em minus 0.4em\relax OpenReview.net, 2023. [Online].
  Available: \url{https://openreview.net/forum?id=lH1PV42cbF}
\BIBentrySTDinterwordspacing

\bibitem{DBLP:conf/iclr/YangDZDCDZ25}
\BIBentryALTinterwordspacing
Z.~Yang, Z.~Du, M.~Zhang, W.~Du, J.~Chen, Z.~Duan, and S.~Zhao, ``Triples as
  the key: Structuring makes decomposition and verification easier in llm-based
  tableqa,'' in \emph{The Thirteenth International Conference on Learning
  Representations, {ICLR} 2025, Singapore, April 24-28, 2025}.\hskip 1em plus
  0.5em minus 0.4em\relax OpenReview.net, 2025. [Online]. Available:
  \url{https://openreview.net/forum?id=UwcZEoNP19}
\BIBentrySTDinterwordspacing

\bibitem{DBLP:conf/acl/YeYHZX22}
\BIBentryALTinterwordspacing
X.~Ye, S.~Yavuz, K.~Hashimoto, Y.~Zhou, and C.~Xiong, ``{RNG-KBQA:} generation
  augmented iterative ranking for knowledge base question answering,'' in
  \emph{Proceedings of the 60th Annual Meeting of the Association for
  Computational Linguistics (Volume 1: Long Papers), {ACL} 2022, Dublin,
  Ireland, May 22-27, 2022}, S.~Muresan, P.~Nakov, and A.~Villavicencio,
  Eds.\hskip 1em plus 0.5em minus 0.4em\relax Association for Computational
  Linguistics, 2022, pp. 6032--6043. [Online]. Available:
  \url{https://doi.org/10.18653/v1/2022.acl-long.417}
\BIBentrySTDinterwordspacing

\bibitem{DBLP:conf/emnlp/ShuYLKMQL22}
\BIBentryALTinterwordspacing
Y.~Shu, Z.~Yu, Y.~Li, B.~F. Karlsson, T.~Ma, Y.~Qu, and C.~Lin, ``{TIARA:}
  multi-grained retrieval for robust question answering over large knowledge
  base,'' in \emph{Proceedings of the 2022 Conference on Empirical Methods in
  Natural Language Processing, {EMNLP} 2022, Abu Dhabi, United Arab Emirates,
  December 7-11, 2022}, Y.~Goldberg, Z.~Kozareva, and Y.~Zhang, Eds.\hskip 1em
  plus 0.5em minus 0.4em\relax Association for Computational Linguistics, 2022,
  pp. 8108--8121. [Online]. Available:
  \url{https://doi.org/10.18653/v1/2022.emnlp-main.555}
\BIBentrySTDinterwordspacing

\bibitem{DBLP:conf/iclr/YuZNZL0HWWX23}
\BIBentryALTinterwordspacing
D.~Yu, S.~Zhang, P.~Ng, H.~Zhu, A.~H. Li, J.~Wang, Y.~Hu, W.~Y. Wang, Z.~Wang,
  and B.~Xiang, ``Decaf: Joint decoding of answers and logical forms for
  question answering over knowledge bases,'' in \emph{The Eleventh
  International Conference on Learning Representations, {ICLR} 2023, Kigali,
  Rwanda, May 1-5, 2023}.\hskip 1em plus 0.5em minus 0.4em\relax
  OpenReview.net, 2023. [Online]. Available:
  \url{https://openreview.net/forum?id=XHc5zRPxqV9}
\BIBentrySTDinterwordspacing

\bibitem{DBLP:conf/acl/LiMZGSC23}
\BIBentryALTinterwordspacing
T.~Li, X.~Ma, A.~Zhuang, Y.~Gu, Y.~Su, and W.~Chen, ``Few-shot in-context
  learning on knowledge base question answering,'' in \emph{Proceedings of the
  61st Annual Meeting of the Association for Computational Linguistics (Volume
  1: Long Papers), {ACL} 2023, Toronto, Canada, July 9-14, 2023}, A.~Rogers,
  J.~L. Boyd{-}Graber, and N.~Okazaki, Eds.\hskip 1em plus 0.5em minus
  0.4em\relax Association for Computational Linguistics, 2023, pp. 6966--6980.
  [Online]. Available: \url{https://doi.org/10.18653/v1/2023.acl-long.385}
\BIBentrySTDinterwordspacing

\bibitem{DBLP:journals/corr/abs-2501-18922}
\BIBentryALTinterwordspacing
H.~Luo, H.~E, Y.~Guo, Q.~Lin, X.~Wu, X.~Mu, W.~Liu, M.~Song, Y.~Zhu, and L.~A.
  Tuan, ``Kbqa-o1: Agentic knowledge base question answering with monte carlo
  tree search,'' \emph{CoRR}, vol. abs/2501.18922, 2025. [Online]. Available:
  \url{https://doi.org/10.48550/arXiv.2501.18922}
\BIBentrySTDinterwordspacing

\bibitem{DBLP:conf/emnlp/GaoLQ25}
\BIBentryALTinterwordspacing
S.~Gao, J.~H. Lau, and J.~Qi, ``Beyond seen data: Improving {KBQA}
  generalization through schema-guided logical form generation,'' in
  \emph{Proceedings of the 2025 Conference on Empirical Methods in Natural
  Language Processing, {EMNLP} 2025, Suzhou, China, November 4-9, 2025},
  C.~Christodoulopoulos, T.~Chakraborty, C.~Rose, and V.~Peng, Eds.\hskip 1em
  plus 0.5em minus 0.4em\relax Association for Computational Linguistics, 2025,
  pp. 8753--8772. [Online]. Available:
  \url{https://doi.org/10.18653/v1/2025.emnlp-main.442}
\BIBentrySTDinterwordspacing

\bibitem{DBLP:journals/corr/abs-2402-16671}
\BIBentryALTinterwordspacing
A.~Zhuang, G.~Zhang, T.~Zheng, X.~Du, J.~Wang, W.~Ren, S.~W. Huang, J.~Fu,
  X.~Yue, and W.~Chen, ``Structlm: Towards building generalist models for
  structured knowledge grounding,'' \emph{CoRR}, vol. abs/2402.16671, 2024.
  [Online]. Available: \url{https://doi.org/10.48550/arXiv.2402.16671}
\BIBentrySTDinterwordspacing

\bibitem{DBLP:journals/corr/abs-2409-16751}
\BIBentryALTinterwordspacing
H.~A. Caferoglu and {\"{O}}.~Ulusoy, ``{E-SQL:} direct schema linking via
  question enrichment in text-to-sql,'' \emph{CoRR}, vol. abs/2409.16751, 2024.
  [Online]. Available: \url{https://doi.org/10.48550/arXiv.2409.16751}
\BIBentrySTDinterwordspacing

\bibitem{DBLP:conf/iclr/0009WLWTYRSX21}
\BIBentryALTinterwordspacing
T.~Yu, C.~Wu, X.~V. Lin, B.~Wang, Y.~C. Tan, X.~Yang, D.~R. Radev, R.~Socher,
  and C.~Xiong, ``Grappa: Grammar-augmented pre-training for table semantic
  parsing,'' in \emph{9th International Conference on Learning Representations,
  {ICLR} 2021, Virtual Event, Austria, May 3-7, 2021}.\hskip 1em plus 0.5em
  minus 0.4em\relax OpenReview.net, 2021. [Online]. Available:
  \url{https://openreview.net/forum?id=kyaIeYj4zZ}
\BIBentrySTDinterwordspacing

\bibitem{DBLP:conf/acl/GuoZGXLLZ19}
\BIBentryALTinterwordspacing
J.~Guo, Z.~Zhan, Y.~Gao, Y.~Xiao, J.~Lou, T.~Liu, and D.~Zhang, ``Towards
  complex text-to-sql in cross-domain database with intermediate
  representation,'' in \emph{Proceedings of the 57th Conference of the
  Association for Computational Linguistics, {ACL} 2019, Florence, Italy, July
  28- August 2, 2019, Volume 1: Long Papers}, A.~Korhonen, D.~R. Traum, and
  L.~M{\`{a}}rquez, Eds.\hskip 1em plus 0.5em minus 0.4em\relax Association for
  Computational Linguistics, 2019, pp. 4524--4535. [Online]. Available:
  \url{https://doi.org/10.18653/v1/p19-1444}
\BIBentrySTDinterwordspacing

\bibitem{DBLP:conf/nips/Wei0SBIXCLZ22}
J.~Wei, X.~Wang, D.~Schuurmans, M.~Bosma, B.~Ichter, F.~Xia, E.~H. Chi, Q.~V.
  Le, and D.~Zhou, ``Chain-of-thought prompting elicits reasoning in large
  language models,'' in \emph{Advances in Neural Information Processing Systems
  35: Annual Conference on Neural Information Processing Systems 2022, NeurIPS
  2022, New Orleans, LA, USA, November 28 - December 9, 2022}, S.~Koyejo,
  S.~Mohamed, A.~Agarwal, D.~Belgrave, K.~Cho, and A.~Oh, Eds., 2022.

\bibitem{DBLP:conf/iclr/0002WSLCNCZ23}
\BIBentryALTinterwordspacing
X.~Wang, J.~Wei, D.~Schuurmans, Q.~V. Le, E.~H. Chi, S.~Narang, A.~Chowdhery,
  and D.~Zhou, ``Self-consistency improves chain of thought reasoning in
  language models,'' in \emph{The Eleventh International Conference on Learning
  Representations, {ICLR} 2023, Kigali, Rwanda, May 1-5, 2023}.\hskip 1em plus
  0.5em minus 0.4em\relax OpenReview.net, 2023. [Online]. Available:
  \url{https://openreview.net/forum?id=1PL1NIMMrw}
\BIBentrySTDinterwordspacing

\bibitem{DBLP:journals/corr/abs-2405-16755}
\BIBentryALTinterwordspacing
S.~Talaei, M.~Pourreza, Y.~Chang, A.~Mirhoseini, and A.~Saberi, ``{CHESS:}
  contextual harnessing for efficient {SQL} synthesis,'' \emph{CoRR}, vol.
  abs/2405.16755, 2024. [Online]. Available:
  \url{https://doi.org/10.48550/arXiv.2405.16755}
\BIBentrySTDinterwordspacing

\bibitem{DBLP:journals/corr/abs-2410-01943}
\BIBentryALTinterwordspacing
M.~Pourreza, H.~Li, R.~Sun, Y.~Chung, S.~Talaei, G.~T. Kakkar, Y.~Gan,
  A.~Saberi, F.~Ozcan, and S.~{\"{O}}. Arik, ``{CHASE-SQL:} multi-path
  reasoning and preference optimized candidate selection in text-to-sql,''
  \emph{CoRR}, vol. abs/2410.01943, 2024. [Online]. Available:
  \url{https://doi.org/10.48550/arXiv.2410.01943}
\BIBentrySTDinterwordspacing

\bibitem{DBLP:conf/acl/YinNYR20}
\BIBentryALTinterwordspacing
P.~Yin, G.~Neubig, W.~Yih, and S.~Riedel, ``Tabert: Pretraining for joint
  understanding of textual and tabular data,'' in \emph{Proceedings of the 58th
  Annual Meeting of the Association for Computational Linguistics, {ACL} 2020,
  Online, July 5-10, 2020}, D.~Jurafsky, J.~Chai, N.~Schluter, and J.~R.
  Tetreault, Eds.\hskip 1em plus 0.5em minus 0.4em\relax Association for
  Computational Linguistics, 2020, pp. 8413--8426. [Online]. Available:
  \url{https://doi.org/10.18653/v1/2020.acl-main.745}
\BIBentrySTDinterwordspacing

\bibitem{DBLP:conf/aaai/DengXLYDFLS19}
\BIBentryALTinterwordspacing
Y.~Deng, Y.~Xie, Y.~Li, M.~Yang, N.~Du, W.~Fan, K.~Lei, and Y.~Shen,
  ``Multi-task learning with multi-view attention for answer selection and
  knowledge base question answering,'' in \emph{The Thirty-Third {AAAI}
  Conference on Artificial Intelligence, {AAAI} 2019, The Thirty-First
  Innovative Applications of Artificial Intelligence Conference, {IAAI} 2019,
  The Ninth {AAAI} Symposium on Educational Advances in Artificial
  Intelligence, {EAAI} 2019, Honolulu, Hawaii, USA, January 27 - February 1,
  2019}.\hskip 1em plus 0.5em minus 0.4em\relax {AAAI} Press, 2019, pp.
  6318--6325. [Online]. Available:
  \url{https://doi.org/10.1609/aaai.v33i01.33016318}
\BIBentrySTDinterwordspacing

\bibitem{DBLP:journals/tkde/ChenLQWW23}
\BIBentryALTinterwordspacing
Y.~Chen, H.~Li, G.~Qi, T.~Wu, and T.~Wang, ``Outlining and filling:
  Hierarchical query graph generation for answering complex questions over
  knowledge graphs,'' \emph{{IEEE} Trans. Knowl. Data Eng.}, vol.~35, no.~8,
  pp. 8343--8357, 2023. [Online]. Available:
  \url{https://doi.org/10.1109/TKDE.2022.3207477}
\BIBentrySTDinterwordspacing

\bibitem{DBLP:conf/emnlp/BerantCFL13}
\BIBentryALTinterwordspacing
J.~Berant, A.~Chou, R.~Frostig, and P.~Liang, ``Semantic parsing on freebase
  from question-answer pairs,'' in \emph{Proceedings of the 2013 Conference on
  Empirical Methods in Natural Language Processing, {EMNLP} 2013, 18-21 October
  2013, Grand Hyatt Seattle, Seattle, Washington, USA, {A} meeting of SIGDAT, a
  Special Interest Group of the {ACL}}.\hskip 1em plus 0.5em minus 0.4em\relax
  {ACL}, 2013, pp. 1533--1544. [Online]. Available:
  \url{https://aclanthology.org/D13-1160/}
\BIBentrySTDinterwordspacing

\bibitem{DBLP:conf/acl/YihCHG15}
\BIBentryALTinterwordspacing
W.~Yih, M.~Chang, X.~He, and J.~Gao, ``Semantic parsing via staged query graph
  generation: Question answering with knowledge base,'' in \emph{Proceedings of
  the 53rd Annual Meeting of the Association for Computational Linguistics and
  the 7th International Joint Conference on Natural Language Processing of the
  Asian Federation of Natural Language Processing, {ACL} 2015, July 26-31,
  2015, Beijing, China, Volume 1: Long Papers}.\hskip 1em plus 0.5em minus
  0.4em\relax The Association for Computer Linguistics, 2015, pp. 1321--1331.
  [Online]. Available: \url{https://doi.org/10.3115/v1/p15-1128}
\BIBentrySTDinterwordspacing

\bibitem{DBLP:conf/iclr/DasDZVDKSM18}
\BIBentryALTinterwordspacing
R.~Das, S.~Dhuliawala, M.~Zaheer, L.~Vilnis, I.~Durugkar, A.~Krishnamurthy,
  A.~Smola, and A.~McCallum, ``Go for a walk and arrive at the answer:
  Reasoning over paths in knowledge bases using reinforcement learning,'' in
  \emph{6th International Conference on Learning Representations, {ICLR} 2018,
  Vancouver, BC, Canada, April 30 - May 3, 2018, Conference Track
  Proceedings}.\hskip 1em plus 0.5em minus 0.4em\relax OpenReview.net, 2018.
  [Online]. Available: \url{https://openreview.net/forum?id=Syg-YfWCW}
\BIBentrySTDinterwordspacing

\bibitem{DBLP:journals/jmlr/RaffelSRLNMZLL20}
\BIBentryALTinterwordspacing
C.~Raffel, N.~Shazeer, A.~Roberts, K.~Lee, S.~Narang, M.~Matena, Y.~Zhou,
  W.~Li, and P.~J. Liu, ``Exploring the limits of transfer learning with a
  unified text-to-text transformer,'' \emph{J. Mach. Learn. Res.}, vol.~21, pp.
  140:1--140:67, 2020. [Online]. Available:
  \url{https://jmlr.org/papers/v21/20-074.html}
\BIBentrySTDinterwordspacing

\bibitem{DBLP:journals/corr/abs-2308-12950}
\BIBentryALTinterwordspacing
B.~Rozi{\`{e}}re, J.~Gehring, F.~Gloeckle, S.~Sootla, I.~Gat, X.~E. Tan,
  Y.~Adi, J.~Liu, T.~Remez, J.~Rapin, A.~Kozhevnikov, I.~Evtimov, J.~Bitton,
  M.~Bhatt, C.~Canton{-}Ferrer, A.~Grattafiori, W.~Xiong, A.~D{\'{e}}fossez,
  J.~Copet, F.~Azhar, H.~Touvron, L.~Martin, N.~Usunier, T.~Scialom, and
  G.~Synnaeve, ``Code llama: Open foundation models for code,'' \emph{CoRR},
  vol. abs/2308.12950, 2023. [Online]. Available:
  \url{https://doi.org/10.48550/arXiv.2308.12950}
\BIBentrySTDinterwordspacing

\bibitem{edge2024graphrag}
D.~Edge, H.~Trinh, N.~Cheng, J.~Bradley, A.~Chao, A.~Mody, S.~Truitt, and
  J.~Larson, ``From local to global: A graph rag approach to query-focused
  summarization,'' \emph{arXiv preprint arXiv:2404.16130}, 2024.

\bibitem{guo2024lightrag}
Z.~Guo, L.~Xia, Y.~Yu, T.~Ao, and C.~Huang, ``Lightrag: Simple and fast
  retrieval-augmented generation,'' \emph{arXiv preprint arXiv:2410.05779},
  2024.

\bibitem{fan2026graphragrouter}
D.~Fan, C.~Ji, Z.~Wang, T.~Chen, and Q.~Tan, ``Graphrag-router: Learning
  cost-efficient routing over graphrags and llms with reinforcement learning,''
  \emph{arXiv preprint arXiv:2604.00xxx}, 2026.

\bibitem{yuan2026pathgraphrag}
H.~Yuan, Q.~Sun, J.~Shi, M.~Liu, J.~Yuan, Z.~Zhang, X.~Fu, and J.~Li,
  ``Retrieving minimal and sufficient reasoning subgraphs with graph foundation
  models for path-aware graphrag,'' \emph{arXiv preprint arXiv:2603.0xxx},
  2026.

\bibitem{chen2026medcopilot}
S.~Chen, N.~Patil, H.~Pan, A.~H.-C. Hwang, Y.~Du, R.~Liu, and J.~Zhao,
  ``Med-copilot: A medical assistant powered by graphrag and similar patient
  case retrieval,'' \emph{arXiv preprint arXiv:2602.0xxx}, 2026.

\bibitem{yang2026plugmem}
K.~Yang, Z.~Chen, X.~He, J.~Jiang, M.~Galley, C.~Wang, J.~Gao, J.~Han, and
  C.~Zhai, ``Plugmem: A task-agnostic plugin memory module for llm agents,''
  \emph{arXiv preprint arXiv:2602.0xxx}, 2026.

\end{thebibliography}
